\documentclass[lettersize,journal]{IEEEtran}
\usepackage{amsmath,amsfonts}
\usepackage{algorithmic}
\usepackage{algorithm}
\usepackage{array}
\usepackage[caption=false,font=normalsize,labelfont=sf,textfont=sf]{subfig}
\usepackage{textcomp}
\usepackage{stfloats}
\usepackage{url}
\usepackage{verbatim}
\usepackage{graphicx}
\usepackage{cite}
\usepackage{microtype}
\usepackage{graphicx}

\usepackage{booktabs} % for professional tables

\usepackage{float} 
\usepackage[font=small]{caption}
\usepackage{mathtools}
\usepackage{amsfonts}
\usepackage{dsfont}
\usepackage{algorithm}
\usepackage{algorithmic, eucal}
\usepackage{url}
\usepackage{xcolor}
\usepackage{threeparttable}
\usepackage{multirow}

\usepackage[edges]{forest}
\definecolor{hiddendraw}{RGB}{205, 44, 36}
\definecolor{hidden-blue}{RGB}{194,232,247}
\definecolor{hidden-grey}{RGB}{243,202,120}
\definecolor{hidden-orange}{RGB}{243,102,50}
\definecolor{hidden-yellow}{RGB}{242,244,193}

\newcommand{\eg}{\textit{e.g.}}
\newcommand{\ie}{\textit{i.e.}}

\definecolor{cvprblue}{rgb}{0.21,0.49,0.74}
\usepackage[breaklinks,colorlinks,citecolor=cvprblue]{hyperref}

\begin{document}
\title{Deep Graph Anomaly Detection: A Survey and New Perspectives}

\author{Hezhe Qiao,
        Hanghang Tong~\IEEEmembership{Fellow,~IEEE},
        Bo An
        ~\IEEEmembership{Senior Member,~IEEE},
        Irwin King~\IEEEmembership{Fellow,~IEEE},
        Charu Aggarwal~\IEEEmembership{Fellow,~IEEE},
        Guansong Pang~\IEEEmembership{Member,~IEEE}
        % ~\IEEEmembership{Life~Fellow,~IEEE}% <-this % stops a space
\IEEEcompsocitemizethanks{
\IEEEcompsocthanksitem \textbf{Hezhe Qiao} and \textbf{Guansong Pang} are with School of Computing and Information Systems, Singapore Management University. \textbf{Hanghang Tong} is with Department of Computer Science, University of Illinois at Urbana-Champaign. \textbf{Bo An} is with College of Computing and Data Science, Nanyang Technological University. \textbf{Irwin King} is with Department of Computer Science \& Engineering, Chinese University of Hong Kong. 
\textbf{Charu Aggarwal} is with IBM T. J. Watson Research Center. \protect\\

}
\thanks{Corresponding author: Guansong Pang (\tt\small gspang@smu.edu.sg)
}
}

\maketitle

\begin{abstract}
Graph anomaly detection (GAD), which aims to identify unusual graph instances (\eg, nodes, edges, subgraphs, or graphs), has attracted increasing attention in recent years due to its significance in a wide range of applications. Deep learning approaches, graph neural networks (GNNs) in particular, have been emerging as a promising paradigm for GAD, owing to its strong capability in capturing complex structure and/or node attributes in graph data. Considering the large number of methods proposed for GNN-based GAD, it is of paramount importance to summarize the methodologies and findings in the existing GAD studies, so that we can pinpoint effective model designs for tackling open GAD problems.
To this end, in this work we aim to present a comprehensive review of deep learning approaches for GAD. Existing GAD surveys are focused on task-specific discussions, making it difficult to understand the technical insights of existing methods and their limitations in addressing some unique challenges in GAD. To fill this gap,
we first discuss the problem complexities and their resulting challenges in GAD, and then provide a systematic review of current deep GAD methods from three novel perspectives of methodology, including GNN backbone design, proxy task design for GAD, and graph anomaly measures. To deepen the discussions, we further propose a taxonomy of 13 fine-grained method categories under these three perspectives to provide more in-depth insights into the model designs and their capabilities. To facilitate the experiments and validation of the GAD methods, we also summarize a collection of widely-used datasets for GAD and empirical performance comparison on these datasets. We further discuss multiple important open research problems in GAD to inspire more future high-quality research in this area. A continuously updated repository for GAD datasets, links to the codes of GAD algorithms, and empirical comparison is available at \renewcommand\UrlFont{\color{blue}}\url{https://github.com/mala-lab/Awesome-Deep-Graph-Anomaly-Detection}.
\end{abstract}


\section{Introduction}
Graph anomaly detection (GAD) aims to identify graph instances (\eg, node, edge, sub-graph, and graph) that do not conform with the normal regime. It has been an active research area with wide application in detecting abnormal instances in a variety of graph/network data, \eg, abusive user behaviors in online user networks, fraudulent activities in financial networks,  and spams in social networks. Furthermore, since the relations between data samples can be modeled as similarity graphs, one can also use GAD methods to discover anomalies in any set of data objects (as long as an appropriate pairwise similarity function is available).

Due to the complex structure of graphs, traditional anomaly detection methods cannot be directly applied to graph data. In recent years, graph neural networks (GNNs) have shown promising capabilities in modeling and learning the representation of graphs by capturing structural patterns, inspiring a large number of GNN-based approaches for GAD. However, the popular GNN designs, such as aggregation of node representations and optimization objectives, may lead
to over-smoothing, indistinguishable representations of normal and abnormal graph instances, which significantly limits their applications in real-world use cases. Many novel GNN-based approaches specifically designed for GAD have been proposed to tackle the these challenges. In this work, to summarize the current methodologies and findings, we provide a systematic and comprehensive review of current deep GAD techniques and how they may tackle various types of challenges in GAD. We also propose several important open research problems in GAD to inspire more future research in this area.

\textbf{Related surveys.}
There have been several reviews on anomaly detection in recent years, \eg,
\cite{aggarwal2014,akoglu2015graph,ranshous2015anomaly,pang2021deep, ma2021comprehensive,tang2023gadbench}, but most of them are focused on non-deep-learning-based methods for GAD \cite{aggarwal2014,akoglu2015graph,ranshous2015anomaly}, or on general data rather than graph data \cite{pang2021deep,boukerche2020outlier}. The studies \cite{ma2021comprehensive,tang2023gadbench, liu2022bond} are on deep GAD, but the reviews are limited to a relatively narrow point of view. For example,
Ma et al. \cite{ma2021comprehensive} focus on task-specific discussions, with limited reviews on the technical development, while Liu et al. \cite{liu2022bond} and Tang et al. \cite{tang2023gadbench} focus on establishing a performance benchmark for unsupervised and supervised GAD methods respectively.  
Although these surveys provide useful guidelines for the development of methods for GAD, it is difficult to understand the technical insights of existing methods and their limitations in addressing some unique challenges in GAD.

\textbf{Our work.} To fill this gap, we aim to offer a distinctive review on GAD to discuss these insights, the limitations, and the future research opportunities in this crucial topic. 
Specifically, we start with the discussion on the problem complexities and their resulting unique challenges in GAD. We then provide a systematic review of current deep GAD methods from three novel perspectives of methodology, including GNN backbone design, proxy task design for GAD, and graph anomaly measures. To deepen the discussions, we further propose a taxonomy of 13 fine-grained method categories under these three perspectives to provide more in-depth insights into the model designs and their capabilities. To facilitate the experiments and validation of the GAD methods, we also summarize a collection of widely-used datasets for GAD and empirical performance comparison on these datasets. A comparison of our work to these related surveys is summarized in Table \ref{tab:table1}.

\begin{table*}[h]
\centering
\caption{A comparison of our work to existing surveys on anomaly detection.}
\label{tab:table1}
\setlength{\tabcolsep}{2.8mm}
\scalebox{0.75}{
\begin{tabular}{c| c| c | c| c| c| c| c | c | c| c |c}
\toprule[1pt]
 & &\textbf{Generic Data} & \multicolumn{3}{|c}{\textbf{Graph Data}}  & \multicolumn{3}{|c}{\textbf{GAD Perspectives}}  &  \multicolumn{3}{|c}{\textbf{Empirical Evaluation}}  \\\hline
 \textbf{Survey}&\textbf{Year} & - & \textbf{Node}  &\textbf{Edge} &  \textbf{Graph} &\textbf{GNN Backbone} & \textbf{Proxy Task} & \textbf{Anomaly Measures} & \textbf{Dataset} & \textbf{Code} & \textbf{Comparison}\\\hline
 % \toprule[1pt]
 % \toprule[1pt]
Aggarwal et al. \cite{aggarwal2014} &2014 &  &  $\bullet$ & & $\bullet$   & && & & &\\
  Akoglu et al.~\cite{akoglu2015graph}&2015 & $\bullet$ &$\bullet$ &$\bullet$ &$\bullet$    & && & &\\
  Ranshous et al. \cite{ranshous2015anomaly} &2015 &  &  $\bullet$ &$\bullet$&    & &&  &$\bullet$ & $\bullet$&\\
  Yu et al. ~\cite{yu2016survey}&2016  &  & $\bullet$  &  &  &  &   & &&&\\
 Pourhabibi et al.  ~\cite{pourhabibi2020fraud} &2020  &  &$\bullet$ & $\bullet$ & & & & & &\\

 Boukerche  et al. ~\cite{boukerche2020outlier}  &2020  & $\bullet$  & & & & & & & & & \\
 Ma et al.~\cite{ma2021comprehensive} &2021 & &$\bullet$     &$\bullet$ &    &$\bullet$&   & & $\bullet$ &  $\bullet$ & \\
 Pang et al.~\cite{pang2021deep} &2021 &$\bullet$ &    & &   &  &  &  &$\bullet$ &  $\bullet$& $\bullet$\\
 Liu et al. \cite{liu2022bond}  &2022 &   & $\bullet$ &   &&  & &  & $\bullet$& $\bullet$ & $\bullet$\\
 % Kim  et al.\cite{kim2022graph}&2022 &  & $\bullet$ & $\bullet$ & $\bullet$   & $\bullet$&  &&& &\\
  Tang et al.~\cite{tang2023gadbench}&2023  &  &$\bullet$  & &   &&  & & $\bullet$ & $\bullet$& $\bullet$\\
  Liu et al. \cite{liu2023survey}  & 2023 &  &$\bullet$ &$\bullet$ & $\bullet$& $\bullet$ & & & &$\bullet$& \\ \hline
 \textbf{Ours} &2024 & & $\bullet$ &$\bullet$ & $\bullet$ &$\bullet$ & $\bullet$ &$\bullet$ & $\bullet$ & $\bullet$& $\bullet$\\

\bottomrule[1pt]

\end{tabular}
}

\end{table*}
 
In summary, our major contributions are as follows:
\begin{itemize}
    \item The survey provides important insights into the problem complexities and the resulting challenges that are unique for the task of GAD (Sec. \ref{sec:problem}).
    \item We introduce a novel taxonomy of current deep GAD methods, which offers in-depth understanding of the methods from three technical design perspectives, including GNN backbone design, proxy task design, and graph anomaly measures (Sec. \ref{sec:categorization}).
    \item We then introduce 13 fine-grained method categories under these three perspectives to provide more in-depth insights into the model designs (\ie, key intuition, assumption, learning objectives, advantages and disadvantages) and their capabilities in addressing the unique challenges in GAD (Secs. \ref{ref:backbone}, \ref{ref:proxy}, and \ref{ref:measure}). 
    \item  We further discuss multiple important future research directions that involve largely unsolved open problems in GAD. Solutions in these directions would open up new opportunities for addressing the unique challenges in GAD (Sec. \ref{sec:opportunity}).
    \item  We also summarize a large number of representative deep GAD methods from the 13 categories and a large collection of GAD benchmark datasets, and further provide quantitative comparison results on these datasets (Appendices A, B, and C).
\end{itemize}

\section{Problems and Challenges in GAD}\label{sec:problem}
This section discusses some unique complexities and challenges in GAD.

\subsection{Major Problem Complexities}
The complexities in GAD can be summarized in two ways. One source of the complexities lies in some inherent characteristics of graph data.

\begin{itemize}
 \item \textbf{P1. Structural dependency.}  The samples are typically correlated/connected with each other instead of being independent. The connections are of different semantics, \eg, it could be a purchase relationship in a social network, or a citation relationship in a citation network. The complexity of graph structure is reflected in connectivity patterns, dependency, or influence at different levels of graph data, which play a significant role in defining what is abnormal on graphs \cite{akoglu2015graph}.
 For example, different from i.i.d. data, where the anomalies are independent of the context, the anomalies in graphs often depend on the context of a graph data instance, \eg, the neighboring nodes of a node. Anomalies may be considered as normal in one context but abnormal in another.

    \item  \textbf{P2. Diverse types of graph.} There are many types of graphs in the real world, each serving different purposes and applications.
  Graphs can be categorized into static and dynamic types, depending on whether they change over time. It can also be divided into heterophilic and homophilic graphs according to the type of connection \cite{zhu2020beyond, zheng2022graph}. The definition of anomaly in one type of graph can differ significantly from that in other types of graph. In particular, a graph instance (\eg, node/edge/graph) that is clearly abnormal in a dynamic graph at a specific time step (\ie, a static graph) can demonstrate strong normality when looking at the evolution of the graph; similarly, we can have opposite abnormality of a graph instance in a homophilic graph vs. in a heterophilic graph.
  Dealing with diverse types of graphs requires the GAD methods to adapt its learning strategy based on its unique properties of graphs.

    \item \textbf{P3. Computational complexity in handling large-scale graphs.}
    With the increasing amount of online data, modern applications can include very large-scale graph data with millions/billions of nodes and/or edges \cite{hu2021ogb,huang2022dgraph,hezhe2023truncated,tang2023gadbench}, such as those in web-scale social networks, financial transaction networks, cyber networks, user-product e-commerce networks, and citation networks.
    To identify anomalies using global structural contexts, it is essential to consider the full graph structural information, or a large proportion of the structural relations. The key complexity here is to deal with the time and space complexities when loading such large-scale structural relation data.
\end{itemize}

Another source is from the variety of graph abnormalities.
\begin{itemize}
    \item \textbf{P4. Diverse graph anomaly instances.} In contrast to anomaly detection in other forms of data, anomalies within graph data can arise from different components, such as nodes, edges, sub-graphs, or the entire graph \cite{ma2021comprehensive}. Moreover,  graph anomalies can manifest themselves in diverse ways,  depending on the structure and attribute information of  graph data. This highlights the need for GAD methods to incorporate a range of techniques focused on identifying irregular patterns across nodes, edges, subgraphs, and the entirety of the graph.

   \item  \textbf{P5. Large variation in graph abnormality.} 
   Anomalies in graphs can manifest in different forms, including abnormality in graph attributes, graph structure, or the composition of graph attributes and structure \cite{akoglu2015graph}. Some exemplars include attribute anomalies (\ie, graph instances that are exceptional in a graph attribute set) \cite{pang2021deep,aggarwal2017outlieranalysis}, structural anomalies (\ie, graph instances that connect different communities, forming dense connections with others) \cite{ding2019deep,ma2021comprehensive}, contextual anomalies (\ie, graph instances that have different attribute values compared to other nodes in the same community) \cite{ding2019deep,ma2021comprehensive},
   and local affinity anomalies (\ie, graph instances that demonstrate significantly weak affinity to their connected instances compared to other instances~\cite{hezhe2023truncated,liu2021cola}). Also, graph abnormality may vary from a local context to a global context, \eg, the node attributes in a 1-hop neighborhood vs. that in the full graph. This can lead to a highly complex composition set of graph abnormalities, \ie, abnormality in attributes, structure, or both attributes and structure conditioned on a certain context, having very different definitions to the conventional anomaly types -- point, conditional, and group anomalies \cite{pang2021deep,aggarwal2017outlieranalysis,chandola2009anomaly} -- in general AD.
\end{itemize}

In addition, there are some complexities inherited from general AD but amplified in GAD:
\begin{itemize}
    \item \textbf{P6. Unknowingness of abnormality.} Many abnormalities remain unknown until they actually occur \cite{pang2021deep}, in which no prior knowledge about the abnormality is available for the modeling. Further, one type of abnormality can show very different behaviors to the other types of abnormality in a single graph, \eg, the heterogeneous abnormality in graph attributes vs. that in graph structure, in a node set vs. in an edge or subgraph set, in a 1-hop local structure context vs. in a higher-order structural context. Thus, knowing one or more types of graph anomalies may not be generalize to the other types of anomalies.
    \item \textbf{P7. Data imbalance.} Due to the rare occurrence of anomalies, there is typically a large sample imbalance between normal and anomaly classes \cite{pang2021deep,aggarwal2017outlieranalysis,chandola2009anomaly}. This also applies to GAD, but this complexity is largely amplified in GAD due to potential long-tailed distributions in graph structure/attributes \cite{aggarwal2014, liu2023survey}, in addition to the imbalance in the class sample size. 
    \item \textbf{P8. Abnormality camouflage.} Bad actors may adjust their behaviors to camouflage the abnormality of the anomalous instances, making them difficult to detect using popular AD methods. The anomaly camouflage in GAD refers to the phenomenon where anomalous graph instances disguise themselves as normal within a local neighborhood or the global graph.  This may be done through various mechanisms, \eg, attribute manipulation and structural manipulation in a graph \cite{dou2020enhancing, liu2020alleviating, qiao2024generative, yu2024barely}.
   
\end{itemize}

\subsection{Major Challenges}\label{subsec:challenges}
The aforementioned problem complexities lead to the following largely unsolved challenges in GAD, which deep GAD approaches can tackle to various extent: 

\begin{itemize}
    \item \textbf{C1. Graph structure-aware GAD.}
     As discuss in \textbf{P1},  graph anomalies are not solely determined by their own attributes but also by their structural context. Thus, the GAD methods  are required to effectively capture those structural dependency in their anomaly scoring functions. The effect of this dependency in anomaly scoring may vary significantly from homophilic graphs to heterophilic graphs, and from static graphs to dynamic graphs (\textbf{P2}). On the other hand, \textbf{oversmoothing} is a common issue when modeling the graph structure information, which is referred to as a phenomenon in graph representation learning where the learned representations of different nodes become overly similar due to the iterative aggregation of representations of neighboring nodes to obtain the representations of the target nodes. In GAD, this can lead to node/subgraph representations that smooth out anomalies as well, making them indistinguishable between normal and abnormal graph instances \cite{dou2020enhancing,hezhe2023truncated}. Therefore, it is challenging to model diverse structural influences in the anomaly scoring on graphs, while avoiding adverse effects like representation oversmoothing.

    \item \textbf{C2. GAD at scale.}
    As discussed in \textbf{P}3, large-scale graphs with millions or even billions of nodes/edges presents a significant computational challenge to GAD methods that aim to model global or higher-order structure information \cite{tang2023gadbench}.  Existing large-graph modeling methods are challenging to apply directly to GAD due to the extreme imbalance in the data (\textbf{P7}). While some subgraph and sampling techniques have been proposed to address this issue, they often fail to capture the full structural information, resulting in sub-optimal performance \cite{liu2021cola, dou2020enhancing}, particularly for unsupervised GAD. Consequently, performing anomaly detection on large-scale graphs remains a long-standing challenge in the area.

    \item \textbf{C3. Generalization to different graph anomalies.} 
     As discussed in \textbf{P4}, there are various types of graph anomaly instances, making it hard to apply a one-for-all approach.  Achieving this requires a combination of robust feature extraction and versatile detection models. Further, the anomalies can manifest in various forms, ranging from attributes, structure and their composition (\textbf{P5}). However, most existing methods are designed for a specific type of anomaly in an unsupervised manner, which typically have a low recall rate \cite{ding2019deep,liu2021cola,hezhe2023truncated}. The challenge is amplified when the training data does not illustrate every possible class of anomaly (\textbf{P6}), regardless of unsupervised or supervised methods \cite{wang2023open,pang2021toward,pang2023deep,zhu2024anomaly,acsintoae2022ubnormal,ding2022catching,pang2019deep}.

  \item  \textbf{C4. Balanced GAD.} 
As discussed in \textbf{P}7,   since the number of normal instances is significantly larger than that of abnormal instances, the models tend to bias towards the majority class during the training, \ie, they perceive the normal patterns more frequently. Consequently, the models might be overly specialized in recognizing normal instances while generalizing poorly on the anomalies. Often the detection decision thresholds are crucial for making predictions for some methods~\cite{dou2020enhancing, tang2023gadbench}, and poorly chosen thresholds can worsen the effects of data imbalance. Thus, the challenge is to avoid biased GAD.

  \item  \textbf{C5. Robust and interpretable GAD.}  
GAD in real applications needs to be robust against various adverse conditions, such as abnormality camouflage (\textbf{P8}) \cite{dou2020enhancing, gong2023beyond,tang2023gadbench} and unknown anomaly contamination \cite{qiao2024generative,hezhe2023truncated}.
Addressing the abnormality camouflage or anomaly contamination may require models that can capture subtle differences between normal graph instances and camouflaged instances, and complex relationships within the graph as well.
  Besides, an explanation of why a graph instance is detected as an anomaly can
  be crucial for the utility of the predictions in real applications \cite{liu2023towards,sanchez2020evaluating, sanchez2020evaluating}, but it is a largely unexplored area. For example, in bank fraud detection, it is essential to provide a comprehensive explanation of the detected fraudulent activity for facilitating the subsequent investigation, but it is challenging to link the fraud to specific attributes of particular transactions (nodes) and their relations (edges) at a specific time period.
    
\end{itemize}

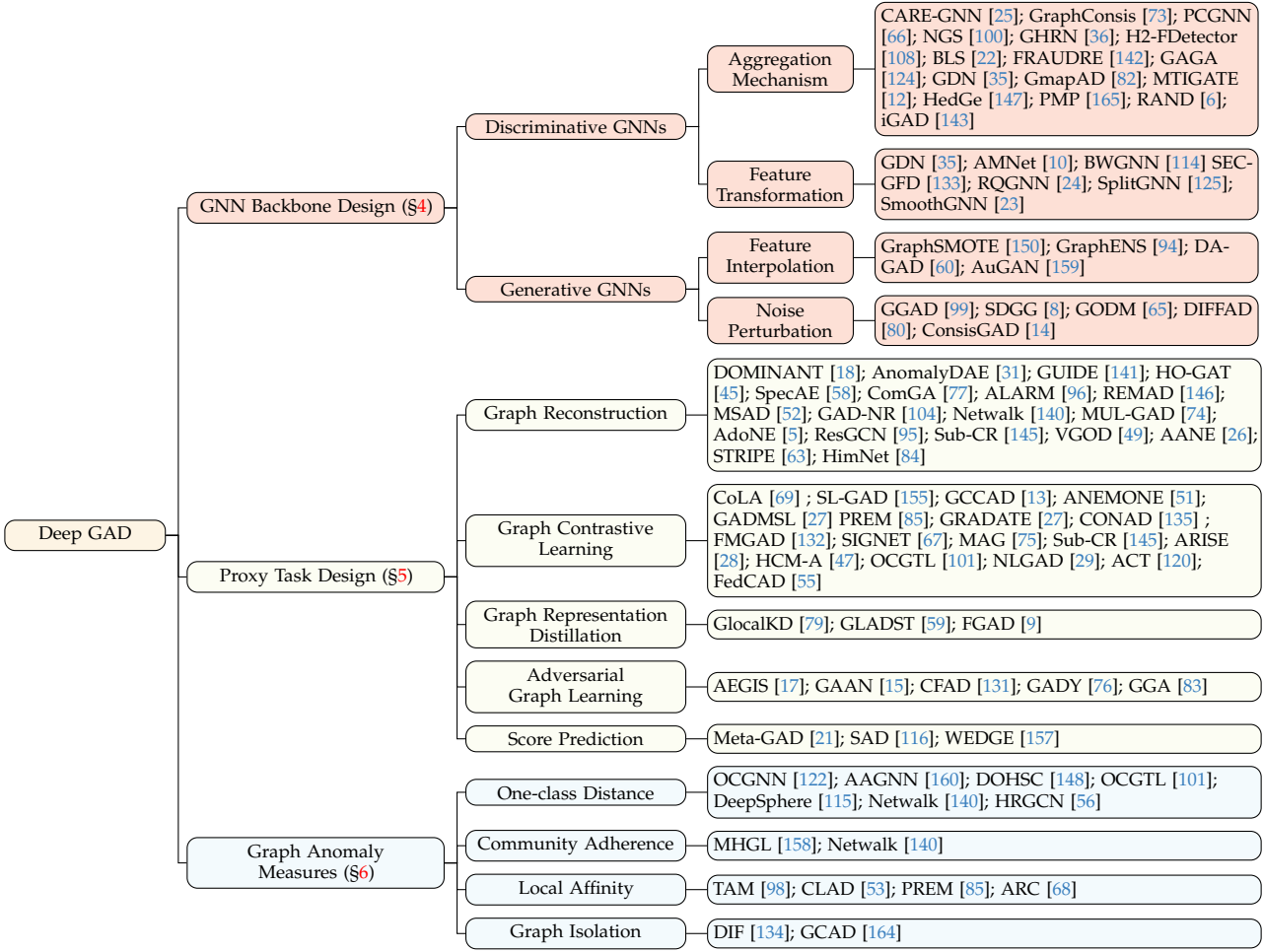
\begin{figure*}[!t]
	\scriptsize
	\begin{forest}
		for tree={
			forked edges,
			grow'=0,
			draw,
			rounded corners,
			node options={align=center},
			calign=edge midpoint,
		},
	    [Deep GAD, text width=2cm, for tree={fill=hidden-grey!20}
     [GNN Backbone Design (\S \ref{ref:backbone}), text width=3.3cm, for tree={fill=hidden-orange!20}
                [Discriminative GNNs, text width=2.8cm, for tree={fill=hidden-orange!20}
				 [Aggregation Mechanism, text width=1.8cm, for tree={fill=hidden-orange!20}
                        [ CARE-GNN \cite{dou2020enhancing}; GraphConsis \cite{liu2020alleviating}; 
                      PCGNN \cite{liu2021pick};  NGS \cite{qin2022explainable}; GHRN \cite{gao2023addressing}; H2-FDetector \cite{shi2022h2}; BLS \cite{dong2022bi}; FRAUDRE \cite{zhang2021fraudre}; GAGA \cite{wang2023label}; GDN \cite{gao2023alleviating}; GmapAD \cite{ma2023towards}; MTIGATE \cite{chang2024multitask}; HedGe \cite{zhang2024generation}; PMP \cite{zhuo2023partitioning}; RAND \cite{bei2023reinforcement}; iGAD\cite{zhang2022dual}; GLHAD \cite{guo2024graph}, text width=5.0cm, node options={align=left}]
                        ]
                        [Feature Transformation, text width=1.8cm, for tree={fill=hidden-orange!20}
                           [ GDN \cite{gao2023alleviating}; AMNet \cite{chai2022can}; BWGNN \cite{tang2022rethinking}  SEC-GFD \cite{xu2023revisiting}; RQGNN \cite{dong2023rayleigh}; SplitGNN \cite{wu2023splitgnn}; SmoothGNN \cite{dong2024smoothgnn} , text width=5.0cm, node options={align=left} ]
                         ]              
				]
	         [Generative GNNs, text width=2.8cm, for tree={fill=hidden-orange!20}
				 [Feature \\Interpolation, text width=1.8cm, for tree={fill=hidden-orange!20}
                            [GraphSMOTE \cite{zhao2021graphsmote}; GraphENS \cite{park2021graphens}; DAGAD \cite{liu2022dagad}; AuGAN \cite{zhou2023improving}; gADAM \cite{zhou2024graph}, text width=5.0cm, node options={align=left} ]
                        ]
                             [Noise \\Perturbation, text width=1.8cm, for tree={fill=hidden-orange!20}
                          [ GGAD \cite{qiao2024generative};  SDGG \cite{cai2023self};   GODM \cite{liu2023data}; DIFFAD \cite{ma2023new}; ConsisGAD \cite{chen2023consistency} , text width=5.0cm, node options={align=left} ]
                         ]
				]
			]
			[Proxy Task Design (\S \ref{ref:proxy}), text width=3.3cm, for tree={fill=hidden-yellow!20}
					[Graph Reconstruction, text width=2.8cm, for tree={fill=hidden-yellow!20}
                     [ DOMINANT \cite{ding2019deep};  AnomalyDAE \cite{fan2020anomalydae};  GUIDE 
                        \cite{yuan2021higher};  HO-GAT \cite{huang2021hybrid}; SpecAE \cite{li2019specae}; ComGA \cite{luo2022comga}; ALARM \cite{peng2020deep}; REMAD \cite{zhang2019robust};  MSAD \cite{kim2023deep};  GAD-NR \cite{roy2024gad};  Netwalk \cite{yu2018netwalk}; MUL-GAD \cite{liu2022mul}; AdoNE \cite{bandyopadhyay2020outlier}; ResGCN \cite{pei2022resgcn}; Sub-CR \cite{zhang2022reconstruction}; VGOD \cite{huang2022we};
	                 	AANE \cite{duan2020aane}; STRIPE \cite{liu2024spatial}; HimNet \cite{niu2023graph}; SI-HGAD \cite{zou2024structural}, 
					text width=7.3cm, node options={align=left}
					]                                       
                    ]                                    
				[Graph Contrastive Learning, text width=2.8cm, for tree={fill=hidden-yellow!20}
                      [CoLA \cite{liu2021cola}
                      ;  SL-GAD \cite{zheng2021generative}; 
                         GCCAD \cite{chen2022gccad};  ANEMONE \cite{jin2021anemone}; GADMSL \cite{duan2023graph}  PREM \cite{pan2023prem}; GRADATE \cite{duan2023graph}; CONAD \cite{xu2022contrastive} ; FMGAD \cite{xu2023few}; SIGNET \cite{liu2024towards}; MAG \cite{liu2023revisiting}; Sub-CR \cite{zhang2022reconstruction}; ARISE \cite{duan2023arise}; HCM-A \cite{huang2022hop}; OCGTL \cite{qiu2022raising}; NLGAD \cite{duan2023normality}; ACT \cite{wang2023cross}; FedCAD \cite{kong2024federated},
				    text width=7.3cm, node options={align=left}
				    ]                                        
                    ]
                    [Graph Representation Distillation, text width=2.8cm, for tree={fill=hidden-yellow!20} 
                    [GlocalKD \cite{ma2022deep}; GLADST \cite{lin2023discriminative}; FGAD \cite{cai2024fgad},
					text width=7.3cm, node options={align=left}
					]                    
                   ]  
                 [Adversarial Graph Learning, text width=2.8cm, for tree={fill=hidden-yellow!20}                 
                  		[AEGIS \cite{ding2021inductive}; GAAN \cite{chen2020generative}; CFAD \cite{xiao2023counterfactual}; GADY \cite{lou2023gady}; GGA \cite{meng2023generative}
					,
					text width=7.3cm, node options={align=left}
					]                   
                   ] 
                 [Score Prediction, text width=2.8cm, for tree={fill=hidden-yellow!20}                 
                  [Meta-GAD \cite{ding2021few}; SAD \cite{tian2023sad}; WEDGE \cite{zhou2023learning}
					,
					text width=7.3cm, node options={align=left}
					]                   
                   ]  
			]
                [Graph Anomaly Measures (\S \ref{ref:measure}), text width=3.3cm, for tree={fill=hidden-blue!20}
                [One-class Distance, text width=2.8cm, for tree={fill=hidden-blue!20}
					  [OCGNN \cite{wang2021one}; 
                         AAGNN \cite{zhou2021subtractive}; DOHSC \cite{zhang2023deep}; OCGTL \cite{qiu2022raising}; DeepSphere \cite{teng2018deep}; Netwalk \cite{yu2018netwalk}; HRGCN \cite{li2023hrgcn},
				    text width=7.3cm, node options={align=left}
				    ]
                ]
                     [Community Adherence, text width=2.8cm, for tree={fill=hidden-blue!20}
					    [ MHGL \cite{zhou2022unseen};  Netwalk \cite{yu2018netwalk},
				    text width=7.3cm, node options={align=left}
				    ]
                ]
                    [Local Affinity, text width=2.8cm, for tree={fill=hidden-blue!20}
						  [TAM \cite{hezhe2023truncated}; CLAD 
                         \cite{kim2023class}; PREM\cite{pan2023prem}; ARC \cite{liu2024arc}; UNPrompt \cite{niu2024zero}; AnomalyGFM \cite{qiao2025anomalygfm},
				    text width=7.3cm, node options={align=left}
				    ]
                ]
                     [Graph Isolation, text width=2.8cm, for tree={fill=hidden-blue!20}
				    [ DIF \cite{xu2023deep}; GCAD \cite{zhuang2023subgraph},
				    text width=7.3cm, node options={align=left}
				    ] 
                ]
			]
	]
 	\end{forest}
	\caption{Overview of the proposed taxonomy of deep GAD from three high-level and 13 fine-grained technical perspectives. 
 }
    \label{fig:taxonomy}
\end{figure*}

\section{Categorization of Deep GAD}\label{sec:categorization}
\subsection{Preliminaries}
GAD aims to recognize the anomaly instances in graph data that may vary from nodes, edges to subgraphs by learning an anomaly scoring function. Traditional GAD methods achieve anomaly detection using matrix decomposition and residual analysis \cite{akoglu2015graph}. However,  their performance is often bottlenecked due to the lack of representation power to capture
the rich structure-attribute semantics of the graph data and to handle high-dimensional node attributes. 
In recent years, GNNs have been widely used in GAD due to their powerful representation learning ability. Some representative GNNs like GCN \cite{kipf2016semi}, GraphSage \cite{hamilton2017inductive}, and GCL\cite{you2020graph} attract much attention in node representation learning in graphs. These GNNs can be leveraged to learn the expressive representation of different graph instances for GAD. 

\textbf{Definition and Notation.}
In this section, we introduce the definitions and notations used throughout the paper.  We denote a graph by $G = (\mathcal{V},\mathcal{E})$ where $\mathcal{V}$ and ${\mathcal{E}}$ denote the node set and edge set respectively. For the graph $G$, we use $\mathbf{X} \in \mathbb{R}^{N \times M}$ to denote the matrix of node attributes and $\mathbf{x}_i \in \mathbb{R}^M$ is the attribute vector of $v_i \in \mathcal{V}$, and ${\bf{A}} \in\{0,1\}^{N \times N}$ is the adjacency matrix of $G$ with ${{\bf{A}}_{ij} = 1}$ iff $\left(v_i, v_j\right) \in \mathcal{E}$, where $N$ is the number of node.  

\textbf{Problem Statement.}
GAD can be divided into anomaly detection at the node-level, edge-level, sub-graph level and graph-level settings. The node-, edge- and subgraph-level AD tasks are typically performed within a single large graph $G$, where the input samples are nodes $v \in G$, edges $e \in G$, and subgraphs $s\subset G$, respectively. For the graph-level AD task, the input samples are a set of graphs $\mathcal{G}=\{G_1,G_2,\cdots\}$. 
For the sake of simplicity and generality across different levels of GAD, we uniformly denote the input samples as $o$, \ie, $o$ can denote a node $v$, an edge $e$, a subgraph $s$, or a full graph $G$, depending on their use in specific algorithms or models. 
% For these three tasks, they are node, graph, and edge. 
Then GAD aims to learn an anomaly scoring function $f: \{o_1, o_2,\cdots\} \rightarrow \mathbb{R}$, such that $f(o) < f(o^{\prime})$, $\forall o\in \mathcal{O}_{n}, o^{\prime} \in \mathcal{O}_{a}$, where ${{\mathcal O}_n}$ and ${\mathcal O}_{a}$ denote the set of normal and abnormal graph instances, respectively. Since anomalies are rare samples, it is typically assumed that $\left| {\mathcal O}_n \right| \gg \left| {\mathcal O}_{a} \right|$.

\subsection{Categorization of Deep GAD Methods}

In order to facilitate a comprehensive understanding of the research progress in GAD, we introduce a new taxonomy that categorizes current GAD methods into three main groups,including GNN backbone design, proxy GAD task design, and graph anomaly measures, depending on the insights offered by each method. This enables us to review the GAD methods from three different technical perspectives. To elaborate the insights in each perspective, we further categorize the methods into fine-grained 13 groups. An overview of the taxonomy is shown in Figure \ref{fig:taxonomy}.

More specifically, general GNNs can not be directly applied to GAD due to the aforementioned problem complexities, and thus, there is a group of studies that focus on designing suitable GNN backbones for GAD.
The design of the GNN backbones can be divided into 
discriminative GNNs and generative GNNs according to the improvement of different modules in GNNs.
The second main category of methods is on the GAD models constructed by optimizing a diverse set of well-crafted learning
objective functions to form a proxy task that can guide the GAD models to capture diverse graph anomaly/normal patterns without the
need for ground-truth labels. This category of methods can be further divided into five subcategories based on the modeling in the proxy tasks.  Lastly, there is a group of methods that build GAD models based on anomaly measures that are designed specifically for graph data. These methods can be further grouped into four subcategories depending on the type of graph anomaly measures used. A summarization of representative algorithms for each type of GAD approaches is presented in \textbf{Table 1} in \texttt{Appendix A}.

\section{GNN Backbone Design} \label{ref:backbone}
This category of methods aims at leveraging GNNs to learn effective representations of graph instances for downstream anomaly detection tasks.  Due to its strong capability to represent graph-structured data, GNNs can effectively obtain expressive node representations through aggregation among the connected nodes. However, unlike general node/graph classification datasets, GAD datasets are often extremely class-imbalanced, which prevents GNNs from being directly applied to GAD datasets. Therefore, several GNNs have been proposed to handle the imbalance problem for more effective GAD. Concretely, this type of methods can be roughly divided into discriminative- and generative-based GNNs for GAD.

\subsection{Discriminative GNNs}
\begin{figure}
 \centering
 \includegraphics[width=3.60in,height=1.05in]{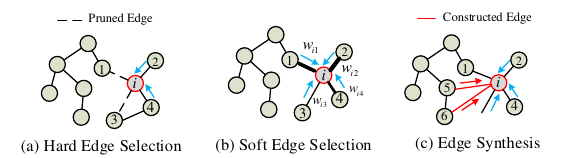 }\\
 \caption{Three categories of aggregation mechanism.
 }
 \vspace{-1em} 
 \label{fig:part1}
\end{figure}

\begin{figure}
 \centering
 \includegraphics[width=3.30in,height=1.15in]{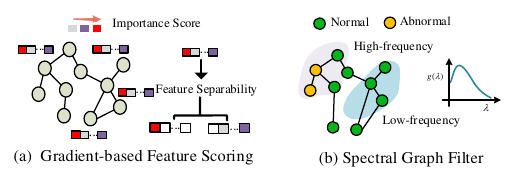 }\\
 \caption{Two categories of feature transformation.
 }
 \vspace{-1em} 
 \label{fig:part1_2}
\end{figure}
Discriminative GNN-based GAD methods refer to a GNN architecture specifically designed for discriminating normal graph instances from the abnormal ones, where the discrimination is typically achieved through a supervised learning manner. Thus, the discriminative GNNs are typically trained on the labeled graph dataset containing examples of both normal and abnormal instances. The core idea in these methods is to adapt the conventional GNN backbones in a way so that the message passing in GNNs can capture the majority patterns or the deviation patterns better.

Let ${\bf{h}}_i$ be the feature representation of a graph instance $o_i$ that is obtained through $l$ layers of \textbf{feature aggregation (FAG)} in a GNN, \ie, ${\bf{h}}_i=\text{FAG}_{\mathit{1:l}}(o_i, \mathcal{N}_i)$ where $\mathcal{N}_i$ represents the neighbor set of a graph instance $o_i$, these methods are typically optimized via a general cross-entropy loss to train the discriminative GAD model. 
\begin{equation}
{L_{{\rm{cls}}}} =  - \sum\limits_{o_i \in G} {\left[ {{y_i}\log {p_i} + (1 - {y_i})\log (1 - {p_i})} \right]},
\end{equation}
where $y_i$ denotes the class label of the instance $o_i$ and $p_i=\text{MLP}_{\mathit{1:m}}({\bf{h}}_i)$ is the output of a mapping function that goes through $m$ layers of multiple perceptrons (MLP) to project the feature $h_i$ to a probability of the sample being abnormal/normal. 
During inference, given a graph instance, $o_j$, $p_j$ or its inverse $1-p_j$ can be used as its anomaly score.

Different GNN-based encoders can be used to obtain the feature representation $\bf{h}$, such as graph convolutional networks (GCNs) \cite{kipf2016semi}, graph attention networks (GAT) \cite{velivckovic2017graph}, or GraphSage \cite{hamilton2017inductive}. 
Depending on which part of the learning pipeline is focused, we group the existing methods in this category into two sub-categories, including the methods that focus on the GNN neighborhood aggregation design, \ie, $\text{FAG}_{\mathit{1:l}}(o_i, \mathcal{N}_i)$, and that focus on feature transformation with the raw attributes or node embeddings as input, \ie, $\textit{Transformation}(\mathbf{h}_i)$ or  $\textit{Transformation}(\mathbf{x}_i)$. Below we introduce each of them in detail.

\subsubsection{Aggregation Mechanism}
As a simple and effective way to obtain the representation of nodes in GNNs, feature aggregation plays a crucial role in learning node representations by aggregating information from neighboring nodes in a graph. Thus, to create GNN-based methods for GAD, one principled approach is to craft suitable feature aggregation FAG designs that are sensible for graph anomaly instances. 

\textbf{Assumption.}  The GAD methods in this line assume that the connected instances from the same class in graphs have similar characteristics, from which we can perform feature aggregation to obtain discriminative normality/abnormality patterns.

A widely-used FAG mechanism is as follows \cite{kipf2016semi}:
\begin{equation} \label{eq:FAG}
{\bf{h}}_i^{(l)}=\sigma\left({\bf{W}}^{(l)}\left({\bf{h}}_i^{(l-1)} + \text{AGG}\left(\left\{{\bf{h}}_j^{(l-1)} \mid o_j \in \mathcal{N}_i \right\}\right)\right)\right),
\end{equation}

\noindent where ${\bf{h}}_i^{(l)}$ is the feature representation of instance $i$ in the $l$-th layer, $\sigma$  is an activation function, and ${\bf{W}}^{(l)}$ is the training parameters in the $l$-th layer. The graph instance $o_i$ is often set as a node $v_i$, and  $\mathcal{N}_i$ is typically the 1-hop neighborhood of the node $v_i$. $\text{AGG}(\cdot)$ is implemented using sum or mean aggregation, which sums or averages the representations of neighbor features. More advanced aggregation methods, such as attention-based aggregation \cite{fan2020anomalydae} and LSTM \cite{zheng2019one}, are also widely used.
However, due to the oversmoothing representation issue (\textbf{C1} in Sec.~\ref{subsec:challenges})
, directly applying such neighborhood aggregation  
mechanism 
can largely reduce the discriminability of of graph anomalies, especially for those whose abnormal behaviors are subtle (\textbf{C3} in Sec.~\ref{subsec:challenges}).
Therefore, a variety of methods were proposed to enforce distinguishable
representations for normal and abnormal graph instances 
throughout a number of feature aggregation iterations. 
These methods can be summarized via the following principled framework: 

\begin{equation}
\begin{array}{l}
{\bf{\hat h}}_i^{(l)} = \text{AGG}\left( {\left\{ {{\bf{h}}_j^{(l - 1)}\mid {o_j} \in \Phi \left( {{{\cal N}_i}} \right) \cup \Psi ({\cal V})} \right\}} \right),\\
{\bf{h}}_i^{(l)} = \sigma \left( {{{\bf{W}}^{(l)}}\left( {{\bf{h}}_i^{(l - 1)} + {\bf{\hat h}}_i^{(l)}} \right)} \right),
\end{array}
\end{equation}
where $\Phi(\cdot)$ and $\Psi(\cdot)$ represent a filtering function on the neighborhood set $\mathcal N$ and an edge synthesizer on the full node set $\mathcal V$, respectively.
Depending on how the methods specify the $\Phi$ or $\Psi$ function, we further categorize them into two fine-grained groups -- hard/soft edge selection and edge synthesis -- to gain better insights into these existing methods.

\vspace{1mm}
\noindent \texttt{$\bullet$ Hard Edge Selection.} 
Popular aggregation mechanisms in GNN methods are built upon a \textbf{homophily assumption} that connected nodes come from the same class.
  
Thus, the existence of \textit{non-homophily edges} (\ie, edges that connect nodes of different classes, also referred to as \textit{heterophily edges} below) in a GAD dataset can greatly hinders the discriminability of the learned feature representations. 
As shown in Fig. \ref{fig:part1}(a), one popular strategy is to instantiate $\Phi\left(\mathcal{N}_i\right)$ that prune the heterophily edges w.r.t. the normal class, referred to as hard edge selection. Below we review the methods in this line.

In order to enhance the homophily relations,  CARE-GNN \cite{dou2020enhancing} devises a label-aware similarity measure to find informative neighboring nodes during the aggregation where $\Phi\left(\mathcal{N}_i\right)$ is instantiated by a node selector that chooses the neighbors with high similarity. Moreover, a reinforcement learning module is also used in \cite{dou2020enhancing} to find the optimal amounts of neighbors to be selected.
MITIGATE~\cite{chang2024multitask} implements $\Phi\left(\mathcal{N}_i\right)$ via a masked aggregation mechanism that utilizes the distance-based clustering algorithm to choose a subset of high-representative nodes, in which the nodes that are closest to the cluster centers are chosen. 
GmpaAD \cite{ma2023towards} takes a similar clustering-based approach as MITIGATE, but it uses 
a differential evolutionary algorithm to find the optimal mapping strategy and generate the representative nodes given the selected candidates from a clustering method.
On the other hand, H2-FDetector \cite{shi2022h2} categorizes the edges into homophily and heterophily connections in the graph, and 
further designs a new information aggregation strategy to ensure that the homophily connections propagate similar information while the heterophily connections propagate different information.

In addition to using distance, $\Phi$ can also be specified via meta learning or reinforcement learning. BLS \cite{dong2022bi} is the representative method that enhances the FAG mechanism under imbalanced and noisy scenarios by selecting important nodes via a meta-learning gradient of the learning loss.  
AO-GNN \cite{huang2022auc} employs a reinforcement learning method supervised by a surrogate reward based on AUC performance to prune the heterophily edges.  
NGS \cite{qin2022explainable} 
takes a meta-graph learning approach that devises a differentiable neural architecture to determine a set of optimized message passing structures and then combines multiple searched meta-graphs in FAG.

\vspace{1mm}
\noindent \texttt{$\bullet$ Soft Edge Selection.}
 Another research line is adopting an attention mechanism in GNNs by assigning the weights for each edge for soft edge selection for GAD, rather than hard edge selection, as demonstrated in Fig. \ref{fig:part1}(c). 
 This weight is generally obtained through the relationship between node embeddings, which serves as an effective way to enforce the importance of some specific edge relations in the feature aggregation. GAT 
 \cite{velivckovic2017graph} is widely used as the basic backbone, on top of which a variety of designs is introduced in the methods of this category for GAD. Specifically, the general attention mechanism in GAT can be formulated as: 
\begin{equation}
{\bf{h}}_i^{(l)} = \sigma \left( {\sum\limits_{{o_j} \in {\cal N}(i)} \Phi  \left( {{{\bf{h}}_i},{{\bf{h}}_j};\theta } \right){\bf{Wh}}_j^{(l)}} \right),
\end{equation}
where $\Phi$ indicates a weight learning function with parameters $\theta$ applied on the embedding of a graph instance $o_i$ and its neighbors $o_j$. It represents the contribution of relations/neighbors  to the target instance $o_i$, where the instance $o_i$ is often specified as a node $v_i$.
For instantiating $\Phi$, GraphConsis \cite{liu2020alleviating} reveals an inconsistency phenomenon in node connections that abnormal nodes can have a high likelihood of being connected to normal nodes to camouflage their abnormality.
It then introduces a consistency scoring-based method based on node embedding similarities
and a self-attention mechanism to assign weights for different connections in the aggregation in $\Phi$. 
FRAUDRE \cite{zhang2021fraudre} extends the inconsistency-based scoring method to include three types of graph inconsistencies in features, topology, and structural relations to consider the importance of different connections. 
On the other hand, PMP \cite{zhuo2023partitioning} introduces a partitioning message passing to independently handle the heterophily and homophily neighbors preventing the gradients in the optimization from being dominated by normal nodes. To achieve this, $\Phi$ is implemented by a weight generator function to adaptively adjust the influence of neighbors from different classes to the target node. 

\vspace{1mm}
\noindent\texttt{$\bullet$ Edge Synthesis.} The edge selection function $\Phi$ focuses on local structure only. Edge synthesis function $\Psi$ can often be used to complement $\Phi$ in capturing more global patterns for GAD, as demonstrated in Fig. \ref{fig:part1}(b). For example, GHRN \cite{gao2023addressing} 
prunes the inter-class edges by emphasizing and delineating the high-frequency components of the graph. 
Apart from edge reduction, the indirect link between nodes can also be beneficial for the normality representation learning \cite{gao2023addressing}. To this end, GHRN introduces
a global node selector $\Psi(\mathcal{V})$ that chooses nodes beyond the neighboring nodes of the target node to introduce this information into the feature aggregation, which can be seen as an edge synthesizer that connects distant nodes to a target node. 
PCGNN \cite{liu2021pick} specifies $\Psi(\mathcal{V})$ using a subgraph construction method consisting of 
class-label-balanced node and edge samplers to tackle potential issues arising from the skewed class distribution. To this end, it incorporates the label information into the sampling process to choose nodes and edges for its sub-graph construction. A distance function is also used to simplify the neighborhood in the sub-graphs. 
NSReg \cite{wang2023open} leverages a novel normal structure regularization method where the normal-node-oriented relation is used to enforce strong normality into the representation learning to avoid overfitting to the labeled abnormal nodes.
 
The relation representations are generated through a learnable transformation that fuses the representations of relevant nodes, which are subsequently used to optimize the normal-node-oriented relation prediction and the representation learner.

\textbf{Advantages.} The key advantages of discriminative GNN-based GAD methods are as follows. (i) Treating anomaly detection as an end-to-end imbalanced classification 
task simplifies the GAD problem, allowing the use of available abnormal samples to detect known graph anomaly instances. (ii) This approach does not rely on the reconstruction of the graph structure, which significantly reduces memory usage and enhances its scalability for large-scale graphs.

\textbf{Disadvantages.} They also have some major disadvantages. (i) Since these methods require some labeled graph data as supervision information, they become inapplicable or less effective in practical applications where such data is difficult to obtain.
(ii) Edge selection may lead to the loss of important structure information while edge synthesis may introduce noise or some irrelevant structure information into the message passing in GNNs. 

\textbf{Challenges Addressed.} The tailored GNNs for GAD enable effective modeling of graph structure and its interaction with graph attribute information (\textbf{C1}). They may also be able to learn more discriminative features for detecting subtle anomalies that are similar to labeled abnormal samples (\textbf{C3}). Since these methods require local feature aggregation, they can often scale up to large graphs (\textbf{C2}).

\subsubsection{Feature Transformation} 
In addition to the efforts on the aggregation mechanism,
another popular approach to obtain discriminative features for GAD is to perform feature transformation on either the graph instance representations from GNNs, \ie, $\mathit{Transformation}(\mathbf{h}_i)$, or raw attributes, \ie, $\mathit{Transformation}(\mathbf{x}_i)$.
This is crucial since datasets used in GAD can often contain a substantial proportion of feature information that is irrelevant, or even noisy, to GAD. There are two popular approaches to instantiate this $\mathit{Transformation}(\cdot)$ function: one is to use gradient information and another is to use spectral graph filters.

\textbf{Assumption.} It is assumed that there is irrelevant or noisy information in the raw attributes or graph structure w.r.t. GAD, which should be discarded during feature aggregation. 

\vspace{1mm}
\noindent \texttt{$\bullet$ Gradient-based Feature Scoring.} 
Extracting class-related features specific to the characteristics of a particular class is one straightforward way to obtain discriminative features.
Inspired by variable decomposition \cite{fan2022debiased}, gradient information-based methods have been emerging as one main approach to
obtain discriminative representations from GNNs \cite{ma2019disentangled}.  
 The key idea here is to select features from the representation ${\bf{h}}_i$ based on the gradient backpropagated from the softmax probability of being a specific class, as illustrated in Fig. \ref{fig:part1_2}(a). Let ${\alpha_k^{c}}$ be the gradient score of an anomaly class $c$ w.r.t. a feature $k$, then it represents the contribution of this feature to anomaly detection, which can be formulated as follows:
\begin{equation}
    \alpha_k^{c}=\frac{1}{N}\left|\sum_{i=1}^N \frac{\partial y^{c}}{\partial \mathbf{h}_{k,i}}\right|,
\end{equation}
where $y^{c}$ is the predicted probability of being the anomaly class $c$ and $\mathbf{h}_{k,i}$ is the $k$-th feature of the representation of node $i$ from a hidden layer. After obtaining a gradient score for each feature dimension, the top $K$ features with the largest gradient scores are selected to represent the nodes in a reduced feature space.
This gradient score-based approach is used in GDN \cite{gao2023alleviating} to select abnormal and normal features in a supervised manner, and these features are found often to be
invariant to structure distribution shift. It 
helps reduce the negative influence of irrelevant features while preserving the extracted abnormal/normal graph patterns, thus enhancing the overall performance of GAD.
Similarly, GraphENS \cite{park2021graphens} determines the importance of each node feature via a gradient score-based method. Apart from gradient score,  
existing feature selection methods for anomaly detection or imbalanced classification \cite{pang2016unsupervised,pang2017learning,pang2017selective,pang2018sparse,li2017feature} may be adapted for GAD.

\vspace{1mm}
\noindent \texttt{$\bullet$ Spectral Graph Filter.}
Spectral graph filter, which combines the strengths of spectral graph theory and GNNs, is widely applied to capture and analyze the structural properties of graphs for GAD tasks.
This approach utilizes a set of graph filters to transform the raw attributes to latent space to extract discriminative graph representations,
as illustrated in Fig. \ref{fig:part1_2}(b). Each graph filter assumes that normal nodes tend to have similar features with their neighbors, which can be regarded as low-frequency information, whereas abnormal nodes in the graph are characterized by deviations from the norm, which are often accompanied by high-frequency information since abnormal nodes often have different characteristics from surrounding nodes. The distinction between low-frequency and high-frequency information is closely related to the spectral properties of the graph. In particular, the low-frequency and high-frequency variations on the graph can be effectively captured by the lower and higher eigenvalues of the graph Laplacian matrix, respectively. To leverage this information for GAD, graph Fourier transformation \cite{shuman2013emerging} based graph filtering operation is often used. Formally, let $\mathbf{L}$ be the symmetrically normalized Laplacian, with eigen decomposition $\mathbf{L}=\mathbf{U} \boldsymbol{\Lambda} \mathbf{U}^T$,  where $\boldsymbol{\Lambda}=\operatorname{diag}\left[\lambda_1, \cdots, \lambda_n\right]$,  then a signal $\mathbf{x} \in {\mathbb{R}^n}$ is transferred by using a graph filter $g$, $\mathit{Transformation}(\bf{x}) = \mathit{g} \star \bf{x} = {\bf{U}}\mathit{g}({\bf{\Lambda }}){{\bf{U}}^T}\bf{x} $, and thus, the graph signal  filtered by the $k$-th filter can be defined as 

\begin{equation}
{{\bf{h}}_{i,k}} = {\bf{U}}{g_k}({\bf{\Lambda }}){{\bf{U}}^T}{{\bf{x}}_i} = {\bf{U}}{\mathop{\rm diag}\nolimits} \left[ {{g_k}\left( {{\lambda _1}} \right), \ldots ,{g_k}\left( {{\lambda _n}} \right)} \right]{{\bf{U}}^T}{{\bf{x}}_i}.
\end{equation}

To better capture the feature separability, a set of multi-frequency filters is often employed to learn the node representations \cite{chai2022can}. Accordingly, the final representation ${\bf h}_i$ of node $v_i$ can be obtained by 
% \hh{how we get the 2nd equation from the first one? can we elaborate it a bit?}

\begin{equation}
{{\bf{h}}_i} = \sum\limits_k {{\alpha _{i,k}}} {{\bf{h}}_{i,k}} = {\bf{U}}\sum\limits_k {{\alpha _{i,k}}} {g_k}({\bf{\Lambda }}){{\bf{U}}^T}{{\bf{x}}_i},
\end{equation}
where $\alpha _{i,k}$ is an important score for the graph filter ${g_k}({\bf{\Lambda }})$.

Various spectral graph filters utilizing the frequency at different level have been proposed 
for GAD. Specifically,
AMNet \cite{chai2022can} is an early work that adaptively integrates different graph signals with mixed frequency patterns via a multi-frequency graph filter group.
It uses a restricted Bernstein polynomial parameterization method to approximate filters in multi-frequency groups. BWGNN \cite{tang2022rethinking} reveals a `right-shift' phenomenon in GAD datasets with synthetic or real-world anomalies, \ie, low-frequency energy is gradually
transferred to the high-frequency part when the degree of
anomaly becomes larger. According to this phenomenon, they justify the necessity of spectral localized band-pass filters in GAD. Current GNNs with adaptive filters cannot guarantee to be band-pass or spectral localized.
Therefore, they build on top of Hammond's graph wavelet theory \cite{hammond2011wavelets} to develop a new GNN architecture with a Beta kernel to better detect higher-frequency anomalies. 

The aforementioned filters can be challenged by heterophily graphs since homophily graphs are assumed in these filters. To tackle this challenge, SEC-GFD \cite{xu2023revisiting}  employs a hybrid band-pass filter to partition the graph spectrum into hybrid frequency bands,
while SplitGNN \cite{wu2023splitgnn} combines a band-pass graph filter with a tunable Beta Wavelet GNN  to address the heterophily issue in node representation learning for GAD (\textbf{C1} in Sec.~\ref{subsec:challenges}).

\textbf{Advantages.}  
(i) The feature transformation approach supports the extraction of discriminative features from data with various amount of irrelevant/noisy information. (ii) It can also provide rich and informative graph representations capturing beyond local graph properties, such as connectivity, centrality, and community structure, for more effective GAD. 

\textbf{Disadvantages.} (i) Depending on what criterion or spectral filter is used, this approach might overlook some important information in feature transformation that could be crucial for GAD, since one feature scoring criterion or filter often fails to capture all possible discriminative features.  
(ii) The approach may confined to a specific type of graph. For example, spectral GNNs are primarily designed for homogeneous graphs, and thus, it may not be suitable for a heterogeneous graph with different types of nodes.  

\textbf{Challenges Addressed.} The feature transformation approach enhances GNNs by focusing them on specific types of discriminative features, thereby providing effective solutions to the graph structure-aware anomaly detection problem (\textbf{C1}). Furthermore, this approach focuses on discriminative features and does not require costly graph structure reconstruction, and thus, it is often good for GAD on large-scale graph data (\textbf{C2}).

\subsection{Generative GNNs}\label{subsec:generative_gnn}
Generative GNN-based methods focus on synthesizing new graph instances to augment the existing graph data for enhancing model training on the new graph for GAD. This approach is motivated by the problem complexities like the scarcity of graph anomaly instances and their large variations (\textbf{P5} and \textbf{P7} in Sec. \ref{subsec:challenges}). The underlying key idea in these GAD methods is to synthesize outliers that can simulate graph anomalies in some specific properties to provide pseudo anomaly information for training the GAD models. In general, these methods
can be summarized by the following formulation:
\begin{equation}\label{eqn:generative_gnn}
 {o_i^{new}} = {g_\phi }(o_i,{\bf{X}},{\bf{A}}; \epsilon),   
\end{equation}
where $o_i$ is an existing graph instance, $g_\phi$ is a graph instance generator with parameters $\phi$ that uses $o_i$, existing attribute information $\bf{X}$, graph structure information $\bf{A}$, and some auxiliary information $\epsilon$ to generate the augmented graph instance, $o_i^{new}$. It is then followed by a one-class or binary classification loss ${L_{cls}}$ defined as
\begin{equation}
{L_{gen}} = \sum
% \limits_{{v_i} \in {\cal V} \cup {{\cal V}^{new}}} 
{\ell \left( {Y,{f_\theta }\left( {{{\bf{X}}_i^{new}},{{\bf{A}}_i^{new}}} \right)} \right)}, 
\end{equation}
where ${\bf{X}}^{new}$ and ${\bf{A}}^{new}$ are the augmented version of $\bf{X}$ and $\bf{A}$,  $Y$ represents the label set of the graph instances consisting of pseudo labels of the generated anomaly instances (and the labels in the original graph if any), and $f$ is a GAD model.
Depending on the source of $\epsilon$ in $g_\phi$ in Eq. \ref{eqn:generative_gnn}, these methods can be categorized into feature interpolation and noise perturbation generation methods. The former focuses on specifying $\epsilon$ using data from existing graphs, while the latter focuses on utilizing prior distribution to specify $\epsilon$.
\begin{figure}
 \centering
 \includegraphics[width=3.00in,height=1.25in]{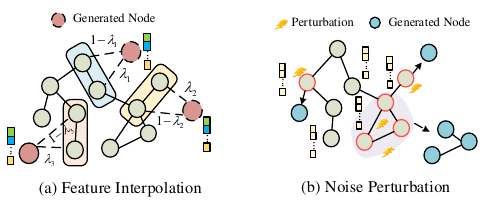 }\\
 \caption{Two categories of generative GNNs.
 }
 \vspace{-1em} 
 \label{fig:part2}
\end{figure}
\subsubsection{Feature Interpolation}
Feature interpolation is a commonly employed technique in imbalance learning for augmenting data, where the representations of synthesized graph instances are created by interpolating the representations of existing graph instances. It has been explored in popular algorithms like SMOTE \cite{fernandez2018smote} and Mixup \cite{zhang2017mixup} to oversample the minority classes or generate diverse, large-scale samples for training deep models. The approach can be generally formulated as follows:
\begin{equation}
    {\mathbf{h}}_{new }=(1-\lambda) \cdot \mathbf{h}_a^{(i)}+\lambda \cdot \mathbf{h}_b^{(j)},
\end{equation}
where $\mathbf{h}_{new}$ is the representation of the generated graph instance using a convex combination of the representations of two existing graph instances from the same class, $\mathbf{h}_a$ and $\mathbf{h}_b$, as illustrated in Fig. \ref{fig:part2}(a). 

\textbf{Assumption.} 
The interpolated feature representations between graph instances can well align with those of the instances from the anomaly class.

GraphSMOTE \cite{zhao2021graphsmote}, GraphMixup \cite{wu2022graphmixup} and GraphENS \cite{park2021graphens} are representative methods that utilize feature interpolation to address the challenge of biased GAD (\textbf{C4} in Sec.~\ref{subsec:challenges}). These methods are designed for imbalanced node classification,  which can be considered as supervised GAD in a closed-set setting since they do not consider the detection of novel/unknown anomaly types that are not illustrated by the labeled training anomaly examples \cite{pang2021toward,pang2023deep,wang2023open,ding2022catching,zhu2024anomaly,acsintoae2022ubnormal}.
In particular, GraphSMOTE \cite{zhao2021graphsmote} extends SMOTE to synthesize new nodes and edges in graph data. Different from generic data, generating new nodes in a graph requires the connections of these nodes to existing nodes. Thus,  
an edge generator is trained simultaneously in GraphSMOTE to model the relations between existing and new nodes when using the SMOTE-style approach to generate the new nodes.  A similar work is done in AugAN \cite{zhou2023improving} that performs interpolation on the feature representations of the nodes that are similar to labeled anomalies to increase the number of training anomaly examples. However, directly performing the interpolation may produce out-of-domain samples due to the extreme sparsity of the minority classes. To alleviate this issue, GraphMixup \cite{park2021graphens} is introduced to construct semantic relation spaces that allow the interpolation to be performed at the semantic level. In GraphENS \cite{park2021graphens}, the synthesized nodes are created using an adaptive interpolation rate that is determined by the distance between the minority and majority class nodes, and its neighborhood is built in a stochastic manner based on the distance between the ego-network nodes of a minority node and a target node.
BAT \cite{liu2023topological} generates virtual nodes for each class
as ``shortcuts'' connecting to the other nodes based on posterior likelihoods, where the  representation of the generated nodes is generated based on a feature interpolation operation.

\textbf{Advantages.} (i) Feature interpolation is a simple yet effective way to create more samples for the under-represented anomaly classes. (ii) Mixing up feature representations from different classes can diversify the training data, enhancing the training of a more generalized GAD model. 

\textbf{Disadvantages.} (i) Feature interpolation focuses on generating node representations but lacks the ability to generate graph structural information. Typically, the methods in this category require the incorporation of an additional local structure generator to address this limitation. (ii)  Feature interpolation also has the risk of producing some ambiguous graph samples which may lead to harder separation between normal and anomaly classes.

\textbf{Challenges Addressed.} Feature interpolation provides a simple but effective way to augment the anomaly data, which helps mitigate the bias due to data imbalance (\textbf{C4}). Further, it may generate abnormal graph instances that are dissimilar to the training anomaly instances, thereby improving the generalization ability of GAD models to some unknown anomalies (\textbf{C3}).

\subsubsection{Noise Perturbation}
Unlike the feature interpolation methods that generate new graph instances based on interpolation between representations of existing instances, the noise perturbation generation methods aim to generate graph instances using prior-driven noise perturbation, as shown in Fig. \ref{fig:part2}(b). This approach can incorporate prior knowledge of the graph normality/abnormality into the generation process for more effective GAD.
The generated graph samples as abnormal graph instances, combined with the given labels for existing nodes, can then be leveraged to guide the training of a discriminator for GAD.

The approach can be generally formulated as follows 
\begin{equation}
\begin{array}{l}
{{\bf{X}}^{new}},{{\bf{A}}^{new}} = g_{\phi}(\bf{X}, \bf{A};\epsilon),\\
{L_{cls}} = \sum\limits_{i = 1}^N \ell  \left( {{y_i},{f_\theta }\left( {{{\bf{X}}_i},{{\bf{A}}_i}} \right)} \right) + \sum\limits_{i = 1}^M \ell  \left( {1,{f_\theta }\left( {{\bf{X}}_i^{new},{\bf{A}}_i^{new}} \right)} \right),
\end{array}
\end{equation}
where $\epsilon$ is noise perturbation typically generated from a probability distribution, such as a Gaussian distribution, $g_{\phi}(\cdot)$ is a generation function parameterized by $\phi$, and $M$ is the number of generated graph instances.

\textbf{Assumption.} Certain prior distributions can be used as a source of 
noise perturbation to generate pseudo-abnormal graph instances and/or diversify normal graph instances.

One group of methods in this category \cite{liu2022dagad,qiao2024generative,cai2023self,chen2023consistency} takes a representation permutation approach, which focuses on applying permutation to graph representations to instantiate $ {g_\phi }(\cdot)$, \ie,  $\mathit{Permutation}({\bf{Z}})$. It first utilizes a GNN to obtain the representations from the original graph data $\bf{X}$ and $\bf{A}$. Then it applies the permutation to the representations to generate the representation of anomalous samples, denoted as $\bf{\tilde Z}$.
\begin{equation}
\begin{array}{l}
 {\bf{\tilde Z}} = \mathit{Permutation}({\bf{Z}}; \epsilon),
 {\bf{Z}} = {GNN}\left( {{\bf{X}},{\bf{A}}} \right)
\end{array}
\end{equation}
where $\epsilon$ are the hyperparameters of permutation.
For example, DAGAD \cite{liu2022dagad} employs the permutation and concatenates on the representation learned from a limited number of labeled instances to generate the anomalous sample, thereby enriching the knowledge of anomalies captured in the training set. 
On the other hand, GGAD \cite{qiao2024generative} aims to generate outlier nodes that assimilate anomaly nodes in both local structure and node representations by leveraging the two priors of anomaly nodes, including asymmetric local affinity and egocentric closeness, to impose constraints the representations. SDGG \cite{cai2023self} takes a similar approach as GGAD, but it is focused on generating abnormal graphs  
that closely resemble fringe normal graphs, which are then used to train graph-level anomaly detectors. 
Unlike GGAD and SDGG that work on exclusively normal training data, ConsisGAD \cite{chen2023consistency} focuses on unlabeled nodes. It creates a noise version of the unlabeled nodes by injecting noise into their representations to synthesize instances that maintain high consistency with the original instances while involving as much diversity as possible. This operation helps mitigate the scarcity of abnormal node instances while also enriching the diversity of normal nodes. 

Apart from the representation permutation-based generation, denoising diffusion probabilistic models (DDPMs) \cite{ho2020denoising} have recently been emerging as another major approach to generate anomalous graph instances \cite{liu2023data, ma2023new}.
The diffusion process is defined as a Markov chain that progressively adds a sequence of scheduled Gaussian noise to corrupt the original data ${\bf{x}}$:
\begin{equation}
    \mathbf{Z}^s=\mathbf{Z}^0+\sigma(s) \boldsymbol{\varepsilon}, \boldsymbol{\varepsilon} \sim \mathcal{N}(\mathbf{0}, \boldsymbol{I}),
\end{equation}
where $\bf{Z^0}$ is the feature representations of the graph instances and the variance of the noise (the noise level) is exclusively determined by $\sigma(s)$. The full denoising process is equivalent to a reverse Markov chain that attempts to recover the original data from noise. It iteratively denoises $\hat{\mathbf{Z}}^s$ to obtain $\hat{\mathbf{Z}}^s_{i-1}$ via
denoising function. The estimated $\hat{\mathbf{Z}}^0$ is then fed into a graph instance generator to generate anomalous graphs for GAD.
In particular, GODM \cite{liu2023data} 
employs iterative denoising to synthesize pseudo anomalous graph instances that have close distribution with the real anomalies in a latent space. DIFFAD \cite{ma2023new} leverages the generative power of denoising diffusion models to synthesize training samples that align with the original graph instance in egonet similarity. The generation of supplementary and effective training samples is utilized to mitigate the shortage of labeled anomalies.

\textbf{Advantages.} (i) The graph instance generation may allow the synthesis of novel anomalies that facilitate the identification of unseen anomalies. (ii) Noise perturbation generation can create more diversified training samples, making the trained models less sensitive to small changes in the input data.
\textbf{Disadvantages.} (i) The generated instances can inevitably include some out-of-domain examples, which may deviate from the optimization objective, rendering the detection models less effective. (ii) Determining the optimal level of noise in the noise perturbation can be challenging and may require extensive experimentation.  

\textbf{Challenge Addressed.} The generation of instances can enhance the generalization of detecting different graph anomalies (\textbf{C3}). It also helps achieve a balanced GAD by enhancing the training samples through the GAD-oriented generation (\textbf{C4}).

\section{Proxy Task Design} \label{ref:proxy}
\begin{figure}
 \centering
 \includegraphics[width=3.70in,height=1.8in]{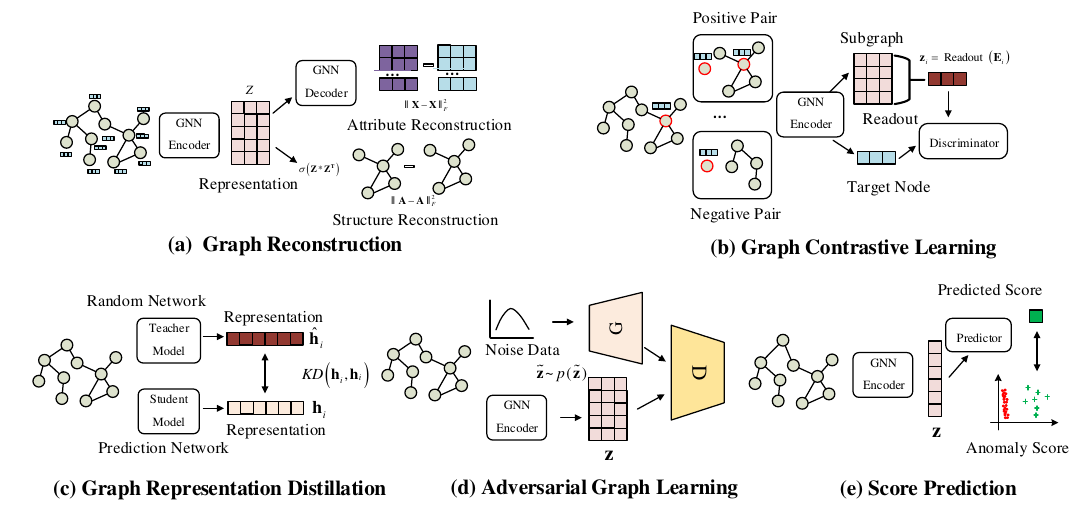 }\\
 \caption{Five categories of proxy task design.
 }
 \vspace{-1em} 
 \label{fig:part3}
\end{figure}

The proxy task design-based approaches aim to capture diverse normal/abnormal graph patterns by optimizing a well-crafted learning objective function that aids the detection of anomalies in the graph data without the use of human-annotated labels. One of the crucial challenges in proxy task design is to guarantee that i) the proxy task is associated with GAD, and ii) it can deal with rich structure information and complex relationships in graphs. We roughly divide the methods in this group into five categories according to the employed proxy tasks, including reconstruction,  contrastive learning, knowledge distillation, adversarial learning, and score prediction. Their respective framework is shown in Fig. \ref{fig:part3}. Below we introduce each of them in detail.
\subsection{Graph Reconstruction}
Data reconstruction aims to learn low-dimensional feature representations of data for reconstructing given data instances, which is widely used in tabular data, and image/video data to detect anomalies \cite{pang2019deep}.  Given that the normal samples often occupy most of the dataset, it helps guarantee that the representations can retain the normal information, and thus, normal samples will be reconstructed with a smaller reconstruction error than anomaly samples. As a commonly used technique in this category, Graph autoencoder (GAE) \cite{hinton2006reducing} consists of a graph encoder and a graph decoder and it is easy to implement the reconstruction process. As shown in Fig. \ref{fig:part3}(a), given a graph with $\bf{X}$ and ${\bf{A}}$, the graph reconstruction can be formulated as
\begin{equation}
\begin{array}{l}
{\bf{Z}} = GN{N_{enc}}\left( {{\bf{X}},{\bf{A}};{\Theta _{enc}}} \right), \\
\widehat {\bf{X}} = GN{N_{dec}}\left( {{\bf{Z}},{\bf{A}};{\Theta _{dec}}} \right),\\
{\widehat {\bf{A}}}_{ij} = p\left( {{{\widehat {\bf{A}}}_{ij}} = 1\mid {{\bf{z}}_i},{{\bf{z}}_j}} \right) = {\mathop{\rm sigmoid}\nolimits} \left( {{{\bf{z}}_i}{\bf{z}}_j^T} \right),
\end{array}
\end{equation}
where the GNN encoder and decoder, which typically follow the general form of the FAG mechanism in Eq. (\ref{eq:FAG}), are designed to obtain the
feature representation for each node by aggregating the representations of its neighbors. They are parameterized by ${\Theta _{Enc}}$ and ${\Theta _{Dec}}$ respectively. $\bf{Z}$ are the representations of the nodes in the latent space. ${\bf{\hat X}}$ 
% . The attribute and structure of the graph are reconstructed using $GNN_{enc}$ and similarly measured. 
and ${\widehat {\bf{A}}}$ are the reconstructed attributes and graph structure respectively. The optimization objective of the methods in this group can be unified as follows
\begin{equation}\label{eqn:reconstruction}
{L_{{\rm{reconstruction }}}} = {\cal R}({\bf{A}},\widehat {\bf{A}}) + {\cal S}({\bf{X}},\widehat {\bf{X}})) + \Phi (\bf{A},\bf{X}; \Theta ),
\end{equation}
where ${\cal R(\cdot)}$ and ${\cal S(\cdot)}$ are two reconstruction functions on graph attributes and structure respectively, and $ \Phi (\bf{A},\bf{X}; \Theta )$ is an auxiliary optimization task to enhance the reconstruction for better GAD. A general assumption underlying the methods of this approach is as follows.

\textbf{Assumption.} Normal graph instances can be well reconstructed while reconstructing the anomalies in terms of graph attributes and/or structure will lead to a large error.

There have been many GAE-based GAD methods, among which early methods typically instantiate the encoder parameter $\Theta_{enc}$ and decoder parameter $\Theta_{dec}$ using different GNN models \cite{ding2019deep, huang2021hybrid, li2019specae,bandyopadhyay2020outlier,fan2020anomalydae,luo2022comga, peng2020deep,yu2018netwalk}. DOMINANT \cite{ding2019deep} is among the seminal studies applying GAE to detect anomalies, in which both $\Theta_{enc}$ and $\Theta_{dec}$ are specified by a three-layer graph convolution.
% and the three-layer graph convolution respectively in DOMINANT.
The optimization terms ${\cal R}$ and ${\cal S}$ are specified using the pointwise difference between matrices:
\begin{equation}
{L_{\rm{reconstruction}}} = (1 - \alpha )||{\bf{A}} - {\bf{\hat A}}||_F^2 + \alpha ||{\bf{X}} - {\bf{\hat X}}||_F^2,
\end{equation}
where $\alpha$ represents the adjustment ratio between attribute and structure reconstruction. The anomaly score is often defined by the mean squared errors (MSE):
% is the common form for anomaly score calculation in reconstruction 
\begin{equation}
\mathit{Score}({v_i}) = (1 - \alpha )||{\bf{a}}_i - {{\hat{\bf{a}}}_i}||_2 + \alpha ||{\bf{x}}_i - {\hat{\bf{x}}_i}||_2,
\end{equation}
where ${\bf{a}}_i$ and ${{\hat{\bf{a}}}_i}$ are the original and reconstructed structure associated with node $v_i$, ${\bf{x}}_i$ and ${\hat {\bf{x}}_i}$ are the original and reconstructed attributes of node $v_i$. Later studies explore the use of more advance GAE by incorporating attention, spectral information, etc \cite{fan2020anomalydae,luo2022comga, peng2020deep}. For example, AnomalyDAE \cite{fan2020anomalydae} and GUIDE \cite{yuan2021higher} employ a graph attention network (GAT) or an attention module to evaluate the significance of neighbors to nodes to enhance the reconstruction of the specific nodes during the training.  
HO-GAT \cite{huang2021hybrid} develops a hybrid order GAT network to specify the $\Theta_{enc}$ that can detect abnormal nodes and motif instances simultaneously. 
Different from the focus on the importance of neighboring nodes, ComGA \cite{luo2022comga} is focused on enhancing node representations for more effective reconstruction. It achieves this by utilizing a community-aware GNN that propagates
a community-specific representation of each node into the feature representations to encode and decode the modularity matrix of the graph. 
Besides node reconstruction, some methods explore node relation construction, \eg, GAD-NR \cite{roy2024gad} that aims to reconstruct the neighborhoods' structure and attributes. GAE can also be extended to perform multi-view graph reconstruction for learning more expressive representations from heterogeneous attributes \cite{peng2020deep}. Despite their simplicity, such GAE-based methods have shown effective performance in not only anomalous node detection but also anomalous edge detection \cite{yu2018netwalk,duan2020aane}.

Another line of research is to complement the reconstruction loss with some auxiliary tasks, \ie, $\Phi (\bf{A},\bf{X}; \Theta)$ in Eq. \ref{eqn:reconstruction}, 
to capture additional information for GAD, since the reconstruction task is often too simple and vulnerable \cite{li2019specae,zhang2019robust, kim2023deep, niu2023graph}. 
In SpecAE \cite{li2019specae}, it integrates a density estimation into the reconstruction by leveraging Laplacian sharpening to amplify the distance between representations of anomalies and the majority of nodes. 
MSAD \cite{kim2023deep} employs a number of weighted meta-paths, \eg, unknown-anomaly-unknown and anomaly-unknown-unknown node paths, to extract context-aware information of nodes as an auxiliary task. 
In HimNet \cite{niu2023graph}, hierarchical memory learning is incorporated via $\Phi (\bf{A},\bf{X}; \Theta)$, where the node-level and graph-level memory modules are jointly optimized to detect both locally and globally anomalous graphs. Netwalk \cite{yu2018netwalk} incorporates clustering as an auxiliary task in the reconstruction to learn the representation of nodes for streaming graph data. 
On the other hand, $\Phi (\bf{A},\bf{X}; \Theta)$ is specified using adversarial learning in DONE \cite{bandyopadhyay2020outlier} to generate node embeddings with an objective to minimize the adverse effects from outlier nodes.

\textbf{Advantages.} (i) Data reconstruction is a simple but effective way to detect anomalies in graph data.
Meanwhile, it does not require labeled data for training. 
(ii) The data reconstruction models can be generalized to detect anomalies at different levels of graph instances, \eg, anomalous nodes, edges, and graphs.

\textbf{Disadvantages.} (i) To perform the reconstruction of graph structure and attributes, it is necessary to load the entire graph, which may require significant computing resources. (ii) The reconstruction is vulnerable to noisy or irrelevant attributes. The presence of noisy or erroneous graph instances can disrupt the reconstruction process, resulting in false positives or diminished detection accuracy.

\textbf{Challenge Addressed.} Reconstruction methods model the entire graph structure, capturing complex structural patterns for GAD (\textbf{C1}). Due to simplicity and straight intuition, they are also applicable to the detection of different levels of graph anomalies (\textbf{C3}).

\subsection{Graph Contrastive Learning}
Graph contrastive learning (GCL) aims to learn effective representations without human annotated labels so that similar graph instances are pulled together in the representation space, while dissimilar instances are far apart. A principled GCL framework for GAD is to 
devise a self-supervised contrastive learning task with suitable positive and negative pairs to learn the underlying dominant (normal) patterns in graph-structured data. As shown in Fig. \ref{fig:part3}(b), the framework generally first utilizes a GNN network to learn the representations of graph instances and build the positive pair and negative pair based on the structure, and then employs a contrastive loss function $L_{\text {c}}$ to maximize the similarity of the positive views while minimizing the similarity of the negative pairs \cite{velivckovic2018deep}:
\begin{equation}
\begin{array}{l}
{{\bf{h}}_i} = GNN\left( {{{\bf{x}}_i},{\bf{X}},{\bf{A}};{\Theta}} \right), \\
{{\bf{s}}_i} = Positive({v_i},{\bf{A}},{\bf{X}}; \Omega_p),
{{{\bf{\tilde s}}}_i} = Negative({v_i},{\bf{A}},{\bf{X}}; \Omega_n),\\
L_{\text {c}}=\mathbb{E}_{(\mathbf{X}, \mathbf{A})}\left[\log \mathcal{D}\left(\mathbf{h}_i, \mathbf{~s}_i\right)+\log \left(1-\mathcal{D}\left(\mathbf{h}_i, \tilde{\mathbf{s}}_i\right)\right)\right],
\end{array}
\end{equation}
where ${{\bf{h}}_i}$ is the representation of target node $v_i$, ${\bf{s}}_i$ and $\tilde{\mathbf{s}}_i$ are the graph instances generated from two sampling functions $Positive(\cdot)$ and $Negative(\cdot)$ with parameters $\Omega_p$ and $\Omega_n$ that are used to sample the positive and negative sample pairs for the target node $v_i$. One key assumption made in the GCL-based GAD methods is that non-neighboring nodes can be treated as `anomalous' graph instances relative to a target node:

\textbf{Assumption.} Non-neighboring nodes to a target node are dissimilar, and thus, they can serve as effective negative ('anomalous') samples to the target node.

This assumption works in that the datasets typically contain a substantial number of normal graph instances. Effective GCL training would enable the learning of the majority patterns on the deviated non-neighboring nodes w.r.t. the target nodes. Built upon this assumption, GCL-based methods are focused on how the positive and negative graph instances $\mathbf{s}$ and $\tilde{\mathbf{s}}$ can be generated to be more aligned with the GAD task.

One group of methods in this line aims to generate node-level contrastive sample pairs \cite{chen2022gccad,wang2021decoupling}. In particular, motivated by the success of the popular GCL method DGI \cite{velivckovic2018deep},
DCI \cite{wang2021decoupling} formulates the GCL objective as the classification of the pairs of target nodes and the nodes under perturbation against the pairs of target nodes and cluster-based global representations. 
Unlike the cluster-based representation, GCCAD \cite{chen2022gccad} 
specifies $\mathbf{s}$ using the neighbors of target normal nodes and $\tilde{\mathbf{s}}$ using the representations of abnormal nodes. During inference, the anomaly score may be defined in various ways, \eg, via the classification probability \cite{wang2021decoupling} or similarity between the target node's representation and the graph representation \cite{chen2022gccad}.

In addition to the node-level contrasts, other methods use subgraphs as the target of contrastive learning to help the model learn better representations for GAD. CoLA \cite{liu2021cola} is an early GCL framework for GAD that learns the relations between each node and its neighboring substructures. Given a target node, $\mathbf{s}$ is a subgraph generated by a random walk around the target node, while the negative subgraph is generated using the random walk around the other nodes. 
Let ${\bf{h}}_i$ and $\mathbf{E}_i$ be the representations of a target node and a subgraph, in which $\mathbf{E}_i$ is often obtained by a readout function as follows
\begin{equation}
    \mathbf{s}_i=\operatorname{Readout}\left(\mathbf{E}_i\right)=\sum_{k=1}^{n_i} \frac{\mathbf{v}_k}{n_i},
\end{equation}
where $n_i$ is the number of nodes in $\mathbf{E}_i$, $\mathbf{v}_k$ is the embedding of node $k$ in the subgraph,
then a  $\rm{ Bilinear(\cdot)}$ function is often used to combine the representations of the node and the subgraphs 
\cite{liu2021cola, zheng2021generative}:
\begin{equation}
{
\begin{array}{l}
{y_i} = {\rm{ Bilinear }}\left( {{{\bf{h}}_i},{{\bf{s}}_i}} \right) = \sigma \left( {{{\bf{h}}_i}{\bf{W}}{{\bf{s}}_i}^ \top } \right), \\
{{\tilde y}_i} = {\rm{ Bilinear }}\left( {{{\bf{h}}_i},{{\bf{s}}_j}} \right) = \sigma \left( {{{\bf{h}}_i}{\bf{W}}{{\bf{s}}_j}^ \top } \right),
\end{array}
}    
\end{equation}
parameterized by $\bf{W}$, 
in which ${y_i} $ and ${{\tilde y}_i}$ are the predicted results for positive pairs $({\bf{h}}_i,{\bf{s}}_i)$ and negative pairs $({\bf{h}}_i,{\bf{s}}_j)$ (\ie, ${\bf{s}}_j$ acts as $\tilde{\mathbf{s}}_i$ here), respectively. 
The anomaly score in CoLA is defined as the difference between the positive and negative pairs:
\begin{equation}
    Score\left( {{v_i}} \right) = \frac{1}{R}{\sum\limits_{r = 1}^R {\left( {{\tilde y}}_{i,r} - y_{i,r} \right)} },
\end{equation}
where $R$ is the number of node-subgraph pairs sampled during inference to obtain a stable anomaly score.
This framework inspires a number of follow-up methods, including the use of prior knowledge of different anomaly types to generate the negative samples \cite{xu2022contrastive}, supervised positive/negative subgraph pairs \cite{zhou2023learning}, and multi-view/scale subgraph generation \cite{jin2021anemone,zheng2021generative,duan2023graph,xu2023few}. 
The above methods are focused on node-level anomaly detection in static graph data. Contrastive learning is also used in dynamic GAD or graph-level anomaly detection. For example,  TADDY \cite{liu2021anomaly} applies a dynamic graph transformer that aggregates spatial and temporal knowledge simultaneously to learn the representations of edges.  It specifies ${\bf{s}}_i$ and ${\tilde{\mathbf{s}}_i}$ by constructing the positive edge using the existing edges in the training set and generates the anomalous edges via negative sampling. 
SIGNET \cite{liu2024towards} is designed for graph-level anomaly detection, which first constructs two different views using dual hypergraph transformation and then maximizes the mutual information between the bottleneck subgraph from two views. The estimated mutual information can be used to evaluate the graph-level abnormality. 

\textbf{Advantages.} (i) Many existing GCL approaches and theories may be adapted to enable GAD.
(ii) The rich graph structure information provides flexible options to generate diverse positive/negative views for effective GAD. (iii) Since many methods rely on only local graph information in their training, they can handle very large graph data. 

\textbf{Disadvantages.} (i) Since GCL is focused on representation learning, it is crucial to develop an effective anomaly scoring method based on the learned representations.
(ii) As GCL methods rely on GNNs without class information, the problem of over-smoothing between normal and abnormal instances remains prevalent.
(iii) The subgraph generation in some methods may incur significant computational costs due to the need to traverse numerous nodes and edges.

\textbf{Challenge Addressed.} Contrastive learning models can be designed to capture different levels of graph structure and graph anomalies (\textbf{C1}, \textbf{C3}). Without the need to load the full graph structure information, they can often scale up to large-scale graph data (\textbf{C2}).

\subsection{Graph Representation Distillation }

Knowledge Distillation (KD) \cite{gou2021knowledge} aims to train a simple model (student) that distills feature representations from a large (teacher) model while maintaining similar accuracy as the large model. 
The key intuition of KD-based GAD is that the representation distillation can capture the majority patterns of graph instances and the difference in the distillation can be used to measure the abnormality of samples.

\textbf{Assumption.} The graph representation distillation can be seen as a process of extracting the prevalent patterns of the graph instances, representing the normal patterns for GAD.

As shown in Fig. \ref{fig:part3} (c), this category of methods learns the representation from a teacher model initialized by a GNN. Subsequently, a GNN-based student network is trained to replicate the representation outputs of the teacher model. The GNN-based teacher and student networks are formulated as follows

\begin{equation}
\begin{array}{l}
{{\bf{h}}_i} = GN{N_{{\rm{teacher}}}}\left( {{{\bf{x}}_i},{\bf{X}},{\bf{A}};{\Theta }} \right),\\
{{{\bf{\hat h}}}_i} = GN{N_{{\rm{student}}}}\left( {{{\bf{x}}_i},{\bf{X}},{\bf{A}};\hat \Theta } \right),
\end{array}
\end{equation}
where $\Theta$ and  $\hat{\Theta}$ are respectively the training parameters of student and teacher networks, ${\bf{h}}_i$ and ${\bf{\hat h}}_i$ are the representations learned by the two networks respectively. Both representations are then integrated into the loss function which can be formulated as the following: 
\begin{equation}
  L_{KD} = \frac{1}{N}\sum\limits_{i = 1}^N {KD\left( {{{\bf{h}}_i},{{\widehat {\bf{h}}}_i}};\hat{\Theta}, \Theta \right)} ,
\end{equation}
where $KD(\cdot)$ is a distillation function that measures the difference between the two feature representations. Overall, the goal of the distillation-based GAD is to make the student model as close as
possible in predicting the corresponding outputs of the teacher model that is built upon normal graph data.  Therefore, the anomaly score can be defined as the difference in the representations between the teacher and student models.
\begin{equation}
    Score\left(v_i ; \hat{\Theta}, \Theta\right)=\left\|\mathbf{h}_i-\widehat{\mathbf{h}}_i\right\|_2.
\end{equation}

GlocaKD \cite{ma2022deep} is an early framework, in which the teacher model $GN{N_{{\rm{teacher}}}}$ is implemented using a random GCN. Then the distillation function $ KD(\cdot)$ is instantiated using KL divergence to minimize the graph- and node-level prediction errors of the representations yielded by the random GCN. The anomaly score in GlocaKD is defined as the prediction error at the graph and node levels. To support better distillation of the normal graph representations, several distillation models have been proposed for GAD with new architectures. For example, the dual-student-teacher model, called GLADST, consists of one teacher model and two student models \cite{lin2023discriminative}, in which the teacher model, trained with a heuristic loss, is designed to make the representations more divergent. Unlike the traditional teacher-student model, GLADST trains the two student models on normal and abnormal graphs separately to capture the normality and abnormality better. Unlike GlocalKD that uses a random GNN to be the teach network, the approach FGAD uses a pre-trained anomaly detector as the teacher \cite{cai2024fgad}. Then the student is designed with a graph isomorphism network (GIN) and a projection head to improve the robustness of GAD.

\textbf{Advantages.} (i) The distillation from the teacher model into a simpler student model provides a new way to extract the normality of graph instances. (ii) Distillation enables the development of smaller, more efficient models by transferring knowledge from a larger model. This compression speeds up inference and reduces computational resource requirements for normality extraction. (iii) Distilling knowledge from a well-trained teacher model enables the student model to better generalize across various types of graphs and quickly adapt to new data or target domains.

\textbf{Disadvantages.}  (i) Choosing an appropriate teacher model can be challenging. If the teacher model is too complex or not well-suited to GAD, the knowledge distilled to the student model may not be optimal.  (ii) The student model may fail to capture all the nuances and intricacies in the teacher model, which can impact the extraction of normality.

\textbf{Challenge Addressed.} 
GNN-enabled knowledge distillation enhances the ability of addressing graph structure-aware GAD. (\textbf{C1}).  
By distilling knowledge from a well-trained teacher, the student gains improved generalization across different GAD scenarios, showcasing enhanced robustness in GAD (\textbf{C5}).

\subsection{Adversarial Graph Learning} 
Generative adversarial learning (GAN) provides an effective solution to generate realistic synthetic samples, 
which can be used for normal pattern learning for GAD.  The key intuition of this group of methods is to learn latent features that can capture the normality perceived in a generative GNN network. Specifically, the GANs employ a generator network aiming to generate samples that are statistically similar to the real data while the discriminator network learns to distinguish between real and generated graph instances \cite{ding2021inductive, chen2020generative}. The normal data with a prior distribution can be easily captured by the generator network while the anomalies struggle to be simulated by the generator due to the nature of the distribution. 

It is worth mentioning that the purpose of generation is different from the noise perturbation generation in the generative GNNs in Sec. \ref{subsec:generative_gnn}. The former mainly uses the graph structure information and interpolation operations in the latent space to generate new node representations without using adversarial learning process, while the latter is focused on generating node representations from a prior distribution through the adversarial learning to learn the latent normality.

\textbf{Assumption.} A generator GNN can capture the majority of patterns in the graph data if it can generate instances that closely resemble the distribution of real graph instances.

This group of methods (see Fig. \ref{fig:part3} (d)) typically employs GNNs to learn the representations of graph instances and a generator network for generating graph instances based on a prior. The discriminator is utilized to distinguish whether a graph instance comes from the generator or the original data. Formally, these methods follow the following framework:
\begin{equation}
\begin{array}{l}
{{\bf{h}}_i} = GNN\left( {{{\bf{x}}_i},{\bf{A}},{\bf{X}};\Theta } \right),\\
\tilde{\mathbf{h}}_i=G\left(\tilde{\mathbf{z}}_i ; \epsilon\right), \tilde{\mathbf{z}}_i \sim p(\tilde{\mathbf{z}}),
\end{array}
\end{equation}
where $\bf{h}$ is the feature representation of a node learned by GNN with parameter $\Theta$, ${\tilde{\bf{h}}}$ is the representation of a generated node, and $p(\widetilde {\bf{z}})$ is the prior distribution.  The generator $G(\cdot)$ takes noises sampled from the prior distribution $p{(\tilde{\bf{z}})}$ as the input and generates synthetic pseudo abnormal graph instances via:
\begin{equation}
\min _G \max _D \mathbb{E}_{\mathbf{h} \sim p(\mathbf{h})}[\log D(\mathbf{h})]+\mathbb{E}_{\tilde{\mathbf{z}} \sim p(\tilde{\mathbf{z}})}[\log (1-D(G(\tilde{\mathbf{z}}; \epsilon)))],
\end{equation}
where the discriminator $D(\cdot)$ is often specified by a classifier that tries to distinguish whether an input is the representation of a normal node or a generated anomaly.
The anomaly score is typically defined based on the output of the discriminator:
\begin{equation}
Score({s_i}) = 1 - D({{\bf{h}}_i}).
\end{equation}

AEGIS \cite{ding2021inductive} and GAAN \cite{chen2020generative} are two representative works that apply GANs to GAD by generating node representations from Gaussian noise. The discriminator is trained to determine whether nodes are real or generated pseudo abnormal instances.
Some graph adversarial learning methods are also proposed to address the multi-class imbalance problem at the node level \cite{shi2020multi, qu2021imgagn}. They incorporate adversarial training to make the model learn robust representations for both majority and minority classes, thereby benefiting the separation of the nodes from different classes.
This graph adversarial learning is also applied to anomalous edge detection. GADY \cite{lou2023gady} employs an anomaly generator to generate abnormal interactions through input noise. The generated interactions are then combined with normal interactions as the input of a discriminator network trained to determine whether the interaction is normal or abnormal.

\textbf{Advantages.} 
(i) GANs provide a distinctive method for learning structural normality by utilizing their graph structure-aware generation capability.
(ii) Its adversarial training can generate realistic samples from noise, enabling the detection models to learn beyond the abnormal samples in the graph. 

\textbf{Disadvantages.} (i) It is difficult to generate samples that accurately simulate real graph instances in terms of both graph structure and attributes, and thus, the generated graph instances may impair the detection performance. (ii) The training of GANs is relatively less stable compared to the GNN training in other groups of methods.

\textbf{Challenge Addressed.} GANs can model and generate different types of abnormal graph instances, facilitating the detection of different graph anomalies, \eg, anomalies that are unseen during training (\textbf{C3}). Also, GANs can generate extensive abnormal samples for more balanced GAD (\textbf{C4}).

\subsection{Score Prediction}

The score prediction based-methods focuses on how to make full use of labeled data to build an end-to-end anomaly score prediction model, as shown in Fig. \ref{fig:part3} (e). Unlike approaches that directly apply GNNs for classification with labeled abnormal and normal nodes, this method is designed for the scenarios where only some graph instances are known to be normal and abnormal instances.

\textbf{Assumption.} The anomaly scores of normal graph instances follow a prior distribution while those for abnormal instances significantly deviate from the distribution. 

The score prediction-based methods refer to training a predictor $f_{\text {pred }}: G \rightarrow \mathbb{R}$ which is instantiated with GNNs to directly predict the anomaly score 
\begin{equation}
Score(s_i) =  {{f_{\text {pred }}}({\bf{x}}_i, {\bf{A}},{\bf{X}};{\Theta _p})}, 
\end{equation}
where  $\Theta _p$ is the training parameters of the score prediction network. DevNet \cite{pang2019deep} is a seminal work for score prediction network, which was originally proposed to identify anomalies in tabular data. It employs a Z-score-based deviation loss to learn the anomaly scores in an end-to-end manner:

\begin{equation}
    L_{\text {deviation}}=\left(1-y_i\right) \cdot\left|Dev(s_i)\right|+y_i \cdot \max \left(0, m-Dev(s_i)\right),
\end{equation}
where $y_i$ is the class label, $m$ is a pre-defined margin based on the prior distribution, and $Dev(s_i)$ is defined as follows 
\begin{equation}
{Dev} \left( {{s_i}} \right) = \frac{{Score(s_i) - {\mu _r}}}{{{\sigma _r}}},
\end{equation}
where $\mu_r$ and $\sigma_r$ are the estimated mean and standard deviation of the anomaly scores based on the prior ${\mathcal N\left(\mu, \sigma\right)}$. 

Meta-GDN \cite{ding2021few} applies the deviation loss to the graph data that leverages a small number of labeled anomalies to enforce significant deviation of the anomaly scores of the normal nodes from the abnormal nodes. 
SAD \cite{tian2023sad} adapts DevNet to dynamic graph data, in which contrastive learning is also used to fully exploit the potential of labeled graph instances on evolving graph streams. In WEDGE \cite{zhou2023learning}, the deviation loss function is defined for at the subgraph level. By minimizing the deviation loss, the score network predictor will enforce a large positive deviation of the anomaly score of an
anomalous subgraph from that of the prior-based reference scores.

\textbf{Advantages.} (i) By integrating the prior distribution into the model’s learning process, it can produce more interpretable anomaly scores compared to other detection methods. (ii)
The studied scenarios where some labeled normal and anomalous graph instances are available are often common in real-world applications.

\textbf{Disadvantages.} (i) The performance of the score prediction model is dependent on the prior and the predefined margin used during training. (ii) The score prediction network is better suited for tabular data because the samples are independent, but a single prior distribution may not be able to effectively capture the dependent scores across the graph instances.

\textbf{Challenge Addressed.} The score prediction provides a GNN-based end-to-end anomaly score learning framework for GAD, having good scalability to large-scale graph data (\textbf{C1}, \textbf{C2}). It also provides an effective way to achieve generalized GAD in the application scenarios where part of the graph instances are labeled (\textbf{C3}).

\section{Graph Anomaly Measures} \label{ref:measure}
\begin{figure}
 \centering
 \includegraphics[width=3.60in,height=1.65in]{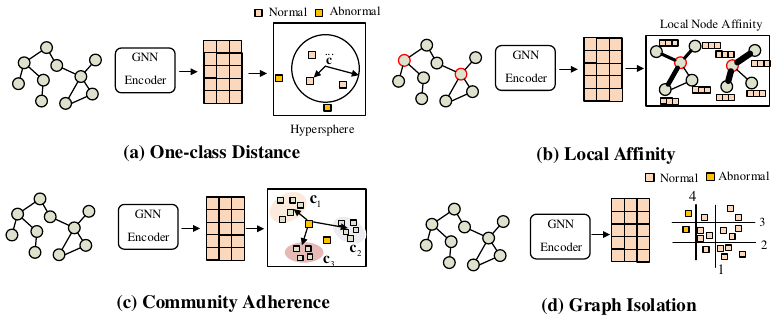 }\\
 \caption{Four categories of graph anomaly measure.}
 \vspace{-1em} 
 \label{fig:part4}
\end{figure}

This category of methods aims to discuss GAD methods that focus on designing anomaly measures for evaluating the abnormality of instances in the graph. These methods generally perform anomaly scoring by incorporating some key abnormal graph characteristics.
As shown in Fig. \ref{fig:part4}, they can be generally divided into four categories, including one-class distance measure, local affinity measure, community adherence measure, and graph isolation-based approaches. 
\subsection{One-class Classification Measure}
One-class classification measures refer to evaluating the distance between each instance and a one-class center of the instances for anomaly scoring \cite{ruff2018deep}. This method can be applied to graph data, with the GNNs trained to minimize the volume of a hypersphere that encloses the representations of the graph instances. As illustrated in Fig. \ref{fig:part4} (a), the key intuition is that anomalous graph instances differ significantly from the normal ones, causing them to fall outside the hypersphere that encompasses most of the normal graph instances.

\textbf{Assumption.} Normal graph instances exhibit similar patterns that can be encapsulated via a one-class hypersphere,
% in a latent space, 
from which anomalies show largely deviated patterns.

The one-class classification on the graph can be generally formulated as the following 
\begin{equation}
  {L_{one - class}} = \frac{1}{N}\sum\limits_{i = 1}^N {||\phi\left({{{\bf{X}}_i},{{\bf{A}}_i}} ; \bf{W}\right) - {\bf{c}}|{|^2}}  + \Phi (\Theta ),\end{equation}
where $\bf{c}$ is the central representation of the one-class hypersphere, 
% ${{\bf{z}}_k}$ \hh{z is not in Eq.30, do u mean phi instead?}
$\phi \left( {{{\bf{X}}_i},{{\bf{A}}_i};{{\bf W}^*}} \right) $ is the representation of graph instance $s_i$ learned by GNN, and $ \Phi(\Theta )$ is a regularization term or auxiliary task which can benefit the one-class distance measure. 
Anomalies are expected to samples that have a large distance to the center.  Thus, the anomaly score can be determined by the distance of a graph instance to the center of the hypersphere:

\begin{equation}
  \operatorname{Score}(s_i)=\left\|\phi\left({{{\bf{X}}_i},{{\bf{A}}_i}} ; \bf{W}^*\right)-\mathbf{c}\right\|_2,
\end{equation}
where ${\bf W}^*$ are the parameters of the trained one-class model and $\mathbf{c}$ is the representation of the one-class center. 

There have been some methods that adopt this one-class classification approach for GAD \cite{zhou2021subtractive,wang2021one}. OCGNN \cite{wang2021one} applies a one-class SVM to graphs, leveraging the powerful representation capabilities of GNNs. The objective of OCGNN is to generate node embeddings that are close to the center.
Since the feature representations are crucial in one-class learning, various methods have explored the use of GNNs from different perspectives to enhance the representation learning for one-class GAD. For example, AAGNN \cite{zhou2021subtractive} designs a subtractive aggregation~\cite{zhou2021subtractive} rather than the commonly used summation-based aggregation for one-class GNN learning. DOHSC \cite{zhang2023deep} adds an orthogonal projection layer \cite{zhang2023deep} to ensure the training data distribution is consistent with the decision hypersphere.
Other methods optimize the one-class learning with some auxiliary tasks, such as node feature reconstruction \cite{teng2018deep}, relation prediction \cite{li2023hrgcn}, and self-supervision \cite{qiu2022raising}, to avoid notorious issues in this approach like model collapse.  

\textbf{Advantages.}
(i) One-class classification does not require labeled anomaly data for training, making it suitable for the scenario where such data is scarce or unavailable.  (ii) The one-class measure can handle isolated nodes well.

\textbf{Disadvantages.} (i) Normal graph patterns can manifest in various ways, making it challenging for a one-class hypersphere to capture the full spectrum of normality. (ii)
Learning the one-class hypersphere is prone to model collapse. 

\textbf{Challenge Addressed.} One-class classification with appropriate GNNs enables the learning of the majority structural pattern in the graph data (normal graph instances) (\textbf{C1}). This approach also does not require the full graph structure information during training, resulting in good scalability to large-scale graphs (\textbf{C2}). 

\subsection{Community Adherence}

Community adherence-based GAD \cite{zhou2022unseen, yu2018netwalk} aims to identify the anomalies based on the adherence of instances to graph communities. The key intuition is that anomalies are not well-distributed and exhibit weak adherence to the communities, whereas normal graph instances typically have strong adherence to at least one community.

\textbf{Assumption.} Normal instances adhere to at least one community, whereas anomalies are unfit to any community.

This type of methods first leverages a mapping function to learn the representation of instances. A clustering or community discovery method is then applied to group the graph instances, as shown in Fig. \ref{fig:part4} (b), where the instances are grouped into three distinct clusters. Since anomalies generally exhibit significantly weaker community adherence, the anomaly score can be defined by the minimum distance to the centers of the communities.
 
\begin{equation}
{\mathop{\rm Score}\nolimits} \left( {{s_i}} \right) = \min \left\| {\phi\left(\mathbf{x}_i, \bf{X}, \bf{A} ; \bf{W}^*\right) - {{\bf{c}}_j}} \right\|_2^2,\forall j \in \{ 1,\ldots ,p\},
\end{equation}
where $p$ is the number of communities, $c_j$ is the center of community $\mathcal{C}_j$, and  $\bf{W}^*$ is the optimal parameters of the mapping function.

This approach is analogous to clustering-based anomaly detection in non-graph data \cite{pang2021deep}, but here it needs to capture the graph characteristics for GAD. To this end, MHGL \cite{zhou2022unseen} utilizes GNNs and a multi-hypersphere learning objective to learn multiple groups of fine-grained normal patterns, enclosing each group using a corresponding hypersphere in the latent space while simultaneously pushing labeled anomalies far away from these hyperspheres. 
The anomaly score is defined as the Euclidean distance between a test instance and the nearest hypersphere center.
 Netwalk \cite{yu2018netwalk} is a method for both anomalous node and edge detection where {\em k}-means clustering was applied to group the existing node/edge into different groups. The anomaly score of node/edge is measured as its distance to the center of its closest cluster.

\textbf{Advantages.} (i) By utilizing graph communities,  the normality of data beyond one-hop graph structures can be more effectively captured. (ii) The community adherence enables a fine-grained modeling of normal patterns, which could be important for identifying some types of graph anomalies that depend on the context of graph communities.

\textbf{Disadvantages.} (i) Community adherence-based methods are sensitive to hyperparameters like the number of clusters. (ii) Community adherence measures rely heavily on the effectiveness of the community detection methods.

\textbf{Challenge Addressed.} Graph communities can be useful for discovering important structural contexts (e.g., those beyond a fixed-hop neighborhood) to detect anomalies that deviate from the communities (\textbf{C1}).
Community adherence can provide one way for interpreting anomalies based on deviations from expected community-based behaviors (\textbf{C5}).

\subsection{Local Affinity}
There are many graph properties that can be important for GAD, such as connectivity, degree distribution, and clustering coefficient. Local affinity is a graph property that integrates multiple properties for evaluating the normality and abnormality of graph instances. As shown in Fig. \ref{fig:part4} (c), the affinity may be defined in various ways, such as the number of connections of a graph instance to neighboring instances, the similarity with neighbors, or the clustering coefficient of the connected subgraphs. The key intuition is that the normal graph instances typically have a strong affinity with their neighbors, whereas an anomalous instance has a significantly weaker affinity with its neighbors. Thus, the local affinity can serve as the inverse of the anomaly score.

\textbf{Assumption.} Normal instances are connected with other normal instances with similar attributes  while anomalies are often graph instances that are less similar to their neighbors.

Formally, the local affinity  $\tau({s_i})$ of an instance $s_i$ can be defined based on its average similarity to its neighbors:

\begin{equation}
    \tau \left( {{s_i}} \right) = \frac{1}{{\left| {{\cal N}\left( {{s_i}} \right)} \right|}}\sum\limits_{{s_j} \in {\cal N}\left( {{s_i}} \right)} {{\mathop{\rm sim}\nolimits} } \left( {{{\bf{h}}_i},{{\bf{h}}_j}} \right),
\end{equation}
where ${\bf{h}}_i$ and ${\bf{h}}_j$ are the representation of instance $s_i$ and $s_j$, ${\cal N}\left( {{s_i}} \right)$ represents the neighboring instance $s_i$.

TAM \cite{hezhe2023truncated} is a seminal work that introduces local affinity as an anomaly measure. It aims to learn tailored node representations for GAD by maximizing the local affinity of nodes to their neighbors. It is optimized on truncated graphs where non-homophily edges are removed iteratively to mitigate its adverse effects on the local affinity measure.  The learned representations result in a significantly stronger local affinity for the normal nodes than the abnormal nodes. CLAD \cite{kim2023class} instead measures the affinity based on the discrepancy between a node and its neighbors using Jenson-Shannon Divergence. The anomaly score is obtained from the affinity for each node in terms of both graph structure and attributes. PREM \cite{pan2023prem} eliminates the message-passing propagation in regular GNNs by using an ego neighbor matching-based contrastive learning module. It aims to learn discriminative patterns between the local ego network and the neighboring instances. These GAD methods are focused on the anomaly score of a node based on its affinity to its neighboring nodes. Similar ideas have also been explored in very recent prompt tuning or in-context learning methods for GAD \cite{niu2024zero,liu2024arc,qiao2025anomalygfm}. Exploring beyond the node-level affinity can be one potential approach for subgraph- or graph-level anomaly detection.

\textbf{Advantages.} (i) The local affinity measure offers a novel way to quantify the abnormality of graph instances at a local scope. (ii) The measure provides a principled framework for evaluating the normality from both the graph structure and attributes. (iii) It can be more interpretable than those identified through task proxy-based GAD methods.
 
\textbf{Disadvantages.} (i) Its effectiveness may vary if the affinity is specified differently.
(ii) It is often designed based on predefined graph properties, making it difficult to generalize to the anomalies that do not conform to the properties.

\textbf{Challenge Addressed.} Local affinity-based GAD methods can adapt to and leverage various structural properties for the detection of various types of anomalies (\textbf{C1}, \textbf{C3}), though the prior knowledge about the properties is required. Local affinity also provides a way to explain why a node is considered anomalous based on the local context, enhancing the interpretability of GAD (\textbf{C5}).

\subsection{Graph Isolation} 
Isolation-based methods \cite{liu2008isolation} is among the most popular methods for anomaly detection.  Due to its general effectiveness across different datasets, it is also applied to identify anomalous graph instances \cite{xu2023deep, zhuang2023subgraph}. Its use for GAD is based on the isolation of graph instances in a feature representation space.

\textbf{Assumption.} Anomalous graph instances can be isolated more easily than normal instances in the representation space.

As shown in Fig. \ref{fig:part4} (d), the methods in this group need to first learn the representations of the instances using a graph encoder $GNN_{enc}$: 
\begin{equation}
{{\bf{h}}_i} = GN{N_{enc}}({s_i},{\bf{X}},{\bf{A}}; \Theta_{enc}),
\end{equation}
where ${\bf{h}}_i$ is the representation of the graph instance $s_i$. An isolation mechanism is then applied on the representations, \ie, $\mathit{Isolation}({\bf{h}}_i)$, where the split process is formulated as the following
\begin{equation}
    \begin{aligned}
& \mathcal{P}_{2 k} \leftarrow\left\{{\bf{h}}_i \mid {\bf{h}}_i^{\left(j_k\right)} \leq \eta_k, {\bf{h}}_i \in \mathcal{P}_k\right\}, \\
& \mathcal{P}_{2 k+1} \leftarrow\left\{{\bf{h}}_i \mid {\bf{h}}_i^{\left(j_k\right)}>\eta_k, {\bf{h}}_i \in \mathcal{P}_k\right\},
\end{aligned}
\end{equation}
where $\mathcal{P}_k$ is the node set of in the $k$-th binary partition tree and $j$ is the dimension in $\mathbf{h}$ used to partition the feature space.
The abnormality of a graph instance  $s$ is evaluated by the isolation difficulty in each tree of the tree set $\mathcal{T}$:
\begin{equation}
    {F}(s \mid \mathcal{T})=\Omega_{\tau_i \sim \mathcal{T}} I\left(s \mid \tau_i\right),
\end{equation}
where $I\left(s \mid \tau_i\right)$ denotes a function to measure the isolation difficulty in tree $\tau_i$ and $\Omega$ denotes an integration function.

DIF \cite{xu2023deep} presents a new representation scheme that combines data partition and deep representation learning to perform isolation in randomly projected deep representations for anomaly detection, showing good effectiveness in anomaly detection in various data types, including graph-level anomaly detection. GCAD \cite{zhuang2023subgraph} uses isolation forest for anomalous node detection. It uses the node representations after subgraph normalization as the input to graph isolation. The anomaly score is defined using a depth-based weighted score that aggregates scores from various associated subgraphs.
If we treat isolation-based measures as simpler alternatives to density estimation, there have been multiple other extensions \cite{ester1996density,li2019specae,ester1996density,zhao2021action} that leverage the learned graph representations to estimate a density-based anomaly score for GAD.

\textbf{Advantages.} (i) Graph isolation measures are built on well established isolation-based methodology for anomaly detection. (ii) Many existing isolation-based methods may be adapted to anomaly detection on graph data.

\textbf{Disadvantages.} (i) The isolation measure operates on a continuous feature space, so its effectiveness relies on the learning of an expressive representation space. (ii) The heuristic of isolation is difficult to be incorporated into GNN-based representation learning, leading to less effective feature representations for the subsequent isolation mechanism.

\textbf{Challenge Addressed.} This measure can adapt traditional anomaly measures to GAD with the power of GNN-based representation learning (\textbf{C1}). The isolation mechanism is highly efficient, allowing good scalability on large graphs (\textbf{C2}).

\section{Research Opportunities}\label{sec:opportunity}
Despite the remarkable success of numerous existing GAD methods, there are a range of research opportunities that could be explored to tackle some largely unsolved GAD problems.

\textbf{Advanced Graph Anomaly Measures.} Most GAD methods are built upon proxy tasks or focused on the GNN backbones using traditional non-graph anomaly measures, as summarized in \textbf{Table 1} in \texttt{Appendix A}. Consequently, they may fail to learn feature representations that encapsulate holistic graph structure and attributes specifically for GAD. Therefore, it is crucial to devise anomaly measures that go beyond traditional anomaly measures and proxy tasks, such as local node affinity \cite{hezhe2023truncated}, for developing more dedicated methods for GAD.

\textbf{GAD on Complex Graphs.} Most existing GAD methods are focused on small-scale (\eg, less than millions of nodes/edges), static, or homogeneous graph data, as shown in \textbf{Tables 2 and 3} in \texttt{Appendix B}. However, many real-world graphs can involve millions/billions of heterogeneous nodes/edges \cite{li2023hrgcn}, such as real-life citation networks, social networks, and financial networks. The nodes/edges may appear in a streaming fashion, where the models can access to only limited graph data at one time step and may need to adapt to new normal/abnormal patterns as the graph evolves \cite{ranshous2015anomaly}. Current methods may be adapted to handle these graphs, but their performance would become less effective since their primary design do not consider those complexity. GAD methods designed for graphs with two or more of these complexities are required.

\textbf{Handling Anomaly Camouflage and Contamination.}
In GAD, anomalies might easily hide their abnormal characteristics by mimicking the structure and attributes of their neighboring instances. There have been some approaches for addressing this problem, \eg, via selecting relevant features, incorporating domain knowledge, or using adversarial training \cite{dou2020enhancing, qiao2024generative}, but they often rely on the prior knowledge about what specific features the attackers may use in the camouflage. A related problem is anomaly contamination in the training data. Current methods are mostly unsupervised, working on anomaly-contaminated training data, but the anomalous instances in the graph can largely bias the message-passing \cite{hezhe2023truncated}, leading to less expressive representations. Recent approaches, such as semi-supervised GAD on a small set of labeled normal instances \cite{qiao2024generative} or training GNNs using truncated graph data \cite{hezhe2023truncated}, may offer effective methodologies for handling these issues.

\textbf{Interpretable GAD.}
As shown by the summarized detection performance results in \textbf{Tables 4, 5, and 6} in \texttt{Appendix C}, current GAD methods have shown impressive success in detecting anomalous graph instances, but they generally ignore the interpretability of their detection results. In addition to accurate detection, interpretable GAD also requires an explanation about why a graph instance is identified as anomalous within a given graph structure, making it different from explaining anomalies in non-graph data. Exploring information such as local graph structure, human feedback, and/or domain knowledge~\cite{liu2024towards} would be some interesting directions for providing the structure-aware anomaly explanation. Also, the obtained anomaly explanation may in turn be further leveraged to improve the detection performance.

\textbf{Open-set Supervised GAD.}
As shown in \textbf{Table 1} in \texttt{Appendix A}, there have been many supervised GAD methods, most of which essentially tackle an imbalanced binary classification problem. Such formulation is often questionable since anomalies per se can draw from very different distributions and cannot be treated as from one concrete class. Open-set supervised GAD also trains the detectors with labeled normal and anomalous examples (\ie, seen anomalies), but it assumes an open set of anomaly classes (\ie, there are anomaly classes that are not illustrated by the training anomaly samples) rather than the closed-set assumption in most existing studies \cite{wang2023open}. Thus, it is a more realistic supervised GAD setting. Methods for this setting have shown significantly better performance than unsupervised and fully supervised methods on visual data \cite{acsintoae2022ubnormal,ding2022catching,yao2023explicit,zhu2024anomaly} and tabular data \cite{pang2019deep,pang2021toward,pang2023deep}. Recent studies also show similar advantages on graph data \cite{wang2023open, zhou2022unseen}. Exploring better modeling of the normal patterns while fitting the seen anomalies could be an effective approach to avoid overfitting of the seen anomalies (\ie, reducing misclassification of the unseen anomalies as normal).

\textbf{Foundation Models for GAD.}
Leveraging foundation models (FMs) for downstream tasks has been emerging as one effective direction to empower the sample-efficient performance in the downstream tasks, including graph-related tasks \cite{xia2024opengraph, sun2023all,liu2023towards, tang2024higpt}, owing to their superior generalization ability. Two main directions for the GAD task involve the training of graph foundation models (GFMs) for GAD and the exploitation of large language models (LLMs) for GAD. There have been a number of successful tuning of FMs for anomaly detection on image data \cite{zhu2024toward,zhou2023anomalyclip,jeong2023winclip} and video data \cite{wu2024vadclip,wu2024weakly,wu2024open,sato2023prompt} via proper prompt crafting, prompt learning, or in-context learning. Similar approaches may be explored for GAD. For example, inspired by \cite{zhu2024toward}, an in-context residual learning-based method is explored for few-shot GAD \cite{liu2024arc}; general abnormality/normality graph patterns and prompts are learned for zero/few-shot cross-dataset GAD in \cite{niu2024zero,qiao2025anomalygfm}. This new paradigm is plausible for GAD in several aspects, such as zero/few-shot detection on target graph data, inductive GAD, 
and interpretable GAD with text description.

\section{Conclusion}
In this survey, we first discuss the complexities and existing challenges in GAD. Then we present a novel taxonomy for deep GAD methods from three new perspectives, including GNN backbone design, proxy task, and graph anomaly measures. We further deepen the discussions in each perspective by discussing more fine-grained categories of methods. Along with each fine-grained methodology category, we not only review the associated GAD methods, but also analyze their general assumption, pros and cons, and their capabilities in addressing the unique challenges in GAD. 
We lastly discuss six important directions for future research on GAD. By tackling the problems in these six directions, we expect a more advanced generation of methods for solving real-life GAD tasks. 

\section*{Acknowledgments}
This research is supported in part by the National Research Foundation, Singapore and CyberSG R\&D Programme Office under its Translation and Innovation Grants (CRPO-GC4-SMU-001), the Ministry of Education, Singapore under its Tier 1 Academic Research Fund (24-SIS-SMU-008), and the Lee Kong Chian Fellowship (T050273).

\bibliographystyle{IEEEtranS}
\bibliography{ref_trans}

% Generated by IEEEtranS.bst, version: 1.14 (2015/08/26)
\begin{thebibliography}{100}
\providecommand{\url}[1]{#1}
\csname url@samestyle\endcsname
\providecommand{\newblock}{\relax}
\providecommand{\bibinfo}[2]{#2}
\providecommand{\BIBentrySTDinterwordspacing}{\spaceskip=0pt\relax}
\providecommand{\BIBentryALTinterwordstretchfactor}{4}
\providecommand{\BIBentryALTinterwordspacing}{\spaceskip=\fontdimen2\font plus
\BIBentryALTinterwordstretchfactor\fontdimen3\font minus \fontdimen4\font\relax}
\providecommand{\BIBforeignlanguage}[2]{{%
\expandafter\ifx\csname l@#1\endcsname\relax
\typeout{** WARNING: IEEEtranS.bst: No hyphenation pattern has been}%
\typeout{** loaded for the language `#1'. Using the pattern for}%
\typeout{** the default language instead.}%
\else
\language=\csname l@#1\endcsname
\fi
#2}}
\providecommand{\BIBdecl}{\relax}
\BIBdecl

\bibitem{acsintoae2022ubnormal}
A.~Acsintoae, A.~Florescu, M.-I. Georgescu, T.~Mare, P.~Sumedrea, R.~T. Ionescu, F.~S. Khan, and M.~Shah, ``Ubnormal: New benchmark for supervised open-set video anomaly detection,'' in \emph{CVPR}, 2022, pp. 20\,143--20\,153.

\bibitem{aggarwal2014}
C.~Aggarwal and K.~Subbian, ``Evolutionary network analysis: A survey,'' \emph{ACM Computing Surveys (CSUR)}, vol.~47, no.~1, pp. 1--36, 2014.

\bibitem{aggarwal2017outlieranalysis}
C.~C. Aggarwal, \emph{Outlier analysis}.\hskip 1em plus 0.5em minus 0.4em\relax Springer, 2017.

\bibitem{akoglu2015graph}
L.~Akoglu, H.~Tong, and D.~Koutra, ``Graph based anomaly detection and description: a survey,'' \emph{Data mining and knowledge discovery}, vol.~29, pp. 626--688, 2015.

\bibitem{bandyopadhyay2020outlier}
S.~Bandyopadhyay, L.~N, S.~V. Vivek, and M.~N. Murty, ``Outlier resistant unsupervised deep architectures for attributed network embedding,'' in \emph{WSDM}, 2020, pp. 25--33.

\bibitem{bei2023reinforcement}
Y.~Bei, S.~Zhou, Q.~Tan, H.~Xu, H.~Chen, Z.~Li, and J.~Bu, ``Reinforcement neighborhood selection for unsupervised graph anomaly detection,'' in \emph{ICDM}.\hskip 1em plus 0.5em minus 0.4em\relax IEEE, 2023, pp. 11--20.

\bibitem{boukerche2020outlier}
A.~Boukerche, L.~Zheng, and O.~Alfandi, ``Outlier detection: Methods, models, and classification,'' \emph{ACM Computing Surveys (CSUR)}, vol.~53, no.~3, pp. 1--37, 2020.

\bibitem{cai2023self}
J.~Cai, Y.~Zhang, and J.~Fan, ``Self-discriminative modeling for anomalous graph detection,'' \emph{arXiv:2310.06261}, 2023.

\bibitem{cai2024fgad}
J.~Cai, Y.~Zhang, Z.~Lu, W.~Guo, and S.-k. Ng, ``Fgad: Self-boosted knowledge distillation for an effective federated graph anomaly detection framework,'' \emph{arXiv:2402.12761}, 2024.

\bibitem{chai2022can}
Z.~Chai, S.~You, Y.~Yang, S.~Pu, J.~Xu, H.~Cai, and W.~Jiang, ``Can abnormality be detected by graph neural networks,'' in \emph{IJCAI}, 2022, pp. 23--29.

\bibitem{chandola2009anomaly}
V.~Chandola, A.~Banerjee, and V.~Kumar, ``Anomaly detection: A survey,'' \emph{ACM Computing Surveys}, vol.~41, no.~3, p.~15, 2009.

\bibitem{chang2024multitask}
W.~Chang, K.~Liu, K.~Ding, P.~S. Yu, and J.~Yu, ``Multitask active learning for graph anomaly detection,'' \emph{arXiv:2401.13210}, 2024.

\bibitem{chen2022gccad}
B.~Chen, J.~Zhang, X.~Zhang, Y.~Dong, J.~Song, P.~Zhang, K.~Xu, E.~Kharlamov, and J.~Tang, ``Gccad: Graph contrastive learning for anomaly detection,'' \emph{IEEE Transactions on Knowledge and Data Engineering}, 2022.

\bibitem{chen2023consistency}
N.~Chen, Z.~Liu, B.~Hooi, B.~He, R.~Fathony, J.~Hu, and J.~Chen, ``Consistency training with learnable data augmentation for graph anomaly detection with limited supervision,'' in \emph{ICLR}, 2023.

\bibitem{chen2020generative}
Z.~Chen, B.~Liu, M.~Wang, P.~Dai, J.~Lv, and L.~Bo, ``Generative adversarial attributed network anomaly detection,'' in \emph{CIKM}, 2020, pp. 1989--1992.

\bibitem{ding2022catching}
C.~Ding, G.~Pang, and C.~Shen, ``Catching both gray and black swans: Open-set supervised anomaly detection,'' in \emph{CVPR}, 2022, pp. 7388--7398.

\bibitem{ding2021inductive}
K.~Ding, J.~Li, N.~Agarwal, and H.~Liu, ``Inductive anomaly detection on attributed networks,'' in \emph{IJCAI}, 2021, pp. 1288--1294.

\bibitem{ding2019deep}
K.~Ding, J.~Li, R.~Bhanushali, and H.~Liu, ``Deep anomaly detection on attributed networks,'' in \emph{SDM}.\hskip 1em plus 0.5em minus 0.4em\relax SIAM, 2019, pp. 594--602.

\bibitem{ding2021towards}
K.~Ding, X.~Shan, and H.~Liu, ``Towards anomaly-resistant graph neural networks via reinforcement learning,'' in \emph{CIKM}, 2021, pp. 2979--2983.

\bibitem{ding2021few}
K.~Ding, Q.~Zhou, H.~Tong, and H.~Liu, ``Few-shot network anomaly detection via cross-network meta-learning,'' in \emph{WebConf}, 2021, pp. 2448--2456.

\bibitem{dong2022bi}
L.~Dong, Y.~Liu, X.~Ao, J.~Chi, J.~Feng, H.~Yang, and Q.~He, ``Bi-level selection via meta gradient for graph-based fraud detection,'' in \emph{DASFAA}.\hskip 1em plus 0.5em minus 0.4em\relax Springer, 2022, pp. 387--394.

\bibitem{dong2024smoothgnn}
X.~Dong, X.~Zhang, Y.~Sun, L.~Chen, M.~Yuan, and S.~Wang, ``Smoothgnn: Smoothing-based gnn for unsupervised node anomaly detection,'' \emph{arXiv:2405.17525}, 2024.

\bibitem{dong2023rayleigh}
X.~Dong, X.~Zhang, and S.~Wang, ``Rayleigh quotient graph neural networks for graph-level anomaly detection,'' \emph{arXiv:2310.02861}, 2023.

\bibitem{dou2020enhancing}
Y.~Dou, Z.~Liu, L.~Sun, Y.~Deng, H.~Peng, and P.~S. Yu, ``Enhancing graph neural network-based fraud detectors against camouflaged fraudsters,'' in \emph{CIKM}, 2020, pp. 315--324.

\bibitem{duan2020aane}
D.~Duan, L.~Tong, Y.~Li, J.~Lu, L.~Shi, and C.~Zhang, ``Aane: Anomaly aware network embedding for anomalous link detection,'' in \emph{ICDM}.\hskip 1em plus 0.5em minus 0.4em\relax IEEE, 2020, pp. 1002--1007.

\bibitem{duan2023graph}
J.~Duan, S.~Wang, P.~Zhang, E.~Zhu, J.~Hu, H.~Jin, Y.~Liu, and Z.~Dong, ``Graph anomaly detection via multi-scale contrastive learning networks with augmented view,'' in \emph{AAAI}, vol.~37, no.~6, 2023, pp. 7459--7467.

\bibitem{duan2023arise}
J.~Duan, B.~Xiao, S.~Wang, H.~Zhou, and X.~Liu, ``Arise: Graph anomaly detection on attributed networks via substructure awareness,'' \emph{IEEE transactions on neural networks and learning systems}, 2023.

\bibitem{duan2023normality}
J.~Duan, P.~Zhang, S.~Wang, J.~Hu, H.~Jin, J.~Zhang, H.~Zhou, and X.~Liu, ``Normality learning-based graph anomaly detection via multi-scale contrastive learning,'' in \emph{ACM MM}, 2023, pp. 7502--7511.

\bibitem{ester1996density}
M.~Ester, H.-P. Kriegel, J.~Sander, X.~Xu \emph{et~al.}, ``A density-based algorithm for discovering clusters in large spatial databases with noise,'' in \emph{kdd}, vol.~96, no.~34, 1996, pp. 226--231.

\bibitem{fan2020anomalydae}
H.~Fan, F.~Zhang, and Z.~Li, ``Anomalydae: Dual autoencoder for anomaly detection on attributed networks,'' in \emph{ICASSP}.\hskip 1em plus 0.5em minus 0.4em\relax IEEE, 2020, pp. 5685--5689.

\bibitem{fan2022debiased}
S.~Fan, X.~Wang, C.~Shi, K.~Kuang, N.~Liu, and B.~Wang, ``Debiased graph neural networks with agnostic label selection bias,'' \emph{IEEE transactions on neural networks and learning systems}, 2022.

\bibitem{fernandez2018smote}
A.~Fern{\'a}ndez, S.~Garcia, F.~Herrera, and N.~V. Chawla, ``Smote for learning from imbalanced data: progress and challenges, marking the 15-year anniversary,'' \emph{Journal of artificial intelligence research}, vol.~61, pp. 863--905, 2018.

\bibitem{gao2024graph}
Y.~Gao, J.~Fang, Y.~Sui, Y.~Li, X.~Wang, H.~Feng, and Y.~Zhang, ``Graph anomaly detection with bi-level optimization,'' in \emph{Proceedings of the ACM on Web Conference 2024}, 2024, pp. 4383--4394.

\bibitem{gao2023alleviating}
Y.~Gao, X.~Wang, X.~He \emph{et~al.}, ``Alleviating structural distribution shift in graph anomaly detection,'' in \emph{WSDM}, 2023, pp. 357--365.

\bibitem{gao2023addressing}
Y.~Gao, X.~Wang, X.~He, Z.~Liu, H.~Feng, and Y.~Zhang, ``Addressing heterophily in graph anomaly detection: A perspective of graph spectrum,'' in \emph{WebConf}, 2023, pp. 1528--1538.

\bibitem{gong2023beyond}
Z.~Gong, G.~Wang, Y.~Sun, Q.~Liu, Y.~Ning, H.~Xiong, and J.~Peng, ``Beyond homophily: Robust graph anomaly detection via neural sparsification,'' in \emph{IJCAI}, 2023, pp. 2104--2113.

\bibitem{gou2021knowledge}
J.~Gou, B.~Yu, S.~J. Maybank, and D.~Tao, ``Knowledge distillation: A survey,'' \emph{International Journal of Computer Vision}, vol. 129, no.~6, pp. 1789--1819, 2021.

\bibitem{guo2024graph}
R.~Guo, M.~Zou, S.~Zhang, X.~Zhang, Z.~Yu, and Z.~Feng, ``Graph local homophily network for anomaly detection,'' in \emph{CIKM}, 2024, pp. 706--716.

\bibitem{hamilton2017inductive}
W.~Hamilton, Z.~Ying, and J.~Leskovec, ``Inductive representation learning on large graphs,'' \emph{NeurIPS}, vol.~30, 2017.

\bibitem{hammond2011wavelets}
D.~K. Hammond, P.~Vandergheynst, and R.~Gribonval, ``Wavelets on graphs via spectral graph theory,'' \emph{Applied and Computational Harmonic Analysis}, vol.~30, no.~2, pp. 129--150, 2011.

\bibitem{hinton2006reducing}
G.~E. Hinton and R.~R. Salakhutdinov, ``Reducing the dimensionality of data with neural networks,'' \emph{science}, vol. 313, no. 5786, pp. 504--507, 2006.

\bibitem{ho2020denoising}
J.~Ho, A.~Jain, and P.~Abbeel, ``Denoising diffusion probabilistic models,'' \emph{NeurIPS}, vol.~33, pp. 6840--6851, 2020.

\bibitem{hu2021ogb}
W.~Hu, M.~Fey, H.~Ren, M.~Nakata, Y.~Dong, and J.~Leskovec, ``Ogb-lsc: A large-scale challenge for machine learning on graphs,'' \emph{arXiv:2103.09430}, 2021.

\bibitem{hu2020open}
W.~Hu, M.~Fey, M.~Zitnik, Y.~Dong, H.~Ren, B.~Liu, M.~Catasta, and J.~Leskovec, ``Open graph benchmark: Datasets for machine learning on graphs,'' in \emph{NeurIPS}, vol.~33, 2020, pp. 22\,118--22\,133.

\bibitem{huang2021hybrid}
L.~Huang, Y.~Zhu, Y.~Gao, T.~Liu, C.~Chang, C.~Liu, Y.~Tang, and C.-D. Wang, ``Hybrid-order anomaly detection on attributed networks,'' \emph{IEEE Transactions on Knowledge and Data Engineering}, 2021.

\bibitem{huang2022auc}
M.~Huang, Y.~Liu, X.~Ao, K.~Li, J.~Chi, J.~Feng, H.~Yang, and Q.~He, ``Auc-oriented graph neural network for fraud detection,'' in \emph{WebConf}, 2022, pp. 1311--1321.

\bibitem{huang2022hop}
T.~Huang, Y.~Pei, V.~Menkovski, and M.~Pechenizkiy, ``Hop-count based self-supervised anomaly detection on attributed networks,'' in \emph{ECMLPKDD}.\hskip 1em plus 0.5em minus 0.4em\relax Springer, 2022, pp. 225--241.

\bibitem{huang2022dgraph}
X.~Huang, Y.~Yang, Y.~Wang, C.~Wang, Z.~Zhang, J.~Xu, L.~Chen, and M.~Vazirgiannis, ``Dgraph: A large-scale financial dataset for graph anomaly detection,'' \emph{NeurIPS}, vol.~35, pp. 22\,765--22\,777, 2022.

\bibitem{huang2022we}
Y.~Huang, L.~Wang, F.~Zhang, and X.~Lin, ``Are we really making much progress in unsupervised graph outlier detection? revisiting the problem with new insight and superior method,'' \emph{arXiv:2210.12941}, 2022.

\bibitem{hyun2024lex}
W.~Hyun, I.~Lee, and B.~Suh, ``Lex-gnn: Label-exploring graph neural network for accurate fraud detection,'' in \emph{Proceedings of the 33rd ACM International Conference on Information and Knowledge Management}, 2024, pp. 3802--3806.

\bibitem{jeong2023winclip}
J.~Jeong, Y.~Zou, T.~Kim, D.~Zhang, A.~Ravichandran, and O.~Dabeer, ``Winclip: Zero-/few-shot anomaly classification and segmentation,'' in \emph{CVPR}, 2023, pp. 19\,606--19\,616.

\bibitem{jin2021anemone}
M.~Jin, Y.~Liu, Y.~Zheng, L.~Chi, Y.-F. Li, and S.~Pan, ``Anemone: Graph anomaly detection with multi-scale contrastive learning,'' in \emph{CIKM}, 2021, pp. 3122--3126.

\bibitem{kim2023deep}
H.~Kim, J.~Kim, B.~S. Lee, and S.~Lim, ``Deep semi-supervised anomaly detection with metapath-based context knowledge,'' \emph{arXiv:2308.10918}, 2023.

\bibitem{kim2023class}
J.~Kim, Y.~In, K.~Yoon, J.~Lee, and C.~Park, ``Class label-aware graph anomaly detection,'' in \emph{CIKM}, 2023, pp. 4008--4012.

\bibitem{kipf2016semi}
T.~N. Kipf and M.~Welling, ``Semi-supervised classification with graph convolutional networks,'' \emph{arXiv:1609.02907}, 2016.

\bibitem{kong2024federated}
X.~Kong, W.~Zhang, H.~Wang, M.~Hou, X.~Chen, X.~Yan, and S.~K. Das, ``Federated graph anomaly detection via contrastive self-supervised learning,'' \emph{IEEE Transactions on Neural Networks and Learning Systems}, 2024.

\bibitem{li2023hrgcn}
J.~Li, G.~Pang, L.~Chen, and M.-R. Namazi-Rad, ``Hrgcn: Heterogeneous graph-level anomaly detection with hierarchical relation-augmented graph neural networks,'' in \emph{DSAA}.\hskip 1em plus 0.5em minus 0.4em\relax IEEE, 2023, pp. 1--10.

\bibitem{li2017feature}
J.~Li, K.~Cheng, S.~Wang, F.~Morstatter, R.~P. Trevino, J.~Tang, and H.~Liu, ``Feature selection: A data perspective,'' \emph{ACM computing surveys (CSUR)}, vol.~50, no.~6, pp. 1--45, 2017.

\bibitem{li2019specae}
Y.~Li, X.~Huang, J.~Li, M.~Du, and N.~Zou, ``Specae: Spectral autoencoder for anomaly detection in attributed networks,'' in \emph{CIKM}, 2019, pp. 2233--2236.

\bibitem{lin2023discriminative}
F.~Lin, X.~Luo, J.~Wu, J.~Yang, S.~Xue, Z.~Wang, and H.~Gong, ``Discriminative graph-level anomaly detection via dual-students-teacher model,'' in \emph{ADMA}.\hskip 1em plus 0.5em minus 0.4em\relax Springer, 2023, pp. 261--276.

\bibitem{liu2022dagad}
F.~Liu, X.~Ma, J.~Wu, J.~Yang, S.~Xue, A.~Beheshti, C.~Zhou, H.~Peng, Q.~Z. Sheng, and C.~C. Aggarwal, ``Dagad: Data augmentation for graph anomaly detection,'' in \emph{ICDM}.\hskip 1em plus 0.5em minus 0.4em\relax IEEE, 2022, pp. 259--268.

\bibitem{liu2008isolation}
F.~T. Liu, K.~M. Ting, and Z.-H. Zhou, ``Isolation forest,'' in \emph{ICDM}.\hskip 1em plus 0.5em minus 0.4em\relax IEEE, 2008, pp. 413--422.

\bibitem{liu2023towards}
J.~Liu, C.~Yang, Z.~Lu, J.~Chen, Y.~Li, M.~Zhang, T.~Bai, Y.~Fang, L.~Sun, P.~S. Yu \emph{et~al.}, ``Towards graph foundation models: A survey and beyond,'' \emph{arXiv:2310.11829}, 2023.

\bibitem{liu2024spatial}
J.~Liu, X.~Shang, X.~Han, W.~Zhang, and H.~Yin, ``Spatial-temporal memories enhanced graph autoencoder for anomaly detection in dynamic graphs,'' \emph{arXiv:2403.09039}, 2024.

\bibitem{liu2022bond}
K.~Liu, Y.~Dou, Y.~Zhao, X.~Ding, X.~Hu, R.~Zhang, K.~Ding, C.~Chen, H.~Peng, K.~Shu \emph{et~al.}, ``Bond: Benchmarking unsupervised outlier node detection on static attributed graphs,'' \emph{NeurIPS}, vol.~35, pp. 27\,021--27\,035, 2022.

\bibitem{liu2023data}
K.~Liu, H.~Zhang, Z.~Hu, F.~Wang, and P.~S. Yu, ``Data augmentation for supervised graph outlier detection with latent diffusion models,'' \emph{arXiv:2312.17679}, 2023.

\bibitem{liu2021pick}
Y.~Liu, X.~Ao, Z.~Qin, J.~Chi, J.~Feng, H.~Yang, and Q.~He, ``Pick and choose: a gnn-based imbalanced learning approach for fraud detection,'' in \emph{WebConf}, 2021, pp. 3168--3177.

\bibitem{liu2024towards}
Y.~Liu, K.~Ding, Q.~Lu, F.~Li, L.~Y. Zhang, and S.~Pan, ``Towards self-interpretable graph-level anomaly detection,'' \emph{NeurIPS}, vol.~36, 2024.

\bibitem{liu2024arc}
Y.~Liu, S.~Li, Y.~Zheng, Q.~Chen, C.~Zhang, and S.~Pan, ``Arc: A generalist graph anomaly detector with in-context learning,'' \emph{arXiv:2405.16771}, 2024.

\bibitem{liu2021cola}
Y.~Liu, Z.~Li, S.~Pan, C.~Gong, C.~Zhou, and G.~Karypis, ``Anomaly detection on attributed networks via contrastive self-supervised learning,'' \emph{IEEE transactions on neural networks and learning systems}, vol.~33, no.~6, pp. 2378--2392, 2021.

\bibitem{liu2021anomaly}
Y.~Liu, S.~Pan, Y.~G. Wang, F.~Xiong, L.~Wang, Q.~Chen, and V.~C. Lee, ``Anomaly detection in dynamic graphs via transformer,'' \emph{IEEE Transactions on Knowledge and Data Engineering}, vol.~35, no.~12, pp. 12\,081--12\,094, 2021.

\bibitem{liu2023survey}
Z.~Liu, Y.~Li, N.~Chen, Q.~Wang, B.~Hooi, and B.~He, ``A survey of imbalanced learning on graphs: Problems, techniques, and future directions,'' \emph{arXiv:2308.13821}, 2023.

\bibitem{liu2023topological}
Z.~Liu, Z.~Zeng, R.~Qiu, H.~Yoo, D.~Zhou, Z.~Xu, Y.~Zhu, K.~Weldemariam, J.~He, and H.~Tong, ``Topological augmentation for class-imbalanced node classification,'' \emph{arXiv:2308.14181}, 2023.

\bibitem{liu2020alleviating}
Z.~Liu, Y.~Dou, P.~S. Yu, Y.~Deng, and H.~Peng, ``Alleviating the inconsistency problem of applying graph neural network to fraud detection,'' in \emph{SIGIR}, 2020, pp. 1569--1572.

\bibitem{liu2022mul}
Z.~Liu, C.~Cao, and J.~Sun, ``Mul-gad: a semi-supervised graph anomaly detection framework via aggregating multi-view information,'' \emph{arXiv:2212.05478}, 2022.

\bibitem{liu2023revisiting}
Z.~Liu, C.~Cao, F.~Tao, and J.~Sun, ``Revisiting graph contrastive learning for anomaly detection,'' \emph{arXiv:2305.02496}, 2023.

\bibitem{lou2023gady}
S.~Lou, Q.~Zhang, S.~Yang, Y.~Tian, Z.~Tan, and M.~Luo, ``Gady: Unsupervised anomaly detection on dynamic graphs,'' \emph{arXiv:2310.16376}, 2023.

\bibitem{luo2022comga}
X.~Luo, J.~Wu, A.~Beheshti, J.~Yang, X.~Zhang, Y.~Wang, and S.~Xue, ``Comga: Community-aware attributed graph anomaly detection,'' in \emph{WSDM}, 2022, pp. 657--665.

\bibitem{ma2019disentangled}
J.~Ma, P.~Cui, K.~Kuang, X.~Wang, and W.~Zhu, ``Disentangled graph convolutional networks,'' in \emph{ICML}.\hskip 1em plus 0.5em minus 0.4em\relax PMLR, 2019, pp. 4212--4221.

\bibitem{ma2022deep}
R.~Ma, G.~Pang, L.~Chen, and A.~van~den Hengel, ``Deep graph-level anomaly detection by glocal knowledge distillation,'' in \emph{WSDM}, 2022, pp. 704--714.

\bibitem{ma2023new}
X.~Ma, R.~Li, F.~Liu, K.~Ding, J.~Yang, and J.~Wu, ``New recipes for graph anomaly detection: Forward diffusion dynamics and graph generation,'' 2023.

\bibitem{ma2021comprehensive}
X.~Ma, J.~Wu, S.~Xue, J.~Yang, C.~Zhou, Q.~Z. Sheng, H.~Xiong, and L.~Akoglu, ``A comprehensive survey on graph anomaly detection with deep learning,'' \emph{IEEE Transactions on Knowledge and Data Engineering}, 2021.

\bibitem{ma2023towards}
X.~Ma, J.~Wu, J.~Yang, and Q.~Z. Sheng, ``Towards graph-level anomaly detection via deep evolutionary mapping,'' in \emph{KDD}, 2023, pp. 1631--1642.

\bibitem{meng2023generative}
L.~Meng, H.~Mostafa, M.~Nassar, X.~Zhang, and J.~Zhang, ``Generative graph augmentation for minority class in fraud detection,'' in \emph{CIKM}, 2023, pp. 4200--4204.

\bibitem{niu2023graph}
C.~Niu, G.~Pang, and L.~Chen, ``Graph-level anomaly detection via hierarchical memory networks,'' in \emph{ECMLPKDD}.\hskip 1em plus 0.5em minus 0.4em\relax Springer, 2023, pp. 201--218.

\bibitem{niu2024zero}
C.~Niu, H.~Qiao, C.~Chen, L.~Chen, and G.~Pang, ``Zero-shot generalist graph anomaly detection with unified neighborhood prompts,'' \emph{arXiv preprint arXiv:2410.14886}, 2024.

\bibitem{pan2023prem}
J.~Pan, Y.~Liu, Y.~Zheng, and S.~Pan, ``Prem: A simple yet effective approach for node-level graph anomaly detection,'' \emph{arXiv:2310.11676}, 2023.

\bibitem{pang2018sparse}
G.~Pang, L.~Cao, L.~Chen, D.~Lian, and H.~Liu, ``Sparse modeling-based sequential ensemble learning for effective outlier detection in high-dimensional numeric data,'' in \emph{AAAI}, vol.~32, no.~1, 2018.

\bibitem{pang2016unsupervised}
G.~Pang, L.~Cao, L.~Chen, and H.~Liu, ``Unsupervised feature selection for outlier detection by modelling hierarchical value-feature couplings,'' in \emph{ICDM}.\hskip 1em plus 0.5em minus 0.4em\relax IEEE, 2016, pp. 410--419.

\bibitem{pang2017learning}
G.~Pang, L.~Cao~Longbing, Chen, and H.~Liu, ``Learning homophily couplings from non-iid data for joint feature selection and noise-resilient outlier detection,'' in \emph{IJCAI}, 2017, pp. 2585--2591.

\bibitem{pang2021deep}
G.~Pang, C.~Shen, L.~Cao, and A.~V.~D. Hengel, ``Deep learning for anomaly detection: A review,'' \emph{ACM computing surveys (CSUR)}, vol.~54, no.~2, pp. 1--38, 2021.

\bibitem{pang2023deep}
G.~Pang, C.~Shen, H.~Jin, and A.~van~den Hengel, ``Deep weakly-supervised anomaly detection,'' in \emph{KDD}, 2023, pp. 1795--1807.

\bibitem{pang2019deep}
G.~Pang, C.~Shen, and A.~van~den Hengel, ``Deep anomaly detection with deviation networks,'' in \emph{KDD}, 2019, pp. 353--362.

\bibitem{pang2021toward}
G.~Pang, A.~van~den Hengel, C.~Shen, and L.~Cao, ``Toward deep supervised anomaly detection: Reinforcement learning from partially labeled anomaly data,'' in \emph{KDD}, 2021, pp. 1298--1308.

\bibitem{pang2017selective}
G.~Pang, H.~Xu, L.~Cao, and W.~Zhao, ``Selective value coupling learning for detecting outliers in high-dimensional categorical data,'' in \emph{CIKM}, 2017, pp. 807--816.

\bibitem{park2021graphens}
J.~Park, J.~Song, and E.~Yang, ``Graphens: Neighbor-aware ego network synthesis for class-imbalanced node classification,'' in \emph{ICLR}, 2021.

\bibitem{pei2022resgcn}
Y.~Pei, T.~Huang, W.~van Ipenburg, and M.~Pechenizkiy, ``Resgcn: attention-based deep residual modeling for anomaly detection on attributed networks,'' \emph{Machine Learning}, vol. 111, no.~2, pp. 519--541, 2022.

\bibitem{peng2020deep}
Z.~Peng, M.~Luo, J.~Li, L.~Xue, and Q.~Zheng, ``A deep multi-view framework for anomaly detection on attributed networks,'' \emph{IEEE Transactions on Knowledge and Data Engineering}, vol.~34, no.~6, pp. 2539--2552, 2020.

\bibitem{pourhabibi2020fraud}
T.~Pourhabibi, K.-L. Ong, B.~H. Kam, and Y.~L. Boo, ``Fraud detection: A systematic literature review of graph-based anomaly detection approaches,'' \emph{Decision Support Systems}, vol. 133, p. 113303, 2020.

\bibitem{qiao2025anomalygfm}
H.~Qiao, C.~Niu, L.~Chen, and G.~Pang, ``Anomalygfm: Graph foundation model for zero/few-shot anomaly detection,'' \emph{arXiv preprint arXiv:2502.09254}, 2025.

\bibitem{hezhe2023truncated}
H.~Qiao and G.~Pang, ``Truncated affinity maximization: One-class homophily modeling for graph anomaly detection,'' \emph{NeurIPS}, vol.~36, 2024.

\bibitem{qiao2024generative}
H.~Qiao, Q.~Wen, X.~Li, E.-P. Lim, and G.~Pang, ``Generative semi-supervised graph anomaly detection,'' in \emph{NeurIPS}, 2024.

\bibitem{qin2022explainable}
Z.~Qin, Y.~Liu, Q.~He, and X.~Ao, ``Explainable graph-based fraud detection via neural meta-graph search,'' in \emph{CIKM}, 2022, pp. 4414--4418.

\bibitem{qiu2022raising}
C.~Qiu, M.~Kloft, S.~Mandt, and M.~Rudolph, ``Raising the bar in graph-level anomaly detection,'' \emph{arXiv:2205.13845}, 2022.

\bibitem{qu2021imgagn}
L.~Qu, H.~Zhu, R.~Zheng, Y.~Shi, and H.~Yin, ``Imgagn: Imbalanced network embedding via generative adversarial graph networks,'' in \emph{KDD}, 2021, pp. 1390--1398.

\bibitem{ranshous2015anomaly}
S.~Ranshous, S.~Shen, D.~Koutra, S.~Harenberg, C.~Faloutsos, and N.~F. Samatova, ``Anomaly detection in dynamic networks: a survey,'' \emph{Wiley Interdisciplinary Reviews: Computational Statistics}, vol.~7, no.~3, pp. 223--247, 2015.

\bibitem{roy2024gad}
A.~Roy, J.~Shu, J.~Li, C.~Yang, O.~Elshocht, J.~Smeets, and P.~Li, ``Gad-nr: Graph anomaly detection via neighborhood reconstruction,'' in \emph{WSDM}, 2024, pp. 576--585.

\bibitem{ruff2018deep}
L.~Ruff, R.~Vandermeulen, N.~Goernitz, L.~Deecke, S.~A. Siddiqui, A.~Binder, E.~M{\"u}ller, and M.~Kloft, ``Deep one-class classification,'' in \emph{ICML}.\hskip 1em plus 0.5em minus 0.4em\relax PMLR, 2018, pp. 4393--4402.

\bibitem{sanchez2020evaluating}
B.~Sanchez-Lengeling, J.~Wei, B.~Lee, E.~Reif, P.~Wang, W.~Qian, K.~McCloskey, L.~Colwell, and A.~Wiltschko, ``Evaluating attribution for graph neural networks,'' in \emph{NeurIPS}, vol.~33, 2020, pp. 5898--5910.

\bibitem{sato2023prompt}
F.~Sato, R.~Hachiuma, and T.~Sekii, ``Prompt-guided zero-shot anomaly action recognition using pretrained deep skeleton features,'' in \emph{CVPR}, 2023, pp. 6471--6480.

\bibitem{shi2022h2}
F.~Shi, Y.~Cao, Y.~Shang, Y.~Zhou, C.~Zhou, and J.~Wu, ``H2-fdetector: A gnn-based fraud detector with homophilic and heterophilic connections,'' in \emph{WebConf}, 2022, pp. 1486--1494.

\bibitem{shi2020multi}
M.~Shi, Y.~Tang, X.~Zhu, D.~Wilson, and J.~Liu, ``Multi-class imbalanced graph convolutional network learning,'' in \emph{IJCAI}, 2020.

\bibitem{shuman2013emerging}
D.~I. Shuman, S.~K. Narang, P.~Frossard, A.~Ortega, and P.~Vandergheynst, ``The emerging field of signal processing on graphs: Extending high-dimensional data analysis to networks and other irregular domains,'' \emph{IEEE signal processing magazine}, vol.~30, no.~3, pp. 83--98, 2013.

\bibitem{sun2023all}
X.~Sun, H.~Cheng, J.~Li, B.~Liu, and J.~Guan, ``All in one: Multi-task prompting for graph neural networks,'' in \emph{KDD}, 2023, pp. 2120--2131.

\bibitem{tang2024higpt}
J.~Tang, Y.~Yang, W.~Wei, L.~Shi, L.~Xia, D.~Yin, and C.~Huang, ``Higpt: Heterogeneous graph language model,'' \emph{arXiv:2402.16024}, 2024.

\bibitem{tang2023gadbench}
J.~Tang, F.~Hua, Z.~Gao, P.~Zhao, and J.~Li, ``Gadbench: Revisiting and benchmarking supervised graph anomaly detection,'' \emph{arXiv:2306.12251}, 2023.

\bibitem{tang2022rethinking}
J.~Tang, J.~Li, Z.~Gao, and J.~Li, ``Rethinking graph neural networks for anomaly detection,'' in \emph{ICML}.\hskip 1em plus 0.5em minus 0.4em\relax PMLR, 2022, pp. 21\,076--21\,089.

\bibitem{teng2018deep}
X.~Teng, M.~Yan, A.~M. Ertugrul, and Y.-R. Lin, ``Deep into hypersphere: Robust and unsupervised anomaly discovery in dynamic networks,'' in \emph{IJCAI}, 2018.

\bibitem{tian2023sad}
S.~Tian, J.~Dong, J.~Li, W.~Zhao, X.~Xu, B.~Song, C.~Meng, T.~Zhang, L.~Chen \emph{et~al.}, ``Sad: Semi-supervised anomaly detection on dynamic graphs,'' \emph{arXiv:2305.13573}, 2023.

\bibitem{velivckovic2017graph}
P.~Veli{\v{c}}kovi{\'c}, G.~Cucurull, A.~Casanova, A.~Romero, P.~Lio, and Y.~Bengio, ``Graph attention networks,'' \emph{arXiv:1710.10903}, 2017.

\bibitem{velivckovic2018deep}
P.~Veli{\v{c}}kovi{\'c}, W.~Fedus, W.~L. Hamilton, P.~Li{\`o}, Y.~Bengio, and R.~D. Hjelm, ``Deep graph infomax,'' \emph{arXiv:1809.10341}, 2018.

\bibitem{wang2023cross}
Q.~Wang, G.~Pang, M.~Salehi, W.~Buntine, and C.~Leckie, ``Cross-domain graph anomaly detection via anomaly-aware contrastive alignment,'' in \emph{AAAI}, vol.~37, no.~4, 2023, pp. 4676--4684.

\bibitem{wang2023open}
Q.~Wang, G.~Pang, M.~Salehi \emph{et~al.}, ``Open-set graph anomaly detection via normal structure regularisation,'' in \emph{ICLR}, 2025.

\bibitem{wang2021one}
X.~Wang, B.~Jin, Y.~Du, P.~Cui, Y.~Tan, and Y.~Yang, ``One-class graph neural networks for anomaly detection in attributed networks,'' \emph{Neural computing and applications}, vol.~33, pp. 12\,073--12\,085, 2021.

\bibitem{wang2021decoupling}
Y.~Wang, J.~Zhang, S.~Guo, H.~Yin, C.~Li, and H.~Chen, ``Decoupling representation learning and classification for gnn-based anomaly detection,'' in \emph{SIGIR}, 2021, pp. 1239--1248.

\bibitem{wang2023label}
Y.~Wang, J.~Zhang, Z.~Huang, W.~Li, S.~Feng, Z.~Ma, Y.~Sun, D.~Yu, F.~Dong, J.~Jin \emph{et~al.}, ``Label information enhanced fraud detection against low homophily in graphs,'' in \emph{WebConf}, 2023, pp. 406--416.

\bibitem{wu2023splitgnn}
B.~Wu, X.~Yao, B.~Zhang, K.-M. Chao, and Y.~Li, ``Splitgnn: Spectral graph neural network for fraud detection against heterophily,'' in \emph{CIKM}, 2023, pp. 2737--2746.

\bibitem{wu2022graphmixup}
L.~Wu, J.~Xia, Z.~Gao, H.~Lin, C.~Tan, and S.~Z. Li, ``Graphmixup: Improving class-imbalanced node classification by reinforcement mixup and self-supervised context prediction,'' in \emph{ECMLPKDD}.\hskip 1em plus 0.5em minus 0.4em\relax Springer, 2022, pp. 519--535.

\bibitem{wu2024open}
P.~Wu, X.~Zhou, G.~Pang, Y.~Sun, J.~Liu, P.~Wang, and Y.~Zhang, ``Open-vocabulary video anomaly detection,'' in \emph{CVPR}, 2024, pp. 18\,297--18\,307.

\bibitem{wu2024weakly}
P.~Wu, X.~Zhou, G.~Pang, Z.~Yang, Q.~Yan, P.~WANG, and Y.~Zhang, ``Weakly supervised video anomaly detection and localization with spatio-temporal prompts,'' in \emph{ACM MM}, 2024.

\bibitem{wu2024vadclip}
P.~Wu, X.~Zhou, G.~Pang, L.~Zhou, Q.~Yan, P.~Wang, and Y.~Zhang, ``Vadclip: Adapting vision-language models for weakly supervised video anomaly detection,'' in \emph{AAAI}, vol.~38, no.~6, 2024, pp. 6074--6082.

\bibitem{xia2024opengraph}
L.~Xia, B.~Kao, and C.~Huang, ``Opengraph: Towards open graph foundation models,'' \emph{arXiv:2403.01121}, 2024.

\bibitem{xiao2023counterfactual}
C.~Xiao, X.~Xu, Y.~Lei, K.~Zhang, S.~Liu, and F.~Zhou, ``Counterfactual graph learning for anomaly detection on attributed networks,'' \emph{IEEE Transactions on Knowledge and Data Engineering}, 2023.

\bibitem{xu2023few}
F.~Xu, N.~Wang, X.~Wen, M.~Gao, C.~Guo, and X.~Zhao, ``Few-shot message-enhanced contrastive learning for graph anomaly detection,'' \emph{arXiv:2311.10370}, 2023.

\bibitem{xu2023revisiting}
F.~Xu, N.~Wang, H.~Wu, X.~Wen, and X.~Zhao, ``Revisiting graph-based fraud detection in sight of heterophily and spectrum,'' \emph{arXiv:2312.06441}, 2023.

\bibitem{xu2023deep}
H.~Xu, G.~Pang, Y.~Wang, and Y.~Wang, ``Deep isolation forest for anomaly detection,'' \emph{IEEE Transactions on Knowledge and Data Engineering}, 2023.

\bibitem{xu2022contrastive}
Z.~Xu, X.~Huang, Y.~Zhao, Y.~Dong, and J.~Li, ``Contrastive attributed network anomaly detection with data augmentation,'' in \emph{PAKDD}.\hskip 1em plus 0.5em minus 0.4em\relax Springer, 2022, pp. 444--457.

\bibitem{yao2023explicit}
X.~Yao, R.~Li, J.~Zhang, J.~Sun, and C.~Zhang, ``Explicit boundary guided semi-push-pull contrastive learning for supervised anomaly detection,'' in \emph{CVPR}, 2023, pp. 24\,490--24\,499.

\bibitem{you2020graph}
Y.~You, T.~Chen, Y.~Sui, T.~Chen, Z.~Wang, and Y.~Shen, ``Graph contrastive learning with augmentations,'' \emph{NeurIPS}, vol.~33, pp. 5812--5823, 2020.

\bibitem{yu2024barely}
H.~Yu, Z.~Liu, and X.~Luo, ``Barely supervised learning for graph-based fraud detection,'' in \emph{AAAI}, vol.~38, no.~15, 2024, pp. 16\,548--16\,557.

\bibitem{yu2016survey}
R.~Yu, H.~Qiu, Z.~Wen, C.~Lin, and Y.~Liu, ``A survey on social media anomaly detection,'' \emph{ACM SIGKDD Explorations Newsletter}, vol.~18, no.~1, pp. 1--14, 2016.

\bibitem{yu2018netwalk}
W.~Yu, W.~Cheng, C.~C. Aggarwal, K.~Zhang, H.~Chen, and W.~Wang, ``Netwalk: A flexible deep embedding approach for anomaly detection in dynamic networks,'' in \emph{KDD}, 2018, pp. 2672--2681.

\bibitem{yuan2021higher}
X.~Yuan, N.~Zhou, S.~Yu, H.~Huang, Z.~Chen, and F.~Xia, ``Higher-order structure based anomaly detection on attributed networks,'' in \emph{BigData}.\hskip 1em plus 0.5em minus 0.4em\relax IEEE, 2021, pp. 2691--2700.

\bibitem{zhang2021fraudre}
G.~Zhang, J.~Wu, J.~Yang, A.~Beheshti, S.~Xue, C.~Zhou, and Q.~Z. Sheng, ``Fraudre: Fraud detection dual-resistant to graph inconsistency and imbalance,'' in \emph{ICDM}.\hskip 1em plus 0.5em minus 0.4em\relax IEEE, 2021, pp. 867--876.

\bibitem{zhang2022dual}
G.~Zhang, Z.~Yang, J.~Wu, J.~Yang, S.~Xue, H.~Peng, J.~Su, C.~Zhou, Q.~Z. Sheng, L.~Akoglu \emph{et~al.}, ``Dual-discriminative graph neural network for imbalanced graph-level anomaly detection,'' \emph{NeurIPS}, vol.~35, pp. 24\,144--24\,157, 2022.

\bibitem{zhang2017mixup}
H.~Zhang, M.~Cisse, Y.~N. Dauphin, and D.~Lopez-Paz, ``mixup: Beyond empirical risk minimization,'' \emph{arXiv:1710.09412}, 2017.

\bibitem{zhang2022reconstruction}
J.~Zhang, S.~Wang, and S.~Chen, ``Reconstruction enhanced multi-view contrastive learning for anomaly detection on attributed networks,'' \emph{arXiv:2205.04816}, 2022.

\bibitem{zhang2019robust}
L.~Zhang, J.~Yuan, Z.~Liu, Y.~Pei, and L.~Wang, ``A robust embedding method for anomaly detection on attributed networks,'' in \emph{IJCNN}.\hskip 1em plus 0.5em minus 0.4em\relax IEEE, 2019, pp. 1--8.

\bibitem{zhang2024generation}
R.~Zhang, D.~Cheng, X.~Liu, J.~Yang, Y.~Ouyang, X.~Wu, and Y.~Zheng, ``Generation is better than modification: Combating high class homophily variance in graph anomaly detection,'' \emph{arXiv:2403.10339}, 2024.

\bibitem{zhang2023deep}
Y.~Zhang, Y.~Sun, J.~Cai, and J.~Fan, ``Deep graph-level orthogonal hypersphere compression for anomaly detection,'' \emph{arXiv:2302.06430}, 2023.

\bibitem{zhao2023using}
L.~Zhao and L.~Akoglu, ``On using classification datasets to evaluate graph outlier detection: Peculiar observations and new insights,'' \emph{Big Data}, vol.~11, no.~3, pp. 151--180, 2023.

\bibitem{zhao2021graphsmote}
T.~Zhao, X.~Zhang, and S.~Wang, ``Graphsmote: Imbalanced node classification on graphs with graph neural networks,'' in \emph{WSDM}, 2021, pp. 833--841.

\bibitem{zhao2021action}
T.~Zhao, B.~Ni, W.~Yu, Z.~Guo, N.~Shah, and M.~Jiang, ``Action sequence augmentation for early graph-based anomaly detection,'' in \emph{CIKM}, 2021, pp. 2668--2678.

\bibitem{zheng2019one}
P.~Zheng, S.~Yuan, X.~Wu, J.~Li, and A.~Lu, ``One-class adversarial nets for fraud detection,'' in \emph{AAAI}, vol.~33, no.~01, 2019, pp. 1286--1293.

\bibitem{zheng2022graph}
X.~Zheng, Y.~Wang, Y.~Liu, M.~Li, M.~Zhang, D.~Jin, P.~S. Yu, and S.~Pan, ``Graph neural networks for graphs with heterophily: A survey,'' \emph{arXiv:2202.07082}, 2022.

\bibitem{zheng2021generative}
Y.~Zheng, M.~Jin, Y.~Liu, L.~Chi, K.~T. Phan, and Y.-P.~P. Chen, ``Generative and contrastive self-supervised learning for graph anomaly detection,'' \emph{IEEE Transactions on Knowledge and Data Engineering}, 2021.

\bibitem{zhou2023anomalyclip}
Q.~Zhou, G.~Pang, Y.~Tian, S.~He, and J.~Chen, ``Anomalyclip: Object-agnostic prompt learning for zero-shot anomaly detection,'' in \emph{ICLR}, 2024.

\bibitem{zhou2024graph}
Q.~Zhou, Y.~Chen, Z.~Xu, Y.~Wu, M.~Pan, M.~Das, H.~Yang, and H.~Tong, ``Graph anomaly detection with adaptive node mixup,'' in \emph{Proceedings of the CIKM}, 2024, pp. 3494--3504.

\bibitem{zhou2023learning}
Q.~Zhou, K.~Ding, H.~Liu, and H.~Tong, ``Learning node abnormality with weak supervision,'' in \emph{CIKM}, 2023, pp. 3584--3594.

\bibitem{zhou2022unseen}
S.~Zhou, X.~Huang, N.~Liu, Q.~Tan, and F.-L. Chung, ``Unseen anomaly detection on networks via multi-hypersphere learning,'' in \emph{SDM}.\hskip 1em plus 0.5em minus 0.4em\relax SIAM, 2022, pp. 262--270.

\bibitem{zhou2023improving}
S.~Zhou, X.~Huang, N.~Liu, H.~Zhou, F.-L. Chung, and L.-K. Huang, ``Improving generalizability of graph anomaly detection models via data augmentation,'' \emph{IEEE Transactions on Knowledge and Data Engineering}, 2023.

\bibitem{zhou2021subtractive}
S.~Zhou, Q.~Tan, Z.~Xu, X.~Huang, and F.-l. Chung, ``Subtractive aggregation for attributed network anomaly detection,'' in \emph{CIKM}, 2021, pp. 3672--3676.

\bibitem{zhu2024anomaly}
J.~Zhu, C.~Ding, Y.~Tian, and G.~Pang, ``Anomaly heterogeneity learning for open-set supervised anomaly detection,'' in \emph{CVPR}, 2024, pp. 17\,616--17\,626.

\bibitem{zhu2024toward}
J.~Zhu and G.~Pang, ``Toward generalist anomaly detection via in-context residual learning with few-shot sample prompts,'' in \emph{CVPR}, 2024, pp. 17\,826--17\,836.

\bibitem{zhu2020beyond}
J.~Zhu, Y.~Yan, L.~Zhao, M.~Heimann, L.~Akoglu, and D.~Koutra, ``Beyond homophily in graph neural networks: Current limitations and effective designs,'' \emph{NeurIPS}, vol.~33, pp. 7793--7804, 2020.

\bibitem{zhuang2023subgraph}
Z.~Zhuang, K.~M. Ting, G.~Pang, and S.~Song, ``Subgraph centralization: a necessary step for graph anomaly detection,'' in \emph{SDM}.\hskip 1em plus 0.5em minus 0.4em\relax SIAM, 2023, pp. 703--711.

\bibitem{zhuo2023partitioning}
W.~Zhuo, Z.~Liu, B.~Hooi, B.~He, G.~Tan, R.~Fathony, and J.~Chen, ``Partitioning message passing for graph fraud detection,'' in \emph{ICLR}, 2023.

\bibitem{zou2024structural}
D.~Zou, H.~Peng, and C.~Liu, ``A structural information guided hierarchical reconstruction for graph anomaly detection,'' in \emph{CIKM}, 2024, pp. 4318--4323.

\end{thebibliography}

\appendix
\begin{appendices}
% \clearpage

% \vspace{.2cm}
\section{ALGORITHMS}\label{app:algorithm}
% \vspace{.2cm}

To gain a more in-depth understanding of Deep GAD methods, in Table \ref{tab:algorithm}, we review and summarize the key characteristics of representative algorithms from each category. Some key observations are as follows. (i) 
Most current methods focus on supervised and unsupervised methods. GNN backbone-based methods are mostly supervised methods, while Proxy task design-based and anomaly measure methods are mostly unsupervised. 
(ii)  The type of datasets used to evaluate each method is different. Supervised methods usually use data sets with real anomalies, while unsupervised methods usually use data sets with synthetic/injected anomalies. There are a small number of methods that use both types of data sets.
(iii) There have been new types of methods for GAD, such as semi-supervised settings with some labeled normal nodes, or open-set supervised GAD.

\begin{table*}[h]
\centering
\caption{Key characteristics of representative deep GAD methods ordered first by publication time and then the methodology.
% \gs{insert the links now pls}
}
% \gs{for methods published in the same year, pls group methods under the same Methodology together.}}\gs{pls add short explanations for the abbr. column names}
\label{tab:algorithm}
\setlength{\tabcolsep}{3.2mm}
\scalebox{0.9}{
\begin{tabular}{c| c | c| c| c| c| c | c }
\toprule[1pt]
 \textbf{Method (Ref.)} & \textbf{Supervision}  &\textbf{Graph Type} & \textbf{Graph Instance}    & \textbf{Anomaly Type} & \textbf{Methodology} & \textbf{Year} & \textbf{Code}\\
 \toprule[1pt]
Netwalk  \cite{yu2018netwalk}  & UnsuperviWd    & Static  & Node   & Injected & Proxy Task Design & 2020  &\href{https://github.com/haoyfan/AnomalyDAE}{Link}  \\ 
 GAAN  \cite{chen2020generative} &Unsupervised  & Static  & Node & Injected & Proxy Task Design  &2020 &\href{https://github.com/Kaslanarian/SAGOD}{Link}\\
ALARM \cite{peng2020deep}   & Unsupervised   & Static  & Node   & Injected & Proxy Task Design  &2020 &\href{https://github.com/Kaslanarian/SAGOD}{Link}\\
AdoNE \cite{bandyopadhyay2020outlier} & Unsupervised   & Static  & Node   & Injected & Proxy Task Design  &2020 &\href{https://github.com/vasco95/DONE_AdONE}{Link} \\
AANE\cite{duan2020aane}  & Unsupervised & Static & Edge  & Genuine & Anomaly Measures &2020 &N/A\\
\hline
PC-GNN \cite{liu2021pick} &Supervised  & Static  &Node    &Genuine  & GNN Backbone  &2021 &\href{https://github.com/PonderLY/PC-GNN}{Link}\\
FRAUDRE\cite{zhang2021fraudre}  &Supervised & Static & Node & Injected & GNN Backbone  &2021 &\href{https://github.com/FraudDetection/FRAUDRE}{Link}\\
DCI \cite{wang2021decoupling}      &Supervised & Static & Node  &Genuine  &GNN Backbone  &2021 &\href{https://github.com/wyl7/DCI-pytorch}{Link}\\
RARE-GNN \cite{ding2021towards} & Supervised    & Static  & Node   & Injected & Proxy Task Design &2021 &N/A \\      
CoLA \cite{liu2021cola}   &Unsupervised & Static  & Node & Injected& Proxy Task Design  &2021 &\href{https://github.com/TrustAGI-Lab/CoLA}{Link}\\
ANEMONE \cite{jin2021anemone} &Unsupervised & Static  & Node & Injected& Proxy Task Design  &2021 &\href{https://github.com/TrustAGI-Lab/ANEMONE}{Link}\\
SL-GAD \cite{zheng2021generative} \ &Unsupervised & Static  & Node & Injected & Proxy Task Design &2021&\href{https://github.com/KimMeen/SL-GAD}{Link} \\
AEGIS \cite{ding2021inductive}  &Unsupervised  & Static  & Node & Injected & Proxy Task Design &2021 &\href{https://github.com/pygod-team/pygod}{Link}\\
Meta-GDN\cite{ding2021few}   &Supervised & Static  & Node  & Injected & Proxy Task Design  &2021  &\href{https://github.com/kaize0409/Meta-GDN_AnomalyDetection}{Link} \\ 
OCGNN \cite{wang2021one}  &Unsupervised  & Static  & Node & Injected & Anomaly Measures &2021&\href{https://github.com/WangXuhongCN/OCGNN}{Link} \\
AAGNN  \cite{zhou2021subtractive} &Unsupervised  & Static  & Node & Injected & Anomaly Measures &2021&\href{https://github.com/betterzhou/AAGNN}{Link}\\
\hline
NGS \cite{qin2022explainable}  &Supervised & Static  &Node    &Genuine  & GNN Backbone &2022&\href{https://github.com/qzzdd/NGS}{Link}\\
H2-FDetector \cite{shi2022h2} &Supervised & Static& Node &Genuine  & GNN Backbone  & 2022 &\href{https://github.com/shifengzhao/H2-FDetector}{Link}\\
iGAD \cite{zhang2022dual} &Supervised & Static& Graph &Genuine  & GNN Backbone  & 2022 &N/A\\
BLS  \cite{dong2022bi}  &Supervised  & Static& Node  &Genuine   & GNN Backbone  &2022 &N/A\\
AO-GNN  \cite{huang2022auc}  &Supervised  & Static& Node  &Genuine   & GNN Backbone       &2022 &N/A           \\
DAGAD \cite{liu2022dagad}     &Supervised & Static   & Nod &Injected & GNN Backbone  &2022 &\href{https://github.com/FanzhenLiu/DAGAD}{Link}\\
BWGNN \cite{tang2022rethinking}    &Supervised & Static  & Node  &Genuine  &GNN Backbone  &2022 &\href{https://github.com/squareRoot3/Rethinking-Anomaly-Detection}{Link}\\
AMNet \cite{chai2022can}  &Supervised & Static  & Node  & Genuine & GNN Backbone  &2022 &\href{https://github.com/zjunet/AMNet}{Link}\\
CONAD \cite{xu2022contrastive} &Unsupervised & Static & Node & Both & Proxy Task Design &2022 &\href{https://github.com/zhiming-xu/conad}{Link}\\
HCM-A \cite{huang2022hop} &Unsupervised & Static  & Node & Injected & Proxy Task Design &2022 &\href{https://github.com/TianjinYellow/GraphAnomalyDetection}{Link}\\
ComGA \cite{luo2022comga}   & Unsupervised   & Static  & Node   & Injected & Proxy Task Design  &2022 &\href{https://github.com/XuexiongLuoMQ/ComGA}{Link}\\  
ResGCN \cite{pei2022resgcn} & Unsupervised   & Static  & Node   & Both & Proxy Task Design  &2022 &\href{https://bitbucket.org/paulpei/resgcn/src/master/}{Link}\\  
GCCAD\cite{chen2022gccad}     &Supervised & Static  & Graph       & Genuine & Proxy Task Design  &2022   &\href{https://github.com/THUDM/GraphCAD}{Link}  \\      
GlocaKD\cite{ma2022deep} &Unsupervised & Static  & Graph  &Genuine & Proxy Task Design  &2022 &\href{https://github.com/RongrongMa/GLocalKD}{Link} \\
Sub-CR \cite{zhang2022reconstruction} & Unsupervised   & Static  & Node   & Injected & Proxy Task Design  &2022 &\href{https://github.com/Zjer12/Sub}{Link}\\
OCGTL \cite{qiu2022raising}   &Unsupervised & Static  & Graph         &Genuine  & Anomaly Measures &2022  &\href{https://github.com/boschresearch/GraphLevel-AnomalyDetection}{Link} \\
MHGL\cite{zhou2022unseen}  &Unsupervised & Static  & Node & Genuine & Anomaly Measures  &2022 &\href{https://github.com/betterzhou/MHGL}{Link}\\
\hline
SDGG \cite{cai2023self} &Supervised & Static  & Graph & Genuine & GNN Backbone  &2023 &N/A\\
GDN \cite{gao2023alleviating}  &Supervised  & Static& Node  &Genuine   & GNN Backbone  &2023 &\href{https://github.com/blacksingular/wsdm_GDN}{Link}\\
GHRN \cite{gao2023addressing}  &Supervised & Static& Node &Genuine   & GNN Backbone  & 2023 &\href{https://github.com/blacksingular/GHRN}{Link}\\
GODM \cite{liu2023data}  &Supervised   &Static &Node &Genuine &GNN Backbone  &2023 &\href{https://github.com/kayzliu/godm}{Link}\\
DIFFAD\cite{ma2023new} &Supervised   &Static &Node  &Genuine&GNN Backbone &2023 &N/A\\
GADY \cite{lou2023gady}  &Unsupervised &  Dynamic & Node & Injected  & GNN Backone  & 2023 &\href{https://github.com/mufeng-74/GADY}{Link}\\
RAND \cite{bei2023reinforcement} &Unsupervised  & Static &Node & Both & GNN Backbone & 2023 &\href{https://github.com/YuanchenBei/RAND}{Link}\\ 
SplitGNN \cite{wu2023splitgnn}  &Supervised   &Static &Node  &Genuine&GNN Backbone &2023 &\href{https://github.com/split-gnn/splitgnn}{Link} \\
SEC-GFD \cite{xu2023revisiting}  &Supervised   &Static &Node  &Genuine&GNN Backbone &2023 &N/A \\
NSReg \cite{wang2023open}  & Supervised & Static  & Node  & Genuine  & GNN Backbone &2023&N/A \\
GmapAD \cite{ma2023towards} & Unsupervised & Static & Graph & Genuine & GNN Backbone &2023 &\href{https://github.com/XiaoxiaoMa-MQ/GmapAD}{Link}\\
RQGNN \cite{dong2023rayleigh} & Unsupervised &Static &Graph &Genunie & GNN Backbone & 2023 & \href{https://github.com/xydong127/RQGNN}{Link}\\\
AuGAN \cite{zhou2023improving}     &Supervised   &Static &Node  &Both  &GNN Backbone &2023 &\href{https://github.com/betterzhou/AugAN}{Link}  \\
HimNet \cite{niu2023graph} & Unsupervised & Static & Graph & Genuine & Proxy Task Design &2023 &\href{https://github.com/Niuchx/HimNet}{Link}\\
GGA \cite{meng2023generative}  &Supervised & Static  & Node & Genuine  &  Proxy Task Design & 2023 &N/A\\
GRADATE \cite{duan2023graph} &Unsupervised  & Static  & Node &Injected & Proxy Task Design &2023 &\href{https://github.com/FelixDJC/GRADATE}{Link} \\
GAD-NR \cite{roy2024gad} & Unsupervised   & Static  & Node   & Injected & Proxy Task Design   &2023 &\href{https://github.com/Graph-COM/GAD-NR}{Link} \\
CFAD \cite{xiao2023counterfactual} &Supervised & Static & Node & Genuine & Proxy Task Design & 2023 &\href{https://github.com/ChunjingXiao/CFAD}{Link} \\
ACT \cite{wang2023cross} & Unsupervised & Static & Node & Injected  & Proxy Task Design &2023 &\href{https://github.com/QZ-WANG/ACT}{Link} \\ 
WEDGE \cite{zhou2023learning}  & Unsupervised & Static & Node & Injected  & Proxy Task Design &2023 &N/A \\
DIF \cite{xu2023deep} &Unsupervised  & Static & Node & Genuine & Anomaly Measures & 2023 &\href{https://github.com/xuhongzuo/deep-iforest}{Link}\\
CLAD \cite{kim2023class}  & Unsupervised  & Static & Node & Injected & Anomaly Measures &2023 &\href{https://github.com/boschresearch/GraphLevel-AnomalyDetection}{Link}\\
PREM \cite{pan2023prem} & Unsupervised  & Static & Node & Injected & Anomaly Measures &2023 &\href{https://github.com/CampanulaBells/PREM-GAD}{Link}\\
HRGCN \cite{li2023hrgcn}  & Unsupervised  & Static & Graph & Genuine & Anomaly Measures &2023 &\href{https://github.com/jiaxililearn/HRGCN}{Link}\\ 
TAM \cite{hezhe2023truncated} & Unsupervised   &Static & Node & Both & Anomaly Measures &2023 &\href{https://github.com/mala-lab/TAM-master}{Link}\\
GCAD \cite{zhuang2023subgraph} &Unsupervised  & Static & Node & Injected & Anomaly Measures & 2023&\href{https://github.com/IsolationKernel/Codes/tree/main/IDK/GraphAnomalyDetection}{Link} \\
\hline
ConsisGAD \cite{chen2023consistency}  & Supervised & Static & Node & Genuine & GNN Backbone &2024 &\href{https://github.com/Xtra-Computing/ConsisGAD}{Link} \\
PMP \cite{zhuo2023partitioning}  & Supervised & Static & Node & Genuine & GNN Backbone &2024 &\href{https://github.com/Xtra-Computing/PMP}{Link} \\
HedGe \cite{zhang2024generation}& Supervised & Static & Node & Genuine & GNN Backbone &2024 &\href{https://github.com/Xtra-Computing/}{Link} \\
BioGNN \cite{gao2024graph}  & Supervised & Static & Node & Genuine & GNN Backbone &2024 &\href{https://github.com/blacksingular/Bio-GNN}{Link} \\
MITIGATE~\cite{chang2024multitask} & Supervised & Static & Node & Injected & GNN Backbone &2024 &\href{https://github.com/AhaChang/MITIGATE}{Link}  \\
GGAD  \cite{qiao2024generative} &Semi-Supervised   &Static &Node &Genuine &GNN Backbone &2024 &\href{https://github.com/mala-lab/GGAD}{Link}\\
SmoothGNN \cite{dong2024smoothgnn}   & Unsupervised   &Static &Node &Genuine &GNN Backbone &2024 &N/A \\
FGAD \cite{cai2024fgad} & Supervised & Static & Graph & Genuine & Proxy Task Design & 2024 &N/A\\
STRIPE \cite{liu2024spatial} &Unsupervised  & Dynamic & Node &  Injected  & Proxy Task design &2024 &N/A
\\
DOHSC \cite{zhang2023deep} &Unsupervised  & Static & Graph & Genuine & Anomaly Measures & 2024 &\href{https://github.com/wownice333/DOHSC-DO2HSC}{Link}\\
ARC \cite{liu2024arc}   & Supervised   &Static &Node &Both & Anomaly Measures &2024 &N/A \\
\bottomrule[1pt]
\end{tabular}
}
\end{table*}

\section{DATASETS} \label{app:dataset}

We also collected and summarized publicly available GAD data sets, including node-level, graph-level, and dynamic graph datasets. 
Typically, these datasets can be categorized into synthetic datasets with injected anomalies and real-world datasets with genuine anomalies.
Some studies inject specific types of exceptional samples into existing graph datasets, such as contextual and structure-based anomalies \cite{ding2019deep, liu2021cola}. In Table \ref{tab:table2} and Table \ref{tab:table3}, we provide some statistical information about the dataset, including the number of nodes, the data volume of edges, the ratio of anomalies, and whether the anomalies are real or injected.
\begin{table*}[h]
\centering
\caption{Publicly accessible node-level GAD datasets.
}
\label{tab:table2}
\setlength{\tabcolsep}{1.5mm}
\scalebox{1.0}{
\begin{tabular}{c | c| c| c| c| c  | c  | c| c}
\toprule[1pt]
 \textbf{Dataset} &\textbf{\# Nodes}  &\textbf{\# Edges}  & \textbf{\# Attributes}  & \textbf{Size} & \textbf{Anomaly}  & \textbf{Anomaly Type} & \textbf{Domain} & \textbf{References} \\
 \toprule[1pt]
 Cora  &  2,708 & 5,429 & 1,433 & Small  &5.5\%  & Injected & Citation Networks &\cite{ding2019deep,liu2021cola, luo2022comga, fan2020anomalydae} \\
 Citersee & 3,327  & 4,732 & 3,703 & Small  &4.5\%  &  Injected & Citation Networks &\cite{ding2019deep, liu2021cola,luo2022comga,fan2020anomalydae}\\
 ACM       & 16,484 &  71,980 & 8,337 & Medium & 3.6\% & Injected & Citation Networks &\cite{ding2019deep,liu2021cola, luo2022comga,fan2020anomalydae}\\
 BlogCatalog & 5,196  &171,743  & 8,189 &Small & 5.8\% & Injected & Social Networks  &\cite{ding2019deep, liu2021cola,luo2022comga,fan2020anomalydae}\\
 Flickr    & 7,575  & 239,738  & 12,407    &Medium  &5.2\%  &Injected & Social Networks &\cite{ding2019deep, liu2021cola,luo2022comga, fan2020anomalydae} \\
 OGB-arXiv    &169,343 & 1,166,243 & 128 & Large  & 3.5\% & Injected & Citation Networks &\cite{hu2020open, liu2021cola}\\
 Amazon  & 11,944 & 4,398,392 & 25 & Large  &  9.5\% &   Genuine &  Transaction Record & \cite{dou2020enhancing, tang2023gadbench, tang2022rethinking, hezhe2023truncated}\\
 YelpChi  & 45,954  & 3,846,979  & 32  &  Large & 14.5\% & Genuine &  Reviewer Interaction & \cite{dou2020enhancing, tang2023gadbench, tang2022rethinking,hezhe2023truncated}\\
 T-Finance &  39,357 & 21,222,543 & 10 &Large & 4.6\% & Genuine &  Transaction Record & \cite{tang2022rethinking,tang2023gadbench, gao2023addressing}\\
 T-Social & 5,781,065 & 73,105,508 &10   & Large &  3.0\% & Genuine &Social Network & \cite{tang2022rethinking,tang2023gadbench, gao2023addressing}\\
 Weibo &8,405 & 407,963  & 400 & Small &  10.3\% & Genuine &  Under Same Hashtag & \cite{tang2022rethinking,tang2023gadbench}\\
 DGraph & 3,700,550 &4,300,999  & 17 &Large &  1.3\% & Genuine &  Loan Guarantor & \cite{huang2022dgraph, tang2023gadbench}\\
 Elliptic & 203,769 & 234,355 &  166& Large & 9.8\% & Genuine  &  Payment Flow & \cite{tang2023gadbench, chai2022can, dong2024smoothgnn, wang2023open}\\
Tolokers & 11,758& 519,000 & 10 &Medium &  21.8\%& Genuine & Work Collaboration & \cite{tang2023gadbench, dong2024smoothgnn} \\ 
Questions & 48,921 &153,540 & 301 & Medium &3.0\% &Genuine &Question Answering & \cite{tang2023gadbench, dong2024smoothgnn}\\ 
 Disney &  124 &335  & 28 & Small  &  4.8\%  & Genuine &Co-purchase  &\cite{liu2022bond, roy2024gad}\\ 
 Books & 1,418 &3,695 & 21 &  Small & 2.0\% & Genuine & Co-purchase &\cite{liu2022bond,roy2024gad}\\
 Enron & 13,533 & 176,987&18 & Medium & 0.4\%  &Genuine  &  Email network &\cite{liu2022bond, roy2024gad}\\
 Reddit &  10,984 & 168,016 &64  & Medium & 3.3\%   &Genuine  & User-subreddit & \cite{liu2022bond, tang2023gadbench,hezhe2023truncated, qiao2024generative}\\
\bottomrule[1pt]
\end{tabular}
}
\end{table*}

\begin{table*}[h]
\centering
\caption{Publicly accessible graph-level GAD datasets. Homo. and Heter. indicate the graph is homogeneous and heterogeneous, respectively. Graph-level GAD methods are typically trained using anomaly-free data, so the anomaly rate is applied to the test data only.}
% \gs{pls add HimNet into the Ref. column where applicable. pls also add the heterogeneous graph datasets used in HRGCN; and add a column to indicate whether the graphs are homogeneous or heterogeneous.} \hz{Two hete. dataset and information have been added in the table.}}
\label{tab:table3}
\setlength{\tabcolsep}{2.0mm}
\scalebox{1.0}{
\begin{tabular}{c | c| c| c| c | c| c | c}
\toprule[1pt]
  \textbf{Dataset} &  \textbf{\# Graphs} &  \textbf{\# Avg. Nodes} & \textbf{\# Edges} & \textbf{Anomaly}  & \textbf{Domain} &\textbf{Homo./Heter.}   & \textbf{References}  \\
 \toprule[1pt]
  KKI  & 83  & 190  &237.4 & 44.6\% & Bioinformatics & Homo.&\cite{ma2022deep, ma2023towards,wang2021one} \\
  OHSU &  79 &	82.01 &	199.66 & 44.3\% & Bioinformatics & Homo.&\cite{ma2022deep, ma2023towards} \\
  MUTAG & 188  & 	17.93 &	19.79  & 33.5\% & Molecules & 
   Homo.&\cite{ma2022deep, ma2023towards, liu2024towards}\\
  PROTEINSfull &  1,113 & 39.06 & 72.82  & 40.4\% & Bioinformatics & Homo.&\cite{ma2022deep, ma2023towards,wang2021one,liu2024towards}\\
  ENZYMES & 600 & 32.63& 62.14  &16.7\% &Bioinformatics &  Homo.&\cite{ma2022deep}\\
  AIDS& 2,000 &15.69 &16.2 & 20.0\% & Chemical Structure & Homo.&\cite{ma2022deep, ma2023towards, wang2021one, liu2024towards} \\
  BZR &405 &35.75 &38.36 & 21.0\% & Molecules & Homo. &\cite{ma2022deep, wang2021one, liu2024towards}\\
  COX2 &467 &41.22& 43.45 &21.8\% & Molecules & Homo.&\cite{ma2022deep,wang2021one,liu2024towards}\\
  DD& 1,178 &284.32 &715.66 &41.3\% & Bioinformatics & Homo.&\cite{ma2022deep,wang2021one,liu2024towards} \\
  NCI1& 4,110 &29.87 &32.3 & 49.9\% & Molecules &Homo. &\cite{ma2023towards,wang2021one,liu2024towards}\\
  IMDB& 1,000 &19.77 &96.53 & 50.0\% & Social Networks & Homo.&\cite{ma2023towards,liu2024towards}\\
  REDDIT& 2,000 &429.63 &497.75 & 50.0\% & Social Networks & Homo.&\cite{ma2022deep, ma2023towards, wang2021one, liu2024towards}\\
  HSE &8,417 &16.89 &17.23 & 5.2\% & Molecules &  Homo. &\cite{ma2022deep,wang2021one} \\
  MMP &7,558 &17.62 &17.98 & 15.6\% & Molecules & Homo.&\cite{ma2022deep,wang2021one} \\
  p53 &8,903 &17.92 &18.34 & 6.3\%  & Molecules &Homo.&\cite{ma2022deep,wang2021one}\\
  PPAR-gamma & 8,451 &17.38 & 17.72 & 2.8\% & Molecules &Homo.&\cite{ma2022deep,wang2021one}\\
  COLLAB &5,000& 74.49& 2,457.78 & 15.5\% & Social Networks & Homo.&\cite{ma2022deep,wang2021one}\\
 Mutagenicit & 	4,337	 &30.32	 & 30.77 &44.6\% &  Molecules  &Homo. &\cite{ma2023towards}  \\
 DHFR &756 &42.43 &44.54 &39.0\%  & Molecules & Homo.&\cite{ma2022deep,wang2021one,liu2024towards}\\ 
 TraceLog & 132,485 & 205 & 224 & 17.6\% & Log Sequences & Heter. & \cite{li2023hrgcn}\\
 FlowGraph &  600& 8,411& 12,730 & 16.7\%& System Flow  & Heter. & \cite{li2023hrgcn} \\
\bottomrule[1pt]
\end{tabular}
}
\end{table*}

\section{QUANTITATIVE COMPARISON} \label{app:comparison}
In this subsection,  we perform an empirical comparison of different graph anomaly detection methods.  The way we compare the performance of different methods is to collect experimental results using the same datasets from their original papers. Tables \ref{tab:Quantitative_node1} and \ref{tab:Quantitative_node2} provide the results of AUROC and AUPRC, two widely used metrics in anomaly detection, for various methods on both synthetic and real datasets for node-level anomaly detection, while Table \ref{tab:graph_level_ad} presents the results of graph-level anomaly detectors on both homogeneous and heterogeneous graph datasets. Additionally, we collect the F1 performance under two setting with 40\% and 1\% labeled nodes used during training on four datasets, with the results shown in Table \ref{tab:macro_f1}. To evaluate the scalability of each method, we also provide a comparison of their runtime in Table \ref{tab:running_time}. It should be noted that the runtime may not be directly comparable across the methods, as these computational costs may be obtained using different experimental environments.

\section{EXPLANATION/ANALYSIS ON PERFORMANCE COMPARISON} \label{app:comparison}
From the empirical comparison results, we can observe that supervised methods generally outperform unsupervised methods, as they leverage labeled information in their detection model learning. However, this does not always hold, particularly on datasets with injected anomalies, where unsupervised methods like reconstruction-based and contrastive learning-based methods achieve the best performance. This may be because supervised methods tend to overfit the limited training anomaly data, while the designs in some unsupervised methods are built upon some prior knowledge that is used for generating the injected anomalies. Additionally, the superior performance of most methods are limited to specific datasets due to their specialized design tailored to those datasets.
Some methods such as reconstruction-based are unable to scale up to very large-scale datasets, such as T-Social and Dgraph. For graph-level anomaly detection, only AUROC results are typically available, as AUPRC is not reported in most studies in this line. Most graph-level anomaly detection methods are unsupervised and designed for homogeneous graphs. 

Since AUROC and AUPRC are widely used metrics for anomaly detection tasks, only a few studies report F1 performance, and all of these methods are supervised. As the number of labeled nodes increases, the F1 performance generally gets improved.

For the runtime of each method,  unsupervised methods typically require significantly more time than semi-supervised and supervised methods due to their inherently high complexity, often involving tasks such as reconstruction, contrastive learning, and local node affinity calculation. In contrast, supervised methods are optimized using a supervised loss function, which generally results in lower computational costs.

\begin{table*}[!ht]
    \centering
    \caption{Quantitative comparison of node-level anomaly detection on datasets with manually injected (synthetic) anomalies. }
    \label{tab:Quantitative_node1}
  % \vskip 0.15in
   \centering
\begin{tabular}{c|c|cccccccccc}
    \hline
    \multirow{2}*{\textbf{Metric}}&\multirow{2}*{\textbf{Method}} & \multicolumn{9}{c}{\textbf{Dataset}}\\
         \multirow{16}*{AUROC} &  & Cora & Citeseer & ACM & BlogCatalog & Flicker & Pubmed & Facebook & Reddit & Weibo  \\
           \hline
          & DOMINANT \cite{ding2019deep} & 0.815 & 0.825 & 0.760 & 0.746 & 0.744 & 0.808 & 0.554 & 0.560 & 0.850  \\ 
        ~ & CoLA \cite{liu2021cola} & 0.878 & 0.896 & 0.823 & 0.785 & 0.751 & 0.951 & / & 0.603 & /  \\
        ~ & SL-GAD \cite{zheng2021generative} & 0.913 & 0.913 & 0.853 & 0.818 & 0.796 & 0.967 & / & 0.567 & /  \\ 
        ~ & CONAD \cite{xu2022contrastive} & 0.788 & / & / & / & / & / & 0.863 & 0.561 & 0.854  \\ 
         ~ & AEGIS \cite{ding2021inductive} & / & /& /& 0.743 &0.738 &0.773 & / & / & /\\
        ~ & OCGNN \cite{wang2021one} & 0.881 & 0.856 & / & / & / & 0.747 & 0.793 & / & /  \\ 
        ~ & ComGA \cite{luo2022comga}  & 0.884 & 0.916 & 0.849 & 0.814 & 0.799 & 0.922 & 0.659 & / & /  \\ 
        ~ & AAGNN \cite{zhou2021subtractive} & / & / & / & 0.818 & 0.829 & 0.856 & / & / & 0.925  \\ 
        ~ & HCM-A \cite{huang2022hop} & / & / & 0.761 & 0.798 & 0.792 & / & / & / & /  \\ 
        ~ & GAAN  \cite{chen2020generative} & 0.742 & / & 0.877 & 0.765 & 0.753 & / & / & 0.554 & 0.925  \\
        ~ & AnomalyDAE \cite{fan2020anomalydae} & 0.762 & 0.727 & 0.778 & 0.783 & 0.751 & 0.810 & / & 0.557 & 0.915  \\ 
        ~ & GAD-NR \cite{roy2024gad} & 0.835 & / & / & / & / & / & / & / & 0.623  \\ 
        ~ & TAM \cite{hezhe2023truncated} & / & / & 0.887 & 0.824 & / & / & 0.914 & 0.602 & /  \\ 
        \hline
         \multirow{14}*{AURPC} & DOMINANT \cite{ding2019deep} & 0.200 & / & / & 0.338 & 0.324 & 0.299 & / & 0.037 & /  \\ 
        ~ & CoLA \cite{liu2021cola} & / & / & 0.323 & 0.327 & / & / & 0.211 & 0.044 & /  \\ 
        ~ & SL-GAD  \cite{zheng2021generative} & / & / & / & 0.388 & 0.378 & / & 0.131 & 0.041 & /  \\ 
        ~ & CONAD \cite{xu2022contrastive} & / & / & / & / & / & / & / & 0.037 & /  \\ 
        ~ & AEGIS \cite{ding2021inductive} & / & / & / & 0.339 & 0.324 & 0.373 & / & / & /  \\ 
        ~ & OCGNN \cite{wang2021one} & / & / & / & / & / & / & / & / & /  \\ 
        ~ & ComGA \cite{luo2022comga} & / & / & / & / & / & / & / & / & /  \\ 
        ~ & AAGNN \cite{zhou2021subtractive} & / & / & / & 0.435 & 0.421 & 0.428 & / & / & /  \\ 
        ~ & HCM-A \cite{huang2022hop}& / & / & / & / & / & / & / & / & /  \\ 
        ~ & GAAN  \cite{chen2020generative} & / & / & / & 0.338 & 0.324 & 0.337 & / & 0.037 & /  \\ 
        ~ & AnomalyDAE \cite{fan2020anomalydae} & 0.183 & / & / & / & / & / & / & / & /  \\ 
        ~ & GAD-NR \cite{roy2024gad} & / & / & / & / & / & / & / & / & /  \\ 
        ~ & TAM \cite{hezhe2023truncated} & / & / & 0.512 & 0.418 & / & / & 0.223 & 0.044 & /  \\ 
        \hline
\bottomrule
    \end{tabular}
\end{table*}

\begin{table*}[!ht]
    \centering
        \caption{Quantitative comparison of node-level anomaly detection on datasets with genuine anomalies. Results of DevNet and PReNet are taken from \cite{wang2023open}. 
        }
        \label{tab:Quantitative_node2}
          \setlength{\tabcolsep}{0.8mm}
          \scalebox{0.9}{
\begin{tabular}{c|c|c|ccccccccccccc}
 \hline
    \multirow{2}*{\textbf{Metric}}&  \multirow{2}*{\textbf{Setting}}& \multirow{2}*{\textbf{Method}} & \multicolumn{12}{c}{\textbf{Dataset}}\\
    
        ~ & ~ & ~ & Amazon  & YelpChi &  T-Finance  & Question  & Elliptic &  Reddit &  Tolokers &  Weibo &  DGraph  & T-Social & Photo & CS    \\ \hline
      \multirow{17}*{AUROC} & \multirow{7}*{Unsupervised}& DOMINANT \cite{ding2019deep} &0.694 & 0.539 &0.538 & / &0.296 & 0.556 & / &  /  & 0.574 &  /&  0.514 & 0.402  \\
      ~ & ~ & CoLA \cite{liu2021cola} & 0.261 & 0.480 & 0.483 & / & / & 0.603 & / &  / &  / &  /&  / & 0.481\\
        ~ & ~ & CLAD \cite{kim2023class}   & 0.203  & 0.476   & 0.139  & 0.621   & 0.419  &0.578 & 0.406 & / & / &  /& /&  /& \\
          ~ & ~ & GRADATE \cite{duan2023graph}  & 0.478  &   0.492  &   0.406  & 0.554  & /   & 0.526  & 0.537&/ & /&  /&/ & /&\\
       ~ & ~ & GAD-NR \cite{roy2024gad} & 0.260 & 0.470 & 0.579 & 0.587 & 0.400 & 0.553  & 0.576 & / & / & / & / & /  \\
        ~ & ~ & Prem \cite{pan2023prem}& 0.278 & 0.490 & 0.448 & 0.603 & 0.497 & 0.551 & 0.565 & / & / & /  & / & / \\
        ~ & ~ & TAM \cite{hezhe2023truncated} & 0.802 & 0.548 & 0.690 & 0.504 & / & 0.572 & 0.469 & / & / & / & / & / \\ \
        ~ & ~ & SmoothGNN \cite{dong2024smoothgnn} & 0.840 & 0.575 & 0.755 & 0.644 & 0.572 & 0.594 & 0.687 & / & 0.649 & 0.703& / & /  \\ \cline{2-15}
         ~ & Semi-supervised & GGAD \cite{qiao2024generative} &  0.944 & 	/ & 0.823&	/&	0.729&	/&	/&	/& 0.594&	/&	0.648&	/ \\ \cline{2-15}
        ~ & \multirow{10}*{Supervised} & BWGNN \cite{tang2022rethinking} & 0.980 & 0.849 & 0.961 & 0.718 & 0.852 & 0.654 & 0.804 & 0.973 & 0.763 & 0.920 & / & /  \\ 
        ~ & ~ & DCI \cite{wang2021decoupling} & 0.946 & 0.778 & 0.868 & 0.692 & 0.828 & 0.665 & 0.755 & 0.942 & 0.747 & 0.808 & / & /  \\ 
        ~ & ~ & AMNet \cite{chai2022can} & 0.970 & 0.826 & 0.937 & 0.681 & 0.773 & 0.684 & 0.768 & 0.953 & 0.731 & 0.536  & / & / \\ 
        ~ & ~ & GHRN \cite{gao2023alleviating} & 0.981 & 0.853 & 0.960 & 0.718 & 0.854 & 0.660 & 0.804 & 0.967 & 0.761 & 0.790 & / & / \\ 
        ~ & ~ & NGS \cite{qin2022explainable} & 0.973 & 0.921 & / & / & / & / & / & / & / & /  & / & / \\ 
        ~ & ~ & PCGNN \cite{liu2021pick} & 0.973 & 0.797 & 0.933 & 0.699 & 0.858 & 0.532 & 0.728 & 0.902 & 0.720 & 0.692 & / & / \\ 
        ~ & ~ & GDN \cite{gao2023addressing} & 0.971 & 0.903 & / & / & / & / & / & / & / & / & / & / \\   
        ~ & ~ & DevNet \cite{pang2019deep} & /	& /	& 0.654& 	/	& /	& /& 	/& 	/	& /	& /& 	0.599& 	0.606 \\
        ~ & ~ & PReNet \cite{pang2023deep} & /& 	/& 	0.892& 	/& 	/	& /& 	/	& /	& /	& /	& 0.698	& 0.632 \\
                ~ & ~ & NSReg \cite{wang2023open} &  /& 	/	& 0.929& 	/& 	/	& /	& /	& /	& /	& /	& 0.908	& 0.797 \\
          \hline
       \multirow{17}*{AUPRC}& \multirow{7}*{Unsupervised} & DOMINANT \cite{ding2019deep} &0.102 &0.165 & 0.047  & / & / &0.036 & / & & 0.008 &  /&  0.104 & 0.187\\
      ~ & ~ & CoLA \cite{liu2021cola} &0.052 &0.136 &0.041 & / &  / &  0.045 & / & / & / &  /& 0.246 & 0.253\\
      ~ & ~ & CLAD \cite{kim2023class} & 0.040  & 0.128   & 0.025  &   0.051 & 0.081  &0.050 & 0.192 & / & / &  /&  / & / &\\
          ~ & ~ & GRADATE \cite{duan2023graph} & 0.063  &  0.145  &  0.038 &  0.035  & /  &0.039 & 0.236 & / & / &  /&  / &  /&\\
       ~ & ~ & GADNR \cite{roy2024gad} & 0.042 & 0.139 & 0.054 & 0.057 & 0.077 & 0.037 & 0.299 & / & / & /& / & /  \\ 
        ~ & ~ & Prem \cite{pan2023prem} & 0.074 & 0.137 & 0.039 & 0.043 & 0.090 & 0.041 & 0.259 & / & / & / & / & / \\ 
        ~ & ~ & TAM \cite{hezhe2023truncated} & 0.332 & 0.173 & 0.128 & 0.039 & / & 0.042 & 0.196 & / & / & / & / & /  \\
        ~ & ~ & SmoothGNN \cite{dong2024smoothgnn} & 0.395 & 0.182 & 0.140 & 0.059 & 0.116 & 0.043  & 0.351 & / & 0.019 & 0.063  & / & / \\  \cline{2-15}
         ~ & Semi-supervised & GGAD \cite{qiao2024generative}&0.792	&/	&0.183	&/&	0.243&	0.061&	/	&/&	0.008&	/&	0.144&	/ \\ \cline{2-15}
        ~ & \multirow{10}*{Supervised} & BWGNN \cite{tang2022rethinking}& 0.891 & 0.551 & 0.866 & 0.167 & 0.260 & 0.069 & 0.497 & 0.930 & 0.040 & 0.549  & / & / \\ 
        ~ & ~ & DCI \cite{wang2021decoupling}& 0.815 & 0.395 & 0.626 & 0.141 & 0.254 & 0.061 & 0.399 & 0.896 & 0.036 & 0.138 & / & /   \\ 
        ~ & ~ & AMNet \cite{chai2022can} & 0.873 & 0.488 & 0.743 & 0.146 & 0.147 & 0.073 & 0.432 & 0.897 & 0.028 & 0.031  & / & / \\ 
        ~ & ~ & GHRN \cite{gao2023addressing} & 0.895 & 0.566 & 0.866 & 0.167 & 0.277 & 0.072 & 0.499 & 0.918 & 0.04 & 0.163 & / & / \\
        ~ & ~ & NGS \cite{qin2022explainable} & / & / & / & / & / & / & / & / & / & /  & / & /   \\ 
        ~ & ~ & PCGNN  \cite{liu2021pick} & 0.878 & 0.437 & 0.698 & 0.144 & 0.356 & 0.042 & 0.381 & 0.819 & 0.028 & 0.087  & / & / \\ 
        % ~ & ~ & GDN \cite{gao2023alleviating} & / & / & / & / & / & / & / & / & / & / & / & / \\
         ~ & ~ & DevNet \cite{pang2019deep} &/&	/	&0.323&	/	&/&	/	&/	&/	&/&	/&	0.468&	0.537 \\
        ~ & ~ & PReNet \cite{pang2023deep}& /	&/	&0.571&	/&	/&	/&	/&	/&	/&	/&	0.460&	0.557 \\
         ~ & ~ & NSReg \cite{wang2023open} &/&	/	&0.757&	/	&/&	/&	/&	/&	/&	/&	0.836&	0.752 \\
          \hline
    \bottomrule
    \end{tabular}}
\end{table*}

\begin{table*}[!ht]
    \centering
       \caption{Quantitative comparison of graph-level anomaly detection.
       }\label{tab:graph_level_ad}
   \setlength{\tabcolsep}{0.7mm}
   \scalebox{0.9}{
   \begin{tabular}{c|c|ccccccccccccc|cc}
    \hline
        \multirow{2}*{\textbf{Metric}}&\multirow{2}*{\textbf{Method}} & \multicolumn{13}{c}{\textbf{Homogeneous Graphs}} & \multicolumn{2}{|c}{\textbf{Heterogeneous Graphs}}\\
        \cline{3-17}
        ~ & ~ & PROTEINS-F &  ENZYMES  & AIDS &  DHFR  & BZR &  COX2 &  DD  & NCI1 &  IMDB & COLLAB & HSE& MMP & P53 &  TraceLog& FlowGraph \\ \hline
       \multirow{9}*{AUROC} & GlocalKD \cite{ma2022deep} & 0.773 & 0.613 & 0.932 & 0.567 & 0.694 & 0.593 & 0.801 & 0.684 & 0.521 & 0.674  & 0.593 & 0.675  & 0.640 &/& /\\
        ~ & OCGIN \cite{zhao2023using} & 0.708 & 0.587 & 0.781 & 0.492 & 0.659 & 0.535 & 0.722 & 0.719 & 0.601 &/ & / & / & / &/& / \\ 
        ~ & SIGNET \cite{liu2024towards} & 0.752 & 0.629 & 0.972 & 0.740 & 0.814 & 0.714 & 0.727 & 0.748 & 0.664 &/  &  /& / & / &/&/ \\ 
        ~ & OCGTL \cite{qiu2022raising} & 0.765 & 0.620 & 0.994 & 0.599 & 0.639 & 0.552 & 0.794 & 0.734 & 0.640 & / & / & / & / &/& /\\
        ~ & OCGCN  \cite{wang2021one} & 0.718 & 0.613 & 0.664 & 0.495 & 0.658 & 0.628 & 0.605 & 0.627 & 0.536 &/ & 0.388  & 0.457  &0.483  &/& / \\ 
        &HimNet \cite{niu2023graph} & 0.772 & 0.589 &0.997 & 0.701 & 0.703 & 0.637 & 0.806 & 0.686 & 0.553 & 0.683  & 0.613 & 0.703 & 0.646 & / &/  \\
        ~ & GLADST \cite{lin2023discriminative} & / & 0.694 & 0.976 & 0.773 & 0.810 & 0.630 & / & 0.681 & / & 0.776 & 0.547 & 0.685 &0.688  &/& /\\ 
         & DIF \cite{xu2023deep}  & / & / & / & / & / & / & / & / & / & / & 0.737  & 0.715 & 0.680 & / &/   \\
          &HRGCN \cite{li2023hrgcn} & / & / & / & / & / & / & / & / & / & / & / & /  & / & 0.864 & 1.000\\
          \hline
              \bottomrule
    \end{tabular}}
\end{table*}
\end{appendices}

\begin{table*}[!ht]
    \centering
          \caption{F1 comparison of node-level anomaly detection
        }\label{tab:macro_f1}
         \setlength{\tabcolsep}{0.7mm}
   \scalebox{0.9}{
    \begin{tabular}{c|c|cccccc}
    
    \hline
       \multirow{2}*{\textbf{Metric}}&\multirow{2}*{\textbf{Method}} &   \multicolumn{4}{c}{\textbf{Datasets}} \\ \cline{3-6}
               &  & Amazon  & YelpChi &  T-Finance & T-Social\\
            \hline
       
         \multirow{10}*{40\%} & CARE-GNN \cite{dou2020enhancing} & 0.863 & 0.633 & 0.825 & 0.518  \\
        ~ & PC-GNN \cite{liu2021pick} & 0.895 & 0.630 & 0.558 & 0.441  \\ 
        ~ & BWGNN \cite{tang2022rethinking} & 0.917 & 0.769 & 0.886 & 0.840  \\ 
        ~ & H2-FDetector \cite{shi2022h2} & 0.869 & 0.743 & 0.742 & /  \\ 
        ~ & GHRN \cite{gao2023addressing} & 0.923 & 0.775 & 0.879 & 0.682  \\ 
        ~ & GDN \cite{pang2019deep} & 0.906 & 0.760 & 0.887 & 0.568  \\ 
        ~ & CONSISGAD \cite{chen2023consistency} & / & / & / & /  \\ 
        ~ & PMP \cite{zhuo2023partitioning} & 0.920 & 0.819 & 0.919 & 0.952  \\ 
        ~ & GLHAD \cite{guo2024graph} & 0.932 & 0.823 & 0.923 & 0.942  \\ 
        ~ & LEX-GNN \cite{hyun2024lex} & 0.934 & 0.863 & / & /  \\ \hline
       \multirow{10}*{1\%} & CARE-GNN \cite{dou2020enhancing} & 0.757 & 0.616 & 0.836 & /  \\ 
        ~ & PC-GNN \cite{liu2021pick}  & 0.852 & 0.642 & 0.869 & 0.496  \\ 
        ~ & BWGNN \cite{tang2022rethinking}& 0.904 & 0.612 & 0.869 & 0.763  \\ 
        ~ & H2-FDetector \cite{shi2022h2} & 0.724 & 0.639 & / & /  \\ 
        ~ & GHRN \cite{gao2023addressing}& 0.863 & 0.613 & 0.8 & 0.712  \\ 
        ~ & GDN \cite{pang2019deep} & 0.897 & 0.648 & 0.766 & 0.557  \\ 
       ~ & CONSISGAD \cite{chen2023consistency} & 0.900  & 0.697 & 0.909 & 0.780  \\ 
        ~ & PMP \cite{zhuo2023partitioning}& 0.877 & 0.686 & 0.888 & 0.845  \\ 
        ~ & GLHAD \cite{guo2024graph} & / & / & / & /  \\
        ~ & LEX-GNN \cite{hyun2024lex} & 0.873  & 0.697 & / & /  \\ 
           \bottomrule
    \end{tabular}
    }
\end{table*}

\begin{table*}[!ht]
    \centering
   
        \caption{Runtime (in seconds) comparison of node-level anomaly detection }\label{tab:running_time}
         \setlength{\tabcolsep}{0.7mm}
   \scalebox{0.9}{
    \begin{tabular}{c|ccccccccc}
    \hline
    \multirow{2}*{\textbf{Method}} &   \multicolumn{8}{c}{\textbf{Datasets}} \\
        ~ & Amazon    & YelpChi & T-Finance & T-Social & Reddit &  Elliptic  & DGraph & Tolokers & Questions  \\ \hline
        DOMINANT \cite{ding2019deep}& 1,592 & / & 10,721 & / & 125 & 1,119 & / & / & /  \\ 
        AnomalyDAE \cite{fan2020anomalydae} & 1,656 & / & 18,560 & / & 161 & 8,296 & / & / & /  \\ 
        OCGNN \cite{wang2021one} & 765 & / & 5,717 & / & 162 & 3,517 & / & / & /  \\ 
        AEGIS \cite{ding2021inductive} & 1,121 & / & 15,258 & / & 166 & 5,638 & / & / & /  \\ 
        GAAN \cite{chen2020generative} & 1,678 & / & 12,120 & / & 94 & 1,866 & / & / & /  \\ 
        TAM \cite{hezhe2023truncated}& 5,050 & 102,232 & 17,360 & / & 432 & 13,200 & / & 5,668 & 11,603  \\ 
        GGAD \cite{qiao2024generative} & 658 & / & 9,345 & / & 368 & 5,146 & 488 & / & /  \\ 
        GADNR \cite{roy2024gad} & 2,048 & 5,046 & 14,255 & / & 692 & 12,568 & / & 861 & 2,795  \\ 
        SmoothGNN \cite{dong2024smoothgnn} & 7 & 19 & 16 & 4,877 & 7 & 205 & 2,924 & 6 & 32  \\ 
        PREM \cite{pan2023prem}& 1,267 & 308 & 266 & / & 73 & 2,149 & / & 74 & 409  \\ 
        CARE-GNN \cite{dou2020enhancing}& / & / & 572 & 9,159 & / & / & / & / & /  \\ 
        PC-GNN \cite{liu2021pick} & / & / & 736 & 13,958 & / & / & / & / & /  \\
        BWGNN \cite{tang2022rethinking}& / & / & 31 & 2,707 & / & / & / & / & /  \\ 
     \bottomrule
    \end{tabular}
    }
\end{table*}

\end{document}